\documentclass[10pt,twocolumn,letterpaper]{article}

\usepackage{cvpr}
\usepackage{times}
\usepackage{epsfig}
\usepackage{graphicx}
\usepackage{amsmath}
\usepackage{amssymb}
\usepackage{subfig}
\usepackage{verbatim}
\usepackage{multirow}

\usepackage[breaklinks=true,bookmarks=false]{hyperref}

\cvprfinalcopy 

\def\cvprPaperID{****} 

\begin{document}

\title{Zooming into Face Forensics: A Pixel-level Analysis}

\author{Jia Li$^1$, Tong Shen$^2$, Wei Zhang$^2$, Hui Ren$^1$, Dan Zeng$^3$, Tao Mei$^2$\\
	{\small $^1$Communication University of China}\\
	{\small $^2$JD AI Research}\\
	{\small $^3$Shanghai University}
}

\maketitle

\begin{abstract}
The stunning progress in face manipulation methods has made it possible to synthesize realistic fake face images, which poses potential threats to our society. It is urgent to have face forensics techniques to distinguish those tampered images. A large scale dataset “FaceForensics++” has provided enormous training data generated from prominent face manipulation methods to facilitate anti-fake research. However, previous works focus more on casting it as a classification problem by only considering a global prediction. Through investigation to the problem, we find that training a classification network often fails to capture high quality features, which might lead to sub-optimal solutions. In this paper, we zoom in on the problem by conducting a pixel-level analysis, \ie formulating it as a pixel-level segmentation task. By evaluating multiple architectures on both segmentation and classification tasks, We show the superiority of viewing the problem from a segmentation perspective. Different ablation studies are also performed to investigate what makes an effective and efficient anti-fake model. Strong baselines are also established, which, we hope, could shed some light on the field of face forensics.
\end{abstract}

\section{Introduction}

Human faces play an important role in human communication, as a face is associated with the identity of a person. The unique face information, working as fingerprints, has been used in many applications such as phone unlocking, payment, etc., thanks to remarkable progress in face detection and recognition systems \cite{taigman2013deepface, sun2014deep, schroff2015facenet}. However, we have also seen stunning progress in image and video manipulation methods, which enable editing the images or videos in a visually plausible way. 
Some face specific manipulation methods \cite{deepfake, Faceswap, 2019Neural_Textures, Thies_2016_CVPR} are able to manipulate the face image of person and create an indistinguishable fake image. 
\begin{figure}[h]
	\setlength{\abovecaptionskip}{0.cm}
	
	\setlength{\belowcaptionskip}{-0.cm}
	\begin{center}
		\setlength{\tabcolsep}{0.3em}
		\begin{tabular}{cccc}
			\includegraphics[width=0.22\linewidth]{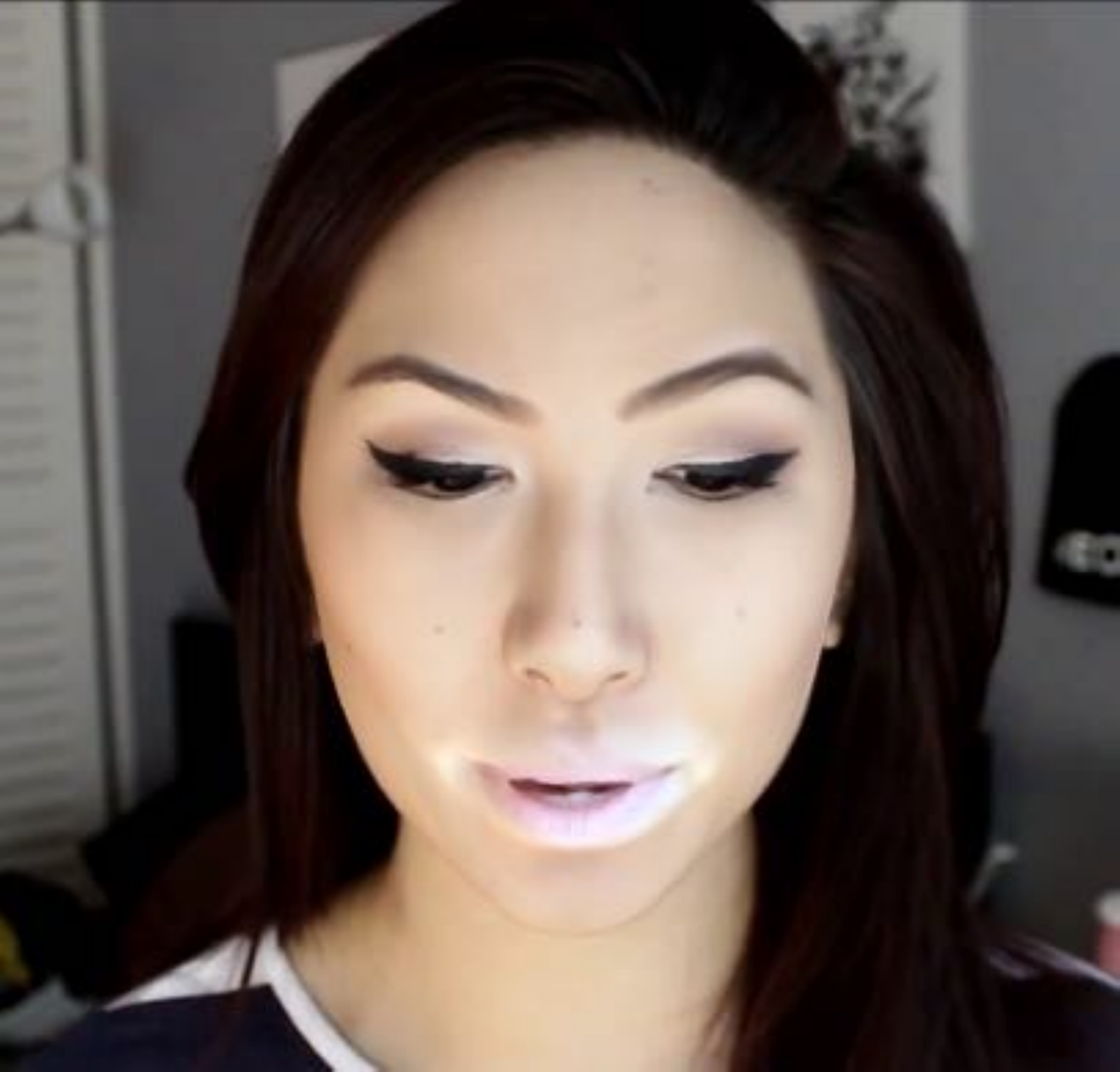} &
			\includegraphics[width=0.22\linewidth]{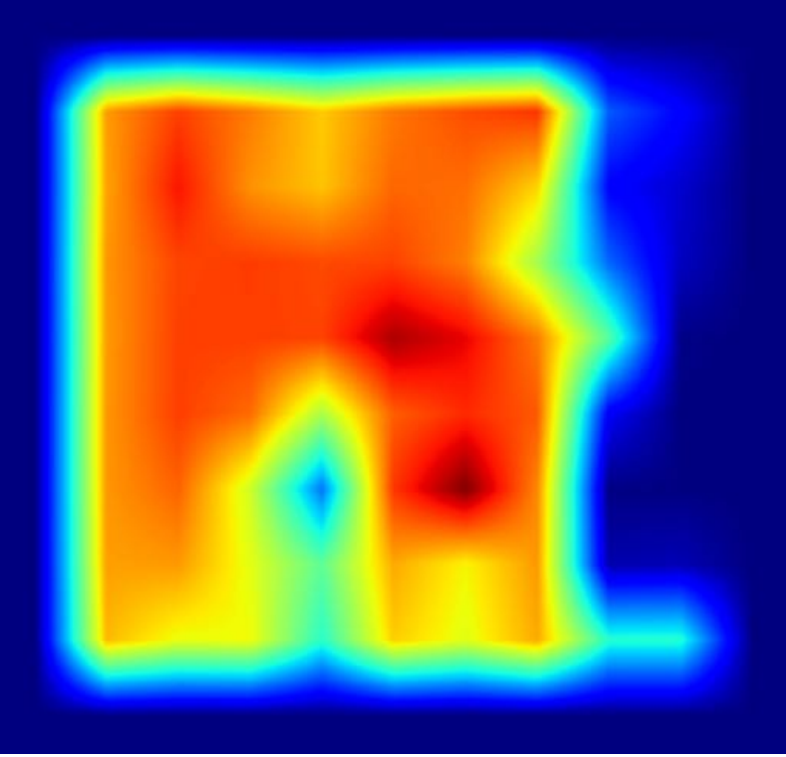} &
			\includegraphics[width=0.22\linewidth]{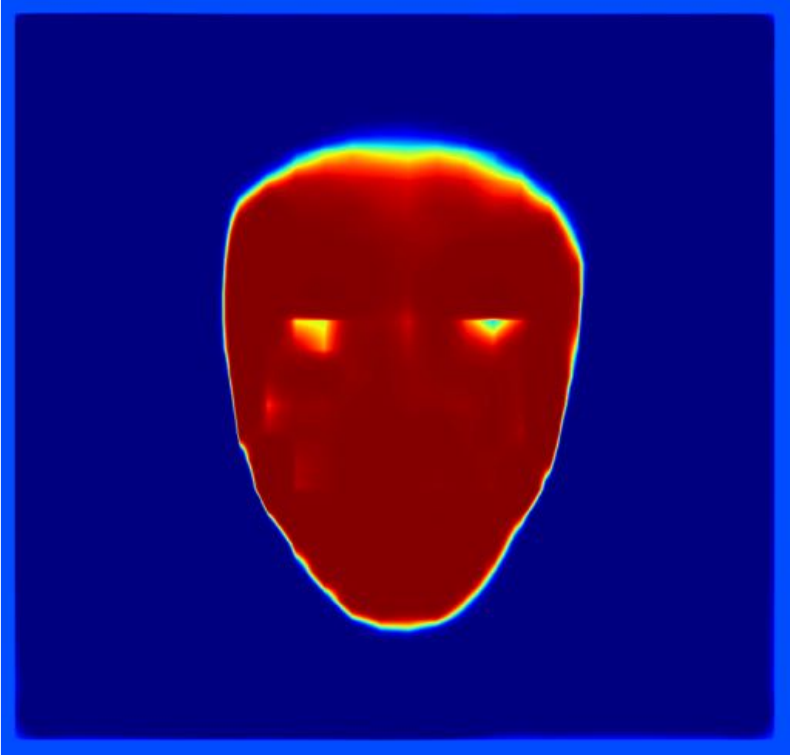} &
			\includegraphics[width=0.22\linewidth]{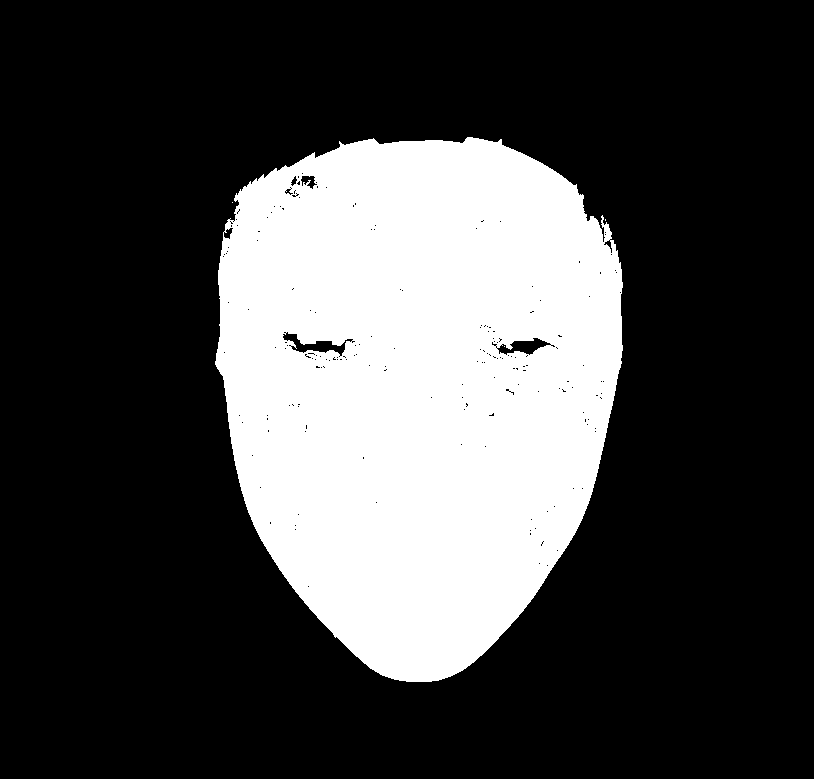} \\			
			Image & Cls & Seg & GT
	\end{tabular}
	\end{center}
	\caption{Predictions of a classification network and a segmentation network. The second image is the activation map of the classification network showing the high-response area. The third is the heatmap given by the segmentation network. Compared with the ground-truth on the right, the segmentation network localizes the tampered pixels on a far accurate level.}
	\label{fig:cls}
\end{figure}
Current face manipulation methods can be roughly divided into two categories, facial reenactment and identity swap. Facial reenactment tries to transfer the facial expressions of one person to another person and synthesize realistic details. Face2Face \cite{Thies_2016_CVPR} and NeuralTextures \cite{2019Neural_Textures} are two representative works. Identity swap is a technique that enables replacing the face of a person with another person's face. Deepfakes \cite{deepfake} and FaceSwap \cite{Faceswap} are two of the most prominent methods. These methods enable effortless creation of fake face images and videos, which poses potential threats to our society. For example, fake news can be easily created by synthesizing a speech video of a politician \cite{Suwajanakorn:2017}.

To alleviate the potential issues caused by the fake face videos and images, great efforts have been dedicated to the field of face forensics, which aims to determine authenticity of a face photo. General image forensics techniques, relying on hand-crafted cues \cite{Farid:2016, Popescu:2005, Double_JPEG_Detection, Obrien:2012:EPM}, might not be suitable for face specific forensics tasks since faces are highly structured data. Recent works take advantage of great representation power of CNNs (Convolutional Neural Networks) and train a network using a large dataset containing authentic and manipulated face images \cite{2019faceforensicspp, mesonet, PS2019detecting, new_cnn_Bayar:2016}. In \cite{2019faceforensicspp}, a large scale dataset called ``FaceForensics++'' is released to address the problem of face forensics. The dataset contains 5,000 videos generated from 4 popular face manipulation methods, Deefakes, FaceSwap, Face2Face and NeuralTextures, which provides rich data to train models as well as a standard benchmark for evaluation.

Most methods for face forensics cast the problem as a classification problem, in which given an image the model is expected to determine whether it is a real face or a manipulated face. Using deep networks has been proved effective in dealing with such a classification problem \cite{2019faceforensicspp, PS2019detecting}. In \cite{2019faceforensicspp}, a modified Xception network \cite{Chollet_2017_CVPR} is trained on ``FaceForensics++'' dataset and achieves remarkable results, accuracy of 99.26 on the raw data. In \cite{mesonet}, a compact network also achieves comparable performance. However, one question is raised: ``\textit{Is the problem well-defined?}'' or ``\textit{Is it a good definition of the problem?}'' In Figure \ref{fig:cls}, the second image shows the activation map of a classification model revealing the high response area for the fake face on the left. It is obviously that the activation map is not actually consistent with the ground-truth, which suggests that the features used to distinguish the fake images might have weak correlation to the real manipulated regions.

The example implies one of the limitations of a classification network that it can only produce a global scalar value representing the confidence of being fake but can not reflect the degree of how the image is manipulated. It would be more beneficial to have a pixel-level output that accurately reflects the manipulated pixels, as shown in the third image of Figure \ref{fig:cls}. Therefore, It would be more natural to formulate the problem of face forensics as a semantic segmentation task so that the model is forced to learn discriminative features to localize manipulated regions.

In this paper, we analyze the problem of face forensics from a pixel-level perspective using segmentation methods to complement the existing classification methods for face forensics. There are some questions that are still under investigated such as: 1) \textit{By nature, whether face forensics is a classification or segmentation problem?} 2) \textit{What is the most suitable network architecture for this problem?} 3) \textit{Should we adopt shallow or deep networks?} 4) \textit{Should we train the model from scratch or initialize it using general vision features.} We conduct experiments to try to answer these questions. By evaluating various architectures, we compare the performance of the segmentation networks and their counterpart classification networks from different aspects. We hope to provide more insight to the problem and establish a new baseline for the benchmark.

Our contributions are three folds:

\begin{itemize}
	\item We conduct a pixel-level analysis to the problem of face forensics by using segmentation methods to be complementary to the existing classification methods.
	
	\item By redefining the problem to be a pixel-level task, we evaluate various architectures and create a strong new baseline for the problem.

	\item By performing different ablation studies, we analyze what makes an effective and efficient anti-fake model, which, we hope, can shed some light on the field of research.  
\end{itemize}

\section{Related Work}

 We cover the most important related papers in the following paragraphs.

\subsection{Digital Face Manipulation}

A comprehensive state-of-the-art report of digital face manipulation can be found in \cite{2018face_star}. Current facial manipulation methods can be separated into four categories: image-based approach, Audio-based approach, computer-graphics-based approach as well as learning based approach.

State of the Arts image-based approaches such as Video Rewrite \cite{Video_Rewrite_Bregler}, Video Face Replacement \cite{2dVideo_face_replacement}, Bringing Portraits to Life \cite{elor2017bringingPortraits} and Deep Video Portraits \cite{kim2018DeepVideo}. These methods employ 2D warps to deform the image to match the expressions of a source actor. “Synthesizing Obama” \cite{Synthesize} learned the mapping between audio and lip motions.

State-of-the-arts computer-graphics-based approaches such as Video Face Replacement \cite{3dVideo_face_replacement}, VDub \cite{Vdub} an Face2Face \cite{Thies_2016_CVPR}. These methods usually reconstruct 3D models using blendshapes or other mesh editing process, based on high-quality 3D face capturing techniques as well as precise and rapid tracking techniques.

 Recently, generative adversarial networks (GANs) are used to apply different facial attributes such as Aging \cite{face_age}, viewpoints \cite{face_rotation_Huang_2017_ICCV}, skin color \cite{attribute_cyclegan_Lu_2018_ECCV}, smiling \cite{Feature_Interpolation2017deep}, or other essential computer graphic renderings \cite{kim2018DeepVideo}, which are implemented as an image-to-image translation, applying a patch-based GAN-loss.

\subsection{Face forensics}

Face forensics aims to ensure authenticity and origin of the face. Face forensics identify computer generated characters from computer graphics faces \cite{mesonet}, print-scanned morphed faces \cite{print_scanned_morphed}, face splicing \cite{Illumination_Color, fight_fake_2018_ECCV}, face swapping \cite{two_stream_tamper, mesonet}, and face reenactment \cite{mesonet, expressions_variation}. Specific artifacts arising from the synthesis process such as color , texture \cite{Illumination_Color} or eye blinking \cite{eye_blinking} can also be exploited. Learning-based approach propose a deep network trained to capture the subtle inconsistencies arising from low-level and/or high level features\cite{mesonet, two_stream_tamper}. Particularly, \cite{rnn_tamper} uses a convolutional neural network to extract frame-level features,which are then used to train a recurrent neural network (RNN) that learns to classify if a video has been subject to manipulation or not. These approaches show impressive results, but can not precisely locate the manipulated area.

\begin{figure*}[t]
	\begin{center}
		\includegraphics[width=0.9\linewidth]{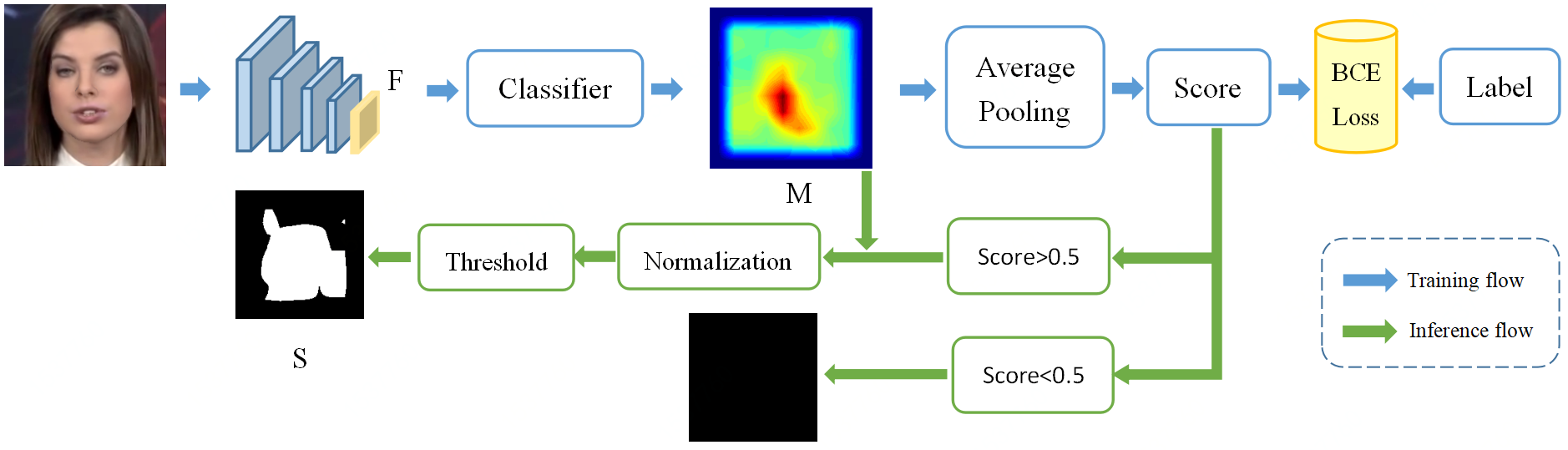}
	\end{center}
	\caption{Pipeline of classification task. Different colors of arrows indicate different stages. Blue is for the training stage, Green is for the inference stage. When the classification score is above 0.5, it is classified as a fake image and is further processed to get the manipulated regions. When the score is below 0.5, indicating a real image, an all-zero mask is produced.}
	\label{fig:cls}
\end{figure*}

\subsection{Pixel-level task}

Instead of a rough prediction in global image-level view, there are many works towards to provide a local or pixel-level prediction, such as Unet \cite{unet_RFB15a}, fully convolutional network (FCN)\cite{FCN_CVPR2015}, Deeplab.for semantic segmentation. As for image generation, pix2pix \cite{pix2pix2016} realize the pixel-level transformation between different domains. There are lots of application concerning face parsing\cite{DBN2012CVPR}, pose parsing or scene segmentation. 

As for face forensics, the mainstream methods are based on global classification at present, we drive the segmentation motivation of face manipulation to predict the region of local manipulation area. The face is often occluded by objects, but the face in the database \cite{2019faceforensicspp} is generally unobstructed, so it can be trained directly.

\section{Problem Setting}
In this section, we first introduce the problem settings and methodologies for both the classification task and the segmentation task. Then we present an overview of the architectures used for evaluation.
\subsection{Classification Task}
\label{sec:cls}

We first revisit the classification task. Formally Let $\mathbf{x} \in \mathbb{R}^{H\times W\times3}$ represent an image containing either an real or a tampered face, and $l \in \{0, 1\}$ represent the label associated to it. We learn a mapping function $f(\cdot): \mathbb{R}^{H\times W\times3} \rightarrow \{0, 1\}$ to predict the authenticity of a face image. Given a dataset $\{(\mathbf{x}^t, l^t)\}^T_{t=1}$ containing T images, the network is trained by the following BCE (Binary Cross Entropy) loss:

\begin{equation}
\mathcal{L}_{cls}=-\frac{1}{T}\sum_{t=1}^{T}(1-l^{t})\log (p^t)+l^t\log (p^t)
\end{equation}

where $p^t$ is the output of the network for $t$th sample.

Since a classification network can only map an image to a scalar indicating the probability of an image being tampered, 
It is unclear whether the model has learned useful features to localize the manipulated regions.
There are some interpretation and visualization works trying to reveal more information from a classification network by investigating the activated regions on featuremaps. \cite{Weakly_Supervised_Object_Localization, Mid_level_Image_Representations, Zhou_2016_CVPR, grad_cam_2017_ICCV} We adopt the most representative method, CAM (Class Activation Map), to help visualize what the model has learned. 

CAM requires the network has an average pooling layer before the classifier, which collapses the output of the last convolution layer to a single vector. Suppose the featuremaps from the last convolution layer is $\mathbf{F} \in \mathbb{R}^{ H_f \times W_f \times K}$; the classifier has weight $\mathbf{w} \in \mathbb{R}^{ K \times 1}$; the activation map $\mathbf{M} \in \mathbb{R}^{ H_f \times W_f}$ of a tampered face is calculated as: 

\begin{equation} \label{eq:cam}
M_{ij}=\underset{k}{\sum}F_{ijk}\cdot w_{k}
\end{equation}

where $M_{ij}$, $F_{ijk}$ and $w_k$ are entries of $\mathbf{M}$, $\mathbf{F}$ and $\mathbf{w}$ respectively.

What Equation \ref{eq:cam} does is actually apply the classifier directly to the featuremaps $\mathbf{F}$, which performs classification on each spatial location. For simplicity, we modify the original CAM setting by switching the average pooling layer and the classifier. As shown in Figure \ref{fig:cls}, the activation map $\mathbf{M}$ can be viewed as a dense prediction output for the image and the classification score is actually produced by averaging the activation map to a scalar.

In order to convert $\mathbf{M}$ to a pixel-level mask, we need to further normalize it to the range of 0 to 1 and quantize it using a threshold. The normalization is operated as:

\begin{equation}
\tilde{\mathbf{M}} = \frac{\mathbf{M} - \min (\mathbf{M})}{\max[\mathbf{M} - \min (\mathbf{M})]}
\end{equation}

The final pixel-level prediction is generated as:

\begin{equation}
\mathbf{S} = \mathbb{I} \{\tilde{\mathbf{M}}\geq \tau_1\}
\end{equation}

where $\mathbb{I}(\cdot)$ is a indicator function and $\tau_1$ a threshold. 

Now we have a pixel-level output that highlights the manipulated regions. Using these outputs makes it easier to investigate and analyze how well the classification model is able to learn discriminative and high-quality features on a pixel-level. Details and analysis are described in Section \ref{sec:exp}.

\begin{figure}[h]

\begin{center}
   \includegraphics[width=1\linewidth]{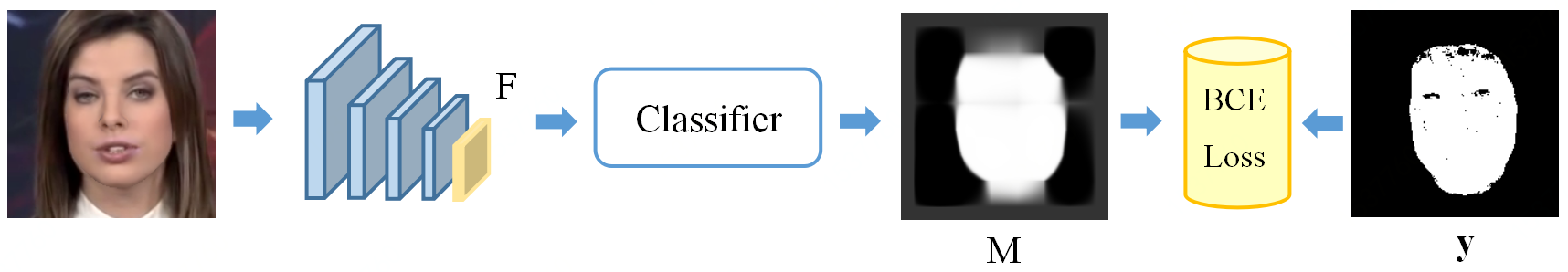}
\end{center}
   \caption{Pipeline of segmentation task. The network predicts a pixel-level output and is supervised directly by a pixel-level mask. }
\label{fig:seg}
\end{figure}
\subsection{Segmentation Task}

\begin{figure}[t]

\begin{center}
	\setlength{\tabcolsep}{0.3em}
	\begin{tabular}{ccccc}
		\includegraphics[width=0.18\linewidth]{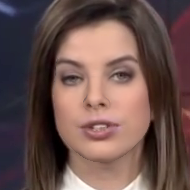} &
		\includegraphics[width=0.18\linewidth]{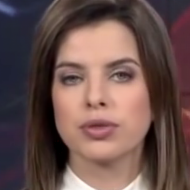} &
		\includegraphics[width=0.18\linewidth]{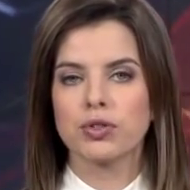} &
		\includegraphics[width=0.18\linewidth]{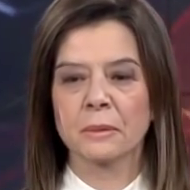} &
	   \includegraphics[width=0.18\linewidth]{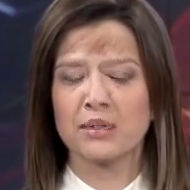} \\	
		\includegraphics[width=0.18\linewidth]{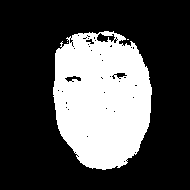} &
		\includegraphics[width=0.18\linewidth]{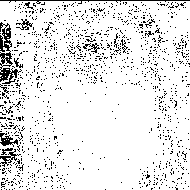} &
		\includegraphics[width=0.18\linewidth]{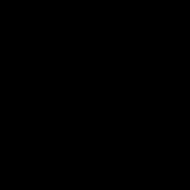} &
		\includegraphics[width=0.18\linewidth]{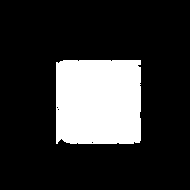} &
		\includegraphics[width=0.18\linewidth]{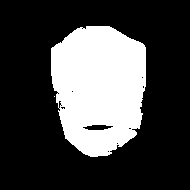} \\			
		F2F & NT & P & DF & FS
	\end{tabular}

\end{center}
   \caption{Illustration of example images and the corresponding masks for the ``FaceForensics++'' dataset. (P: Pristine, DF: DeepFakes, F2F: Face2Face, FS: FaceSwap, NT: NeuralTextures)}
\label{fig:mask}
\end{figure}

A classification network has limited capability to localize manipulated regions with a pixel-level manner because it is supervised only by a global label. Segmentation extends the task to a dense classification problem by assigning a label to each pixel of an image. The model is then forced to learn discriminative features to determine the authenticity of each pixel. Formally, the supervision for an image is defined as a mask $\mathbf{y}\in \{0, 1\}^{ H \times W}$ instead of a single label and the loss is imposed on each pixel:

\begin{equation}
\mathcal{L}_{seg}=-\frac{1}{T}\sum_{t=1}^{T}\sum_{i=1}^{H}\sum_{j=1}^{W}(1-y^{t}_{ij})\log (p^t_{ij})+y^t_{ij}\log (p^t_{ij})
\end{equation}

where $y^t_{ij}$ and $p^t_{ij}$ are the label and the prediction respectively for $t$th sample at position $(i,j)$.

Since a segmentation task requires pixel-level mask as supervision, annotation of the data is usually time-consuming. For example, as mentioned in \cite{Cordts2016Cityscapes}, a high-resolution street view image for semantic segmentation requires around 1.5 hours for labelling. Fortunately for the face forensics task, the mask can be easily calculated by checking the pixel difference between the original image and the forged image without any extra annotation cost. Figure \ref{fig:mask} shows some training images from ``FaceForensics++'' dataset as well as their corresponding mask indicating the manipulated area. 

A classification network can be easily converted to a FCN (Fully Convolutional Network) \cite{FCN_CVPR2015} where the fully connected layers are replaced by convolutional layers. The pipeline for training a segmentation network is illustrated in Figure \ref{fig:seg}. Compared with the classification task in Figure \ref{fig:cls}, the main difference is that the average pooling is dropped and the BCE loss is directly applied to each pixel. The pixel-level prediction can be directly obtained from the trained model.

A segmentation model can be also evaluated from a global classification perspective by aggregating the dense prediction:

\begin{equation} \label{eq:agg}
\hat{l}=\mathbb{I}\{\frac{1}{HW}\sum_{i=1}^{H}\sum_{j=1}^{W}\hat{y}_{ij}\geq\tau_2\}
\end{equation}

where $\hat{y}_{ij}$ represents the prediction at position $(i,j)$ and $\tau_2$ is the threshold.

In this way, we are able to make fair comparison between a segmentation network and its counterpart classification network under classification metrics. With extensive experiments in Section \ref{sec:exp}, we show the superiority of the segmentation networks for the face forensics task.

\subsection{Architectures}

In order to conduct deep analysis on the classification and segmentation task, we choose several representative architectures to evaluate the effectiveness on the problem of face forensics. 

\noindent \textbf{Xception} \cite{xception_2017_CVPR} is a deep network architecture constructed by a series of modified inception modules \cite{inception_CVPR} where the depthwise separable convolution is used. There are totally 36 convolutional layers involved to form the feature extraction base of the network. The architecture is adopted in \cite{2019faceforensicspp} for the classification task of face forensics.

\noindent \textbf{MesoInception-4} \cite{mesonet} is a compact and light-weight network to address the problem of face forensics. It consists of two inception modules followed by two classic convolutional layers with maxpooling layers. We replace the all the operations after the last batchnorm layer with a single convolutional layer as the classifier. 

\noindent \textbf{UNet} \cite{unet_RFB15a} is an effective and popular architecture for pixel-level tasks such as segmentation and pixel-to-pixel translation \cite{pix2pix2016}. A Unet is basically defined by an encoder, consisting of convolutional layers and downsampling operations, and a decoder, consisting of convolutional layers and upsampling operations. There are skip connections between the encoder and the decoder to enable passing information from low-level features. We choose two variants of UNet with different downsampling times in the encoder. \textbf{UNet8x} and \textbf{UNet4x} are downsampled by 8 and 4 times respectively. 

\noindent \textbf{VGG16} \cite{vgg} is a classic deep network for recognition tasks, which consists of 16 convolutional layers. Since we found the full network fails to converge on face forensics tasks, we only use two shallow versions \textbf{VGG8} and \textbf{VGG5}, containing the first 7 and 4 feature layers of vgg respectively and a classifier. 

\noindent \textbf{FN3} is a 3-layer network we design to explore the potential of shallow networks. This architecture only contains two ``Conv-BN-ReLU'' blocks, and a $3\times3$ convolutional layer as the classifier. The first two convolutional layers are with kernel size 7 and stride 2. It is interesting that this minimum structure works surprisingly good, even outperforming most of those deep architectures. Please refer to Section \ref{sec:exp} for more details.



\section{Experiments}
\label{sec:exp}
\subsection{Experiment Setup}

\begin{table}[h]
	\renewcommand{\arraystretch}{1.5}
	\begin{center}
		\scalebox{0.8}{
			\begin{tabular}{|c|ccccc|c|}
				\hline 
				& DF & F2F & FS & NT & P & Avg\tabularnewline
				\hline 
				\hline 
				Xception-cls & \textbf{99.16} & \textbf{98.35} & \textbf{98.88} & \textbf{99.09} & \textbf{99.18} & \textbf{98.93}\tabularnewline
				
				Mesonet-cls & 93.33 & 77.01 & 26.77 & 92.05 & 88.99 & 75.63\tabularnewline
				
				UNet8x-cls &  56.57&  33.4&  22.96&  47.21&  92.55& 50.5\tabularnewline
				
				UNet4x-cls &  66.9&  45.45&  37.48&  55.42&  98.98& 60.8\tabularnewline
				
				VGG7-cls &  41.69&  73.37&  67.78&  38.81&  76.45& 59.6\tabularnewline
				
				VGG4-cls &  56.02&  84.92&  90.72&  40.33&  70.66& 68.53\tabularnewline
				
				Conv3-cls & 94.35 &93.28  &81.13  &94.26  &61.43  &84.89 \tabularnewline
				\hline 
				Xception-seg & 96.45 & 97.98 & 99.02 & 98.39 & 99.92 & 98.35\tabularnewline
				
				Mesonet-seg & 68.86 & 79.58 & 89.77 & 96.92 & 59.56 & 78.94\tabularnewline
				
				UNet8x-seg & \textbf{99.08} & \textbf{98.74} & 97.17 & \textbf{99.42} & 66.65 & 92.21\tabularnewline
				
				UNet4x-seg & 98.61 & 97.32 & \textbf{99.05} & 96.53 & 97.01 & 97.70\tabularnewline
				
				VGG7-seg &98.41  &98.34  &99.05  &99.01  &99.33  &98.83 \tabularnewline
				
				VGG4-seg &98.24  &98.29  &99.03  &99.01  &99.99  &98.91 \tabularnewline
				
				Conv3-seg & 98.16 & 98.32 & 99.03 & 99.06 & \textbf{99.99} & \textbf{98.91}\tabularnewline
				\hline
		\end{tabular}}
	\end{center}
	\caption{Classification accuracy on different manipulation methods. (P: Pristine, DF: DeepFakes, F2F: Face2Face, FS: FaceSwap, NT: NeuralTextures)}
	\label{tab:cls}
\end{table}

\begin{table*} [t]
	
\renewcommand{\arraystretch}{1.5}
	\begin{center}
		\scalebox{0.70}{
			\begin{tabular}{|c|ccccc|c|ccccc|c|ccccc|c|}
				\hline 
				& \multicolumn{6}{c|}{mIoU} & \multicolumn{6}{c|}{Bg-IoU} & \multicolumn{6}{c|}{Fg-IoU}\tabularnewline
				\hline
				& DF & F2F & FS & NT & P & Avg & DF & F2F & FS &NT & P & Avg & DF & F2F & FS & NT & P  & Avg\tabularnewline
				\hline 
				\hline 
				Xception-seg & 89.32 & 88.18 & 87.7 & 62.81 & \textbf{99.95} & 85.59 & 95.95 & 93.79 & 94.19 & 41.94 & \textbf{99.95} & 85.16 & 82.7 & 82.56 & 81.21 & 83.67 & - & 82.54\tabularnewline
				Mesonet-seg & 56.58 & 51.14 & 54.52 & 40.23 & 90.2 & 58.53 & 78.96 & 71.06 & 74.68 & 22.02 & 90.2 & 67.38 & 34.19 & 31.21 & 34.35 & 58.44 & - & 39.55\tabularnewline
				UNet8x-seg & 87.83 & 86.97 & 85.02 & 50.51 & 86.02 & 79.27 & 94.7 & 92.32 & 91.82 & 28.27 & 86.02 & 78.63 & 80.96 & 81.62 & 78.22 & 72.75 & - & 78.39\tabularnewline
				UNet4x-seg & 89.12 & 89.43 & 86.29 & 51.46 & 96.09 & 82.48 & 95.41 & 93.89 & 92.59 & 30.68 & 96.09 & 81.73 & 82.82 & 84.96 & 79.99 & 72.25 & - & 80.00\tabularnewline
				VGG8-seg & 94.68 & 95.21 & 94.33 & \textbf{76.04} & 99.31 & 91.91 & 97.87 & 97.34 & 97.19 & \textbf{59.42} & 99.31 & 90.23 & 91.48 & 93.07 & 91.47 & \textbf{92.67} & - & 92.17\tabularnewline
				VGG5-seg & \textbf{95.78} & \textbf{96.21} & \textbf{94.81} & 75.6 & 99.86 & \textbf{92.45} & \textbf{98.36} & \textbf{97.94} & \textbf{97.51} & 58.97 & 99.86 & \textbf{90.53} & \textbf{93.21} & \textbf{94.48} & \textbf{92.11} & 92.23 & - & \textbf{93.01}\tabularnewline
				FN3-seg & 92.68 & 93.05 & 89.01 & 64.42 & 99.72 & 87.78 & 97.16 & 96.24 & 94.81 & 43.89 & 99.72 & 86.36 & 88.21 & 89.86 & 83.21 & 84.95 & - & 86.56\tabularnewline
				\hline 
				Xception-cls & \textbf{47.6} & \textbf{58.98} & \textbf{56.21} & \textbf{58.83} & \textbf{99.23} & \textbf{64.17} & 59.9 & \textbf{71.9} & \textbf{75.62} & \textbf{23.27} & \textbf{99.23} & \textbf{65.98} & \textbf{35.3} & \textbf{46.06} & \textbf{36.8} & 52.95 & - & \textbf{42.78}\tabularnewline
				Mesonet-cls & 45.96 & 37.14 & 37.48 & 24.78 & 95.46 & 48.16 & \textbf{67.87} & 55.75 & 65.08 & 13.93 & 95.46 & 59.62 & 24.05 & 18.53 & 9.88 & 35.63 & - & 22.02\tabularnewline
				UNet8x-cls & 23 & 33.6 & 34.82 & 29.71 & 86.39 & 41.5 & 28.63 & 49.62 & 53.8 & 11.3 & 86.39 & 45.94 & 17.42 & 17.58 & 15.84 & 48.11 & - & 24.74 \tabularnewline
				UNet4x-cls & 22.3 & 32.95 & 34.38 & 35.14 & 97.59 & 44.47 & 26.25 & 46.11 & 51.92 & 13.99 & 97.59 & 47.17 & 18.35 & 19.79 & 16.83 & 56.29 & - & 27.82 \tabularnewline
				VGG8-cls & 28.73 & 23.45 & 26.12 & 29.84 & 63.66 & 34.36 & 40.91 & 21.65 & 30.72 & 12.1 & 63.66 & 33.81 & 16.54 & 25.25 & 21.51 & 47.57 & - & 27.72 \tabularnewline
				VGG5-cls & 39.18 & 37.92 & 38.85 & 15.73 & 80.39 & 42.41 & 66.81 & 63.83 & 69.56 & 13.51 & 80.39 & 58.82 & 11.56 & 12.01 & 8.14 & 17.96 & - & 12.42\tabularnewline
				FN3-cls & 16.77 & 18.46 & 20.47 & 43.84 & 48.68 & 29.64 & 14.58 & 10.63 & 17.9 & 8.09 & 48.68 & 19.97 & 18.97 & 26.29 & 23.04 & \textbf{79.59} & - & 36.97\tabularnewline
				\hline 
			\end{tabular}}
	\end{center}
	\caption{Segmentation results for different architectures. (P: Pristine, DF: DeepFakes, F2F: Face2Face, FS: FaceSwap, NT: NeuralTextures)}
	\label{tab:seg}
\end{table*}

\noindent \textbf{Dataset:}
FaceForensics++ \cite{2019faceforensicspp} is a large scale face forensics dataset consisting of 5,000 video clips in total. Video sequences are crawled from the internet and a manual screening is adopted to ensure high quality and avoid face occlusion, resulting in 1,000 original videos. Four manipulation methods, Deepfakes, Face2Face, FaceSwap and NeuralTextures, are applied to create forged videos, resulting in 4,000 fake clips. The dataset also provides data with three different compression levels, raw, HQ and LQ. We only focus on the raw quality task because low quality videos usually suffer from strong loss of visual and identity information, which might not cause abuse as those clear ones. \cite{2019faceforensicspp} also suggests a split of 720 videos for training, 140 for validation as well as testing. We follow the same  setting.

\noindent \textbf{Evaluation protocol and metrics:}
In \cite{2019faceforensicspp}, there are two types of training protocols involved, \textit{method specific training} and \textit{mixed training}. The former involves forged data from only one of the manipulation methods. The latter requires training a model with all the real and forged data and the performance is evaluated on each specific method. We only adopt \textit{mixed training} as it poses a more challenging task and real scenario. The evaluation is frame-based, therefore we extract all frames for the training set and partial frames for validation and testing (every 10 frames).

In terms of evaluation metrics, we use classification accuracy for the classification tasks, which represents how many test images are correctly classified. For segmentation tasks, IoU (Intersection over Union) is used to represent the ratio of $\frac{TP}{TP+FP+TN}$, where TP (True Positive), FP (False Positive) and TN (True Negative) are calculated based on pixels. The IoU is calculated for both foreground and background, denoted as Fg-IoU and Bg-IoU. The two IoUs are averaged to get mIoU, the mean IoU.

 \noindent\textbf{Implementation details:} 
 In face forensics, faces are the most important regions. As shown in \cite{2019faceforensicspp}, the model trained with the whole images performs poorly. Therefore, instead of using the whole image, we extract the faces as a pre-processing step using a public face detection tool \cite{facerecognition} and only use the face regions to train the models . In order to include more background information, we enlarge the bounding box to the scale of 2. The segmentation masks are calculated by checking the difference between a manipulated face image and its corresponding original image. For segmentation tasks, the images are randomly cropped by size 256x256 and the same operation is applied to the corresponding mask to get the cropped mask. For classification tasks, it is necessary to include most face regions in the crop. Therefore, the shorter dimension of the image is first resized to 256, then a patch of 256x256 is cropped from the resized image.

 The implementation is based on PyTorch \cite{paszke2017automatic}.
 All the models are trained using the Adam \cite{kingma:adam} optimizer with parameters $\beta_{1}=0.9$ and $\beta_{2}=0.999$. Since the Adam optimizer can adjust the learning rate dynamically, we only set the default learning rate to $10^{-3}$ and do not use any learning rate decay policies. The batchsize is set to 64.

\subsection{Experimental Results}

\noindent \textbf{Classification task:}
Table \ref{tab:cls} shows the classification accuracy of different architectures. The suffixes ``-seg'' and ``-cls'' represent a segmentation model and a classification model respectively. The pixel-wise output is aggregated to a global output according to Equation \ref{eq:agg}. From the scores of  the classification models, Xception-cls reaches the best performance, which is consistent with \cite{2019faceforensicspp}. It can be seen that UNet, as a popular segmentation model for various pixel-level prediction tasks, fails to perform well in the classification task. it is interesting to see that FN3-cls, a minimum structure with only 3 layer works surprisingly good. Although, the performance is lower than Xception, FN3-cls achieves far better performance than other models.  For those segmentation models, it can be easily noticed that they obtain better classification results than their counterpart classification models, which shows the benefit of training models under pixel-level supervision.

\noindent \textbf{Segmentation task:}
Table \ref{tab:seg} shows the segmentation results of different architectures. The classification models are trained using a global image-level label and visualized by the CAM to get a pixel-level output, explained in Section \ref{sec:cls}. For segmentation models, VGG5-seg achieves the best performance in terms of both mIoU, Bg-IoU, and Fg-IoU. Mesonet-seg, as a compact and efficient architecture, does not achieve comparable results, outperformed by other methods by a large margin. We suspect it could be due to the limited capacity of the model.
It is also worth noting that UNet still does not reach promising results as a popular segmentation architecture. On the contrary, the 3-layer network FN3-seg shows better potential, even better than Xception-seg.
For classification models, Xception-cls achieves the best results in most of the scores, which implies that Xception-cls can successfully learn high-quality features to locate manipulated regions even trained with a global image-level label. However, Xception-cls can be hardly compared with its segmentation counterpart that obtains far higher scores. The rest of the classification models all suffer from low scores. Even VGG5-cls, whose segmentation counterpart achieves the best results, is unable to produce plausible predictions without pixel-level supervision.

\begin{figure*}[t]
	\setlength{\abovecaptionskip}{0.cm}
	\setlength{\belowcaptionskip}{-0.1cm}
	\begin{center}
		\setlength{\tabcolsep}{0.3em}
		\begin{tabular}{|ccccccccc|}
			\hline
			\multirow{2}{*}{DF}&\includegraphics[width=0.1\linewidth]{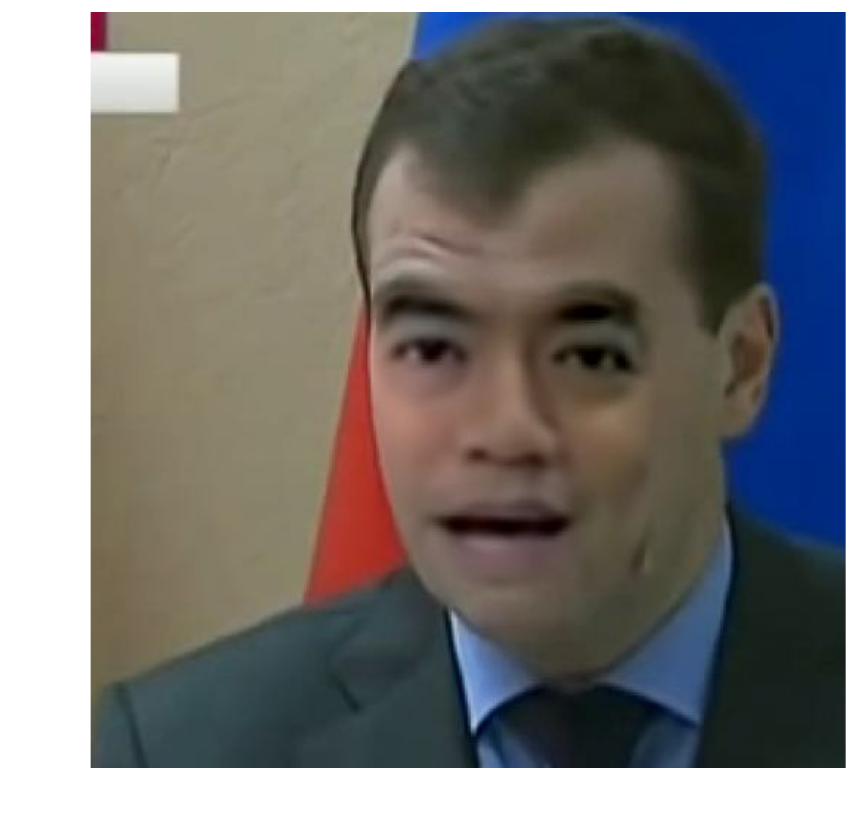}&
			\includegraphics[width=0.1\linewidth]{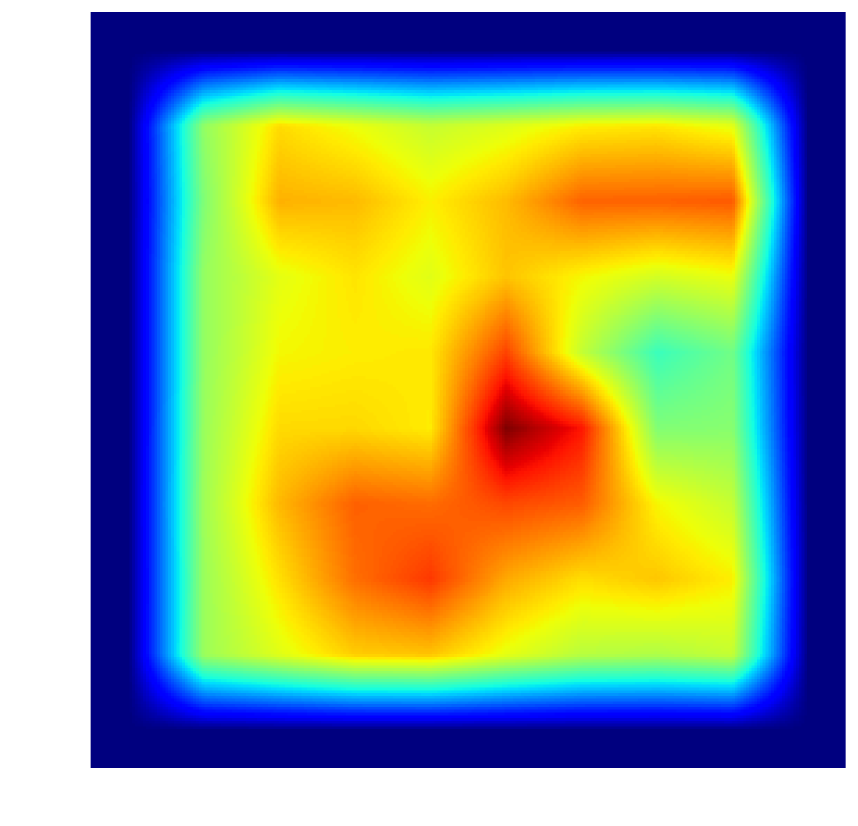}&
			\includegraphics[width=0.1\linewidth]{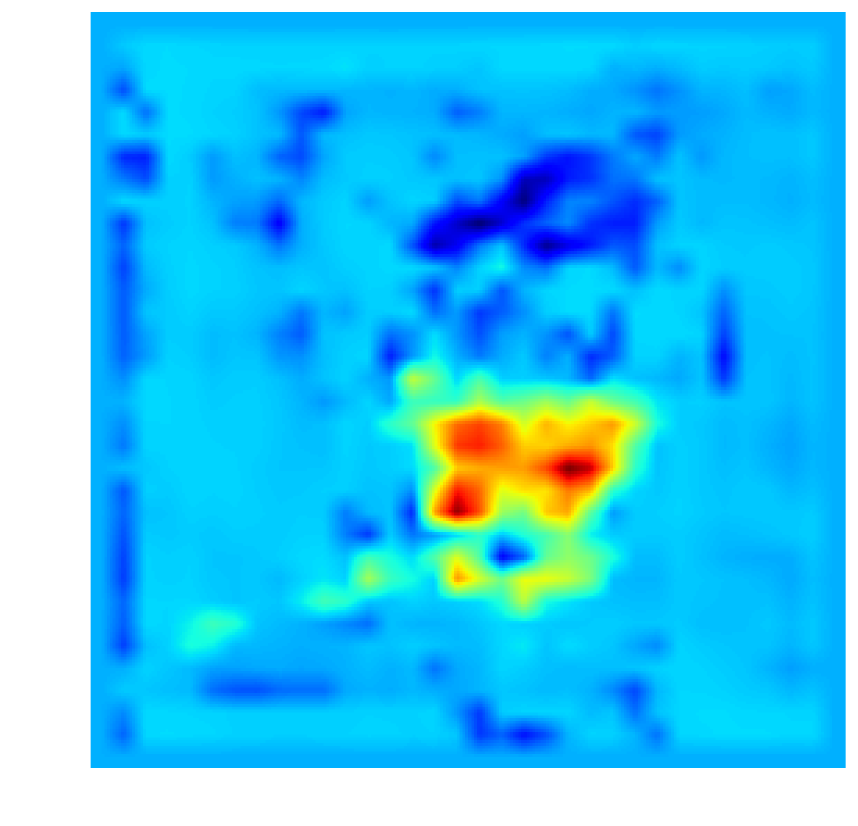}&
			\includegraphics[width=0.1\linewidth]{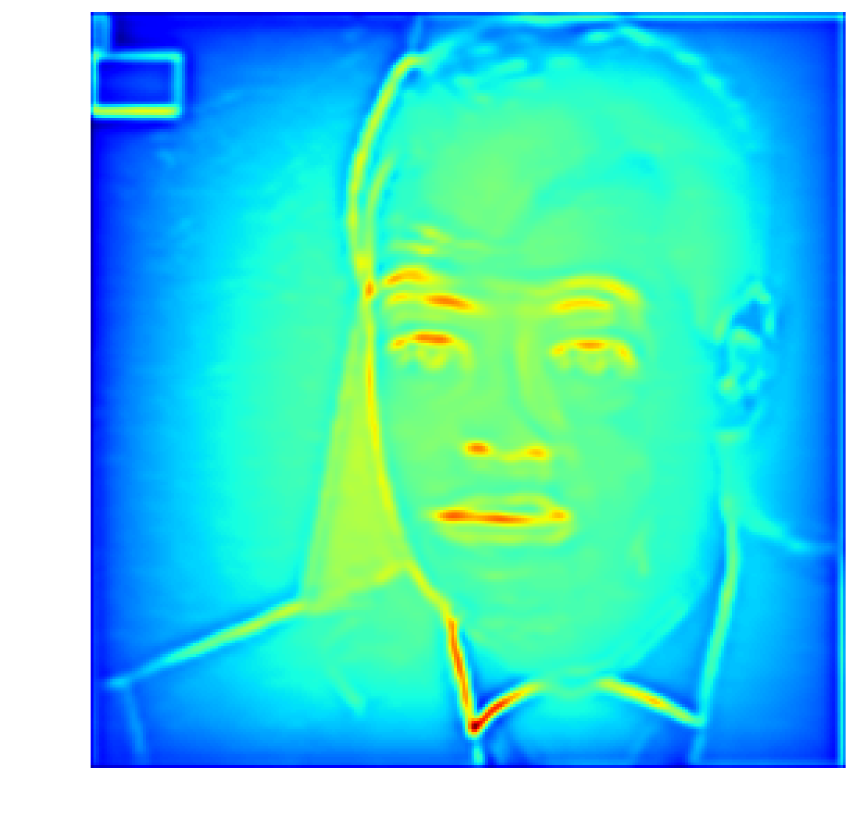}&
			\includegraphics[width=0.1\linewidth]{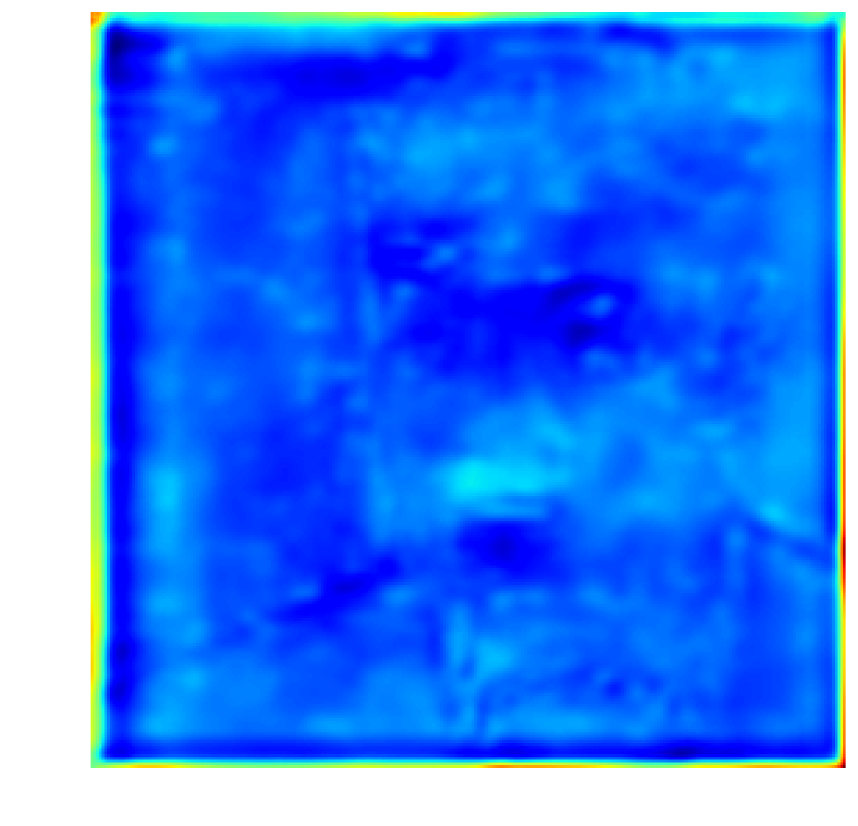}&
			\includegraphics[width=0.1\linewidth]{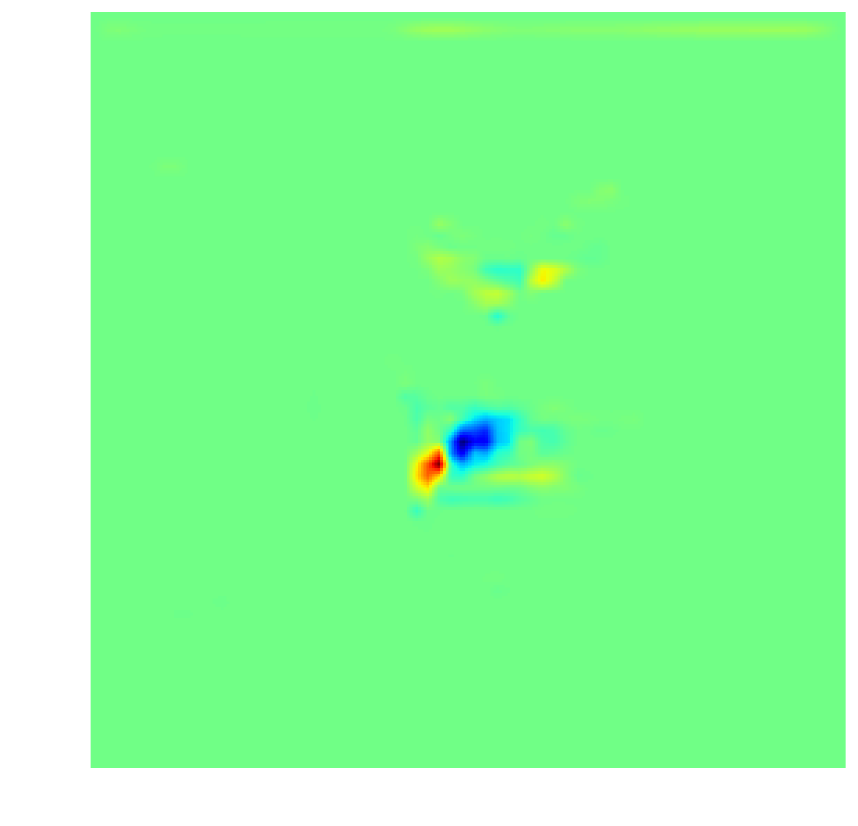}&
			\includegraphics[width=0.1\linewidth]{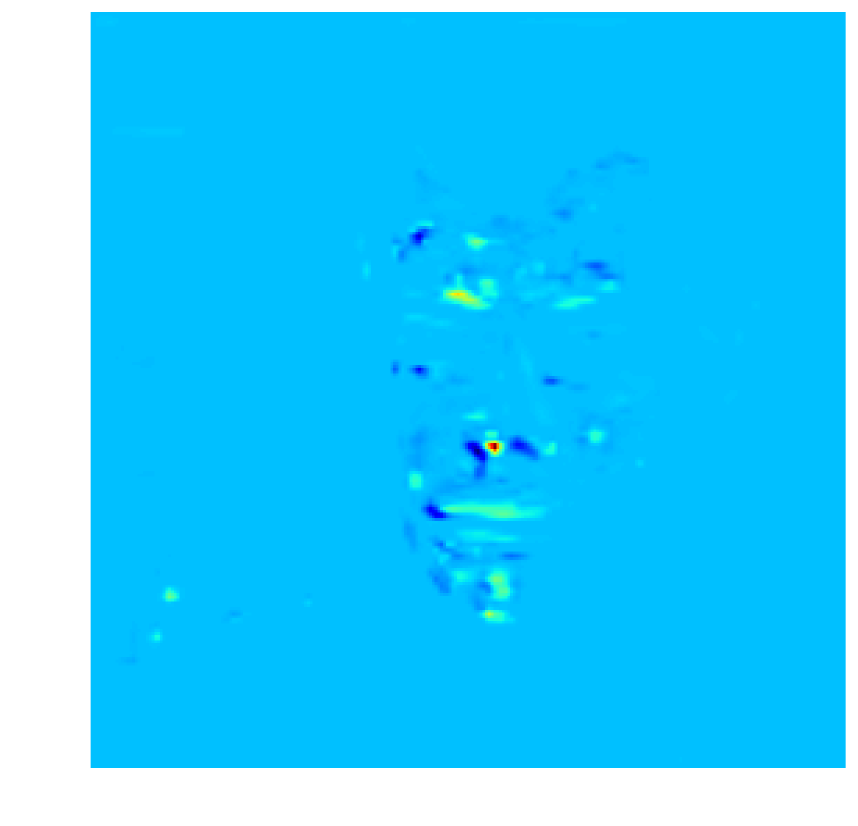}&
			\includegraphics[width=0.1\linewidth]{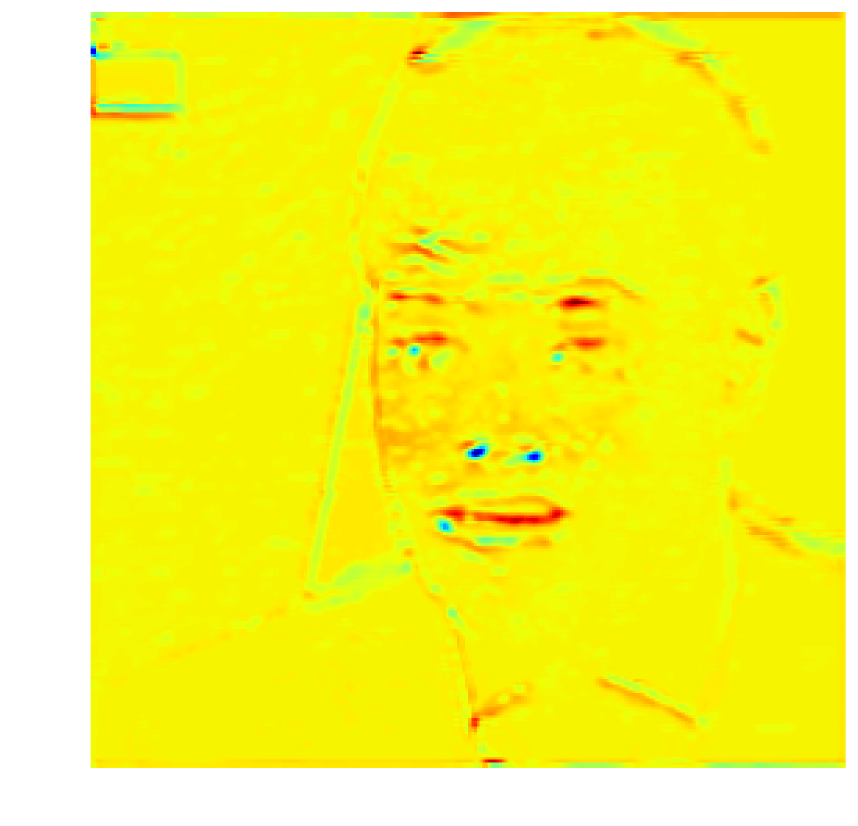}
			\\
			&\includegraphics[width=0.1\linewidth]{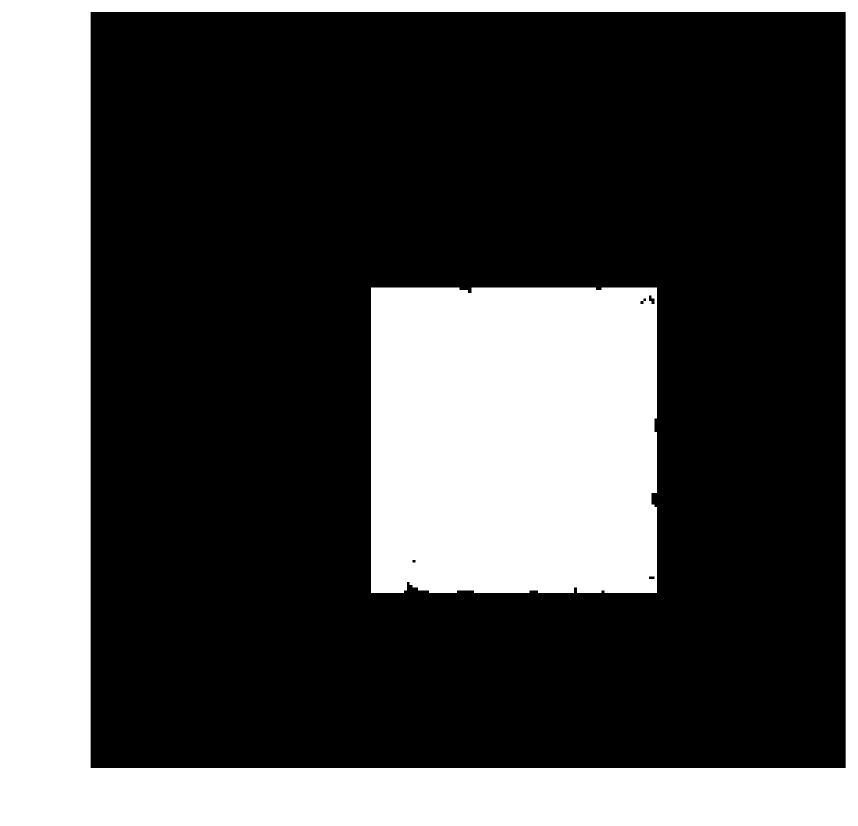}&
			\includegraphics[width=0.1\linewidth]{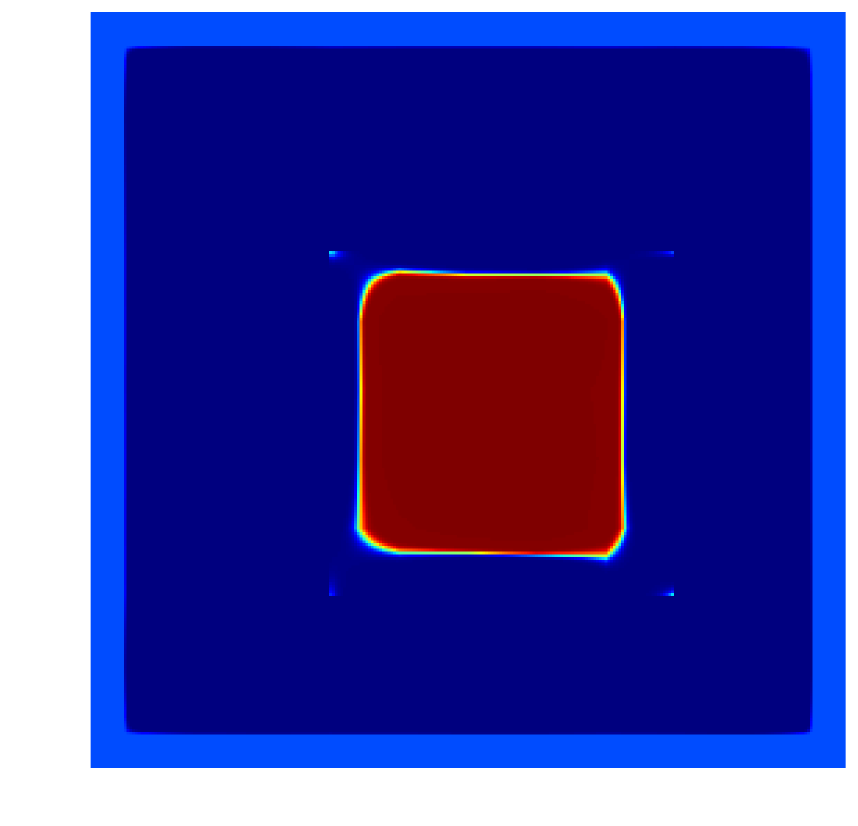}&
			\includegraphics[width=0.1\linewidth]{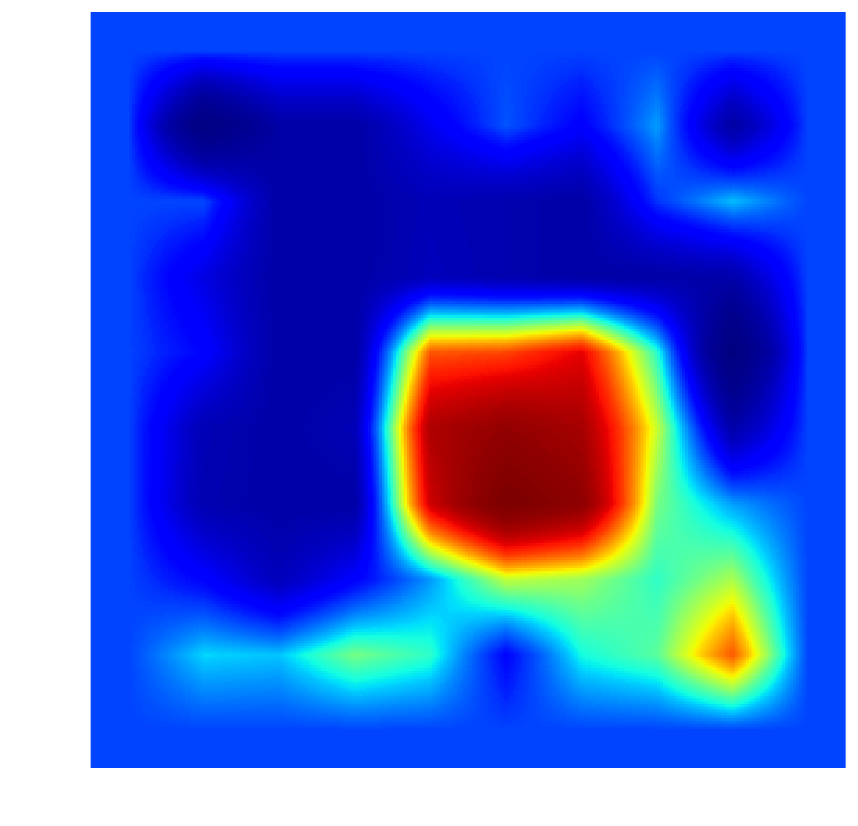}&
			\includegraphics[width=0.1\linewidth]{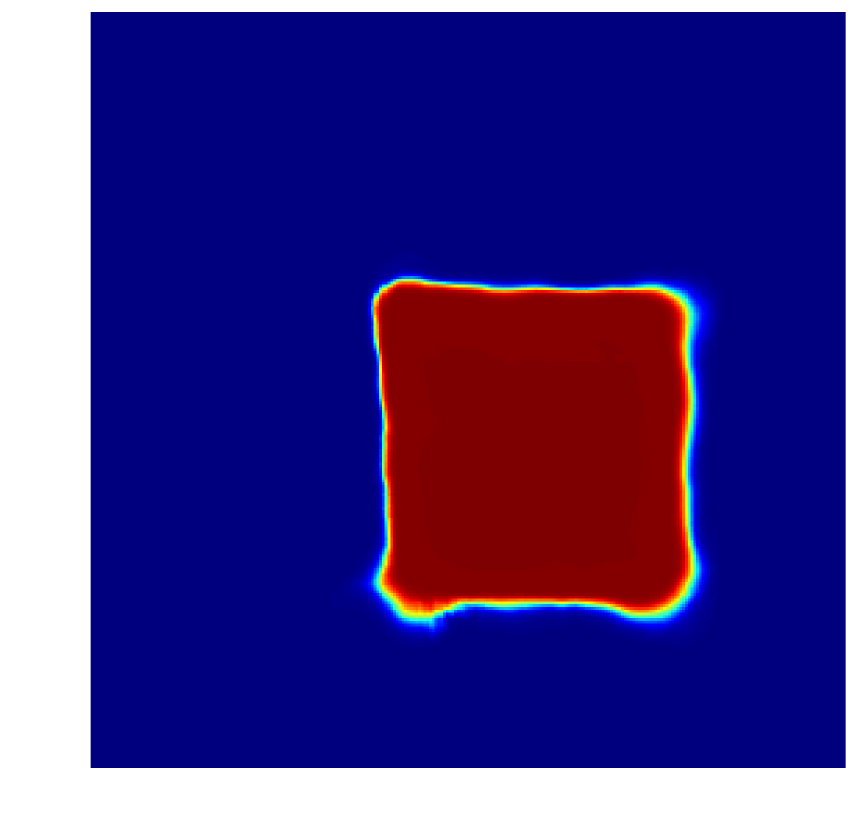}&
			\includegraphics[width=0.1\linewidth]{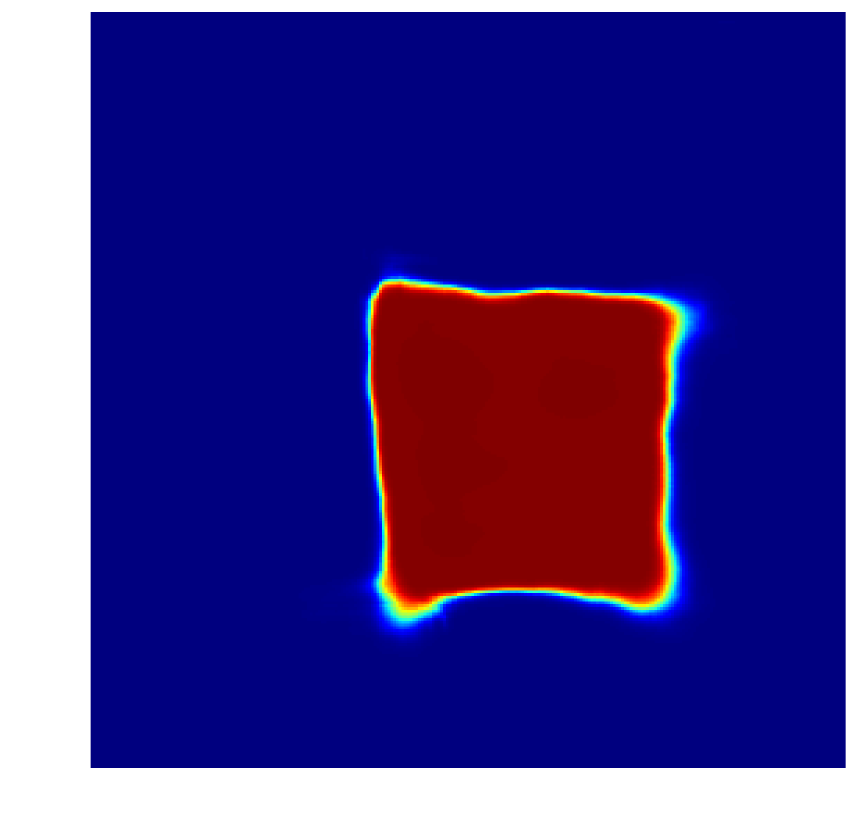}&
			\includegraphics[width=0.1\linewidth]{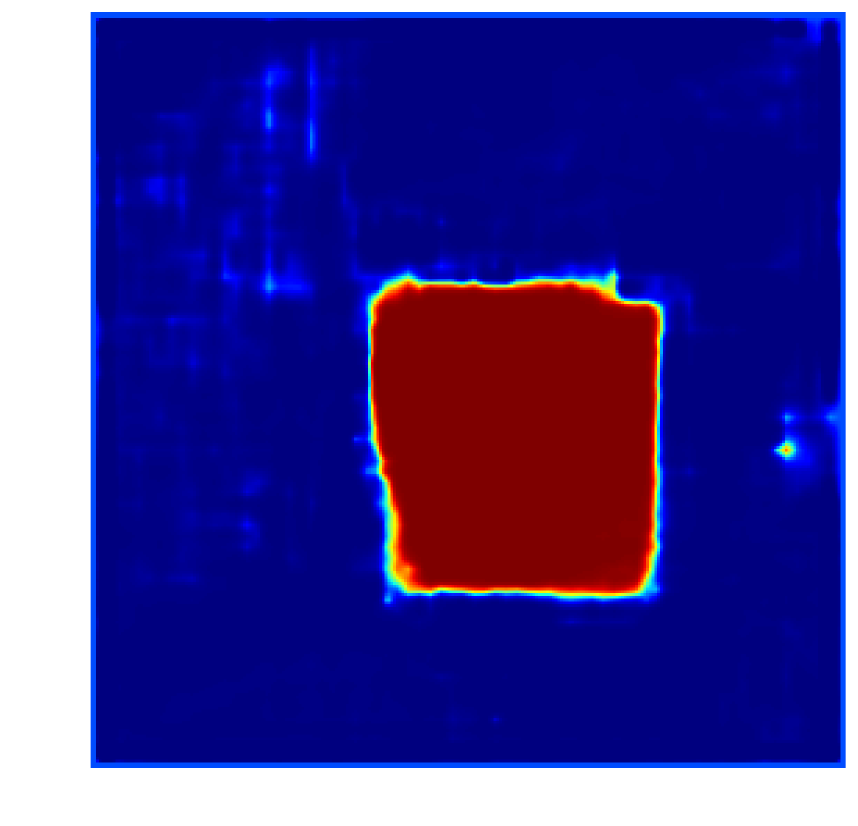}&
			\includegraphics[width=0.1\linewidth]{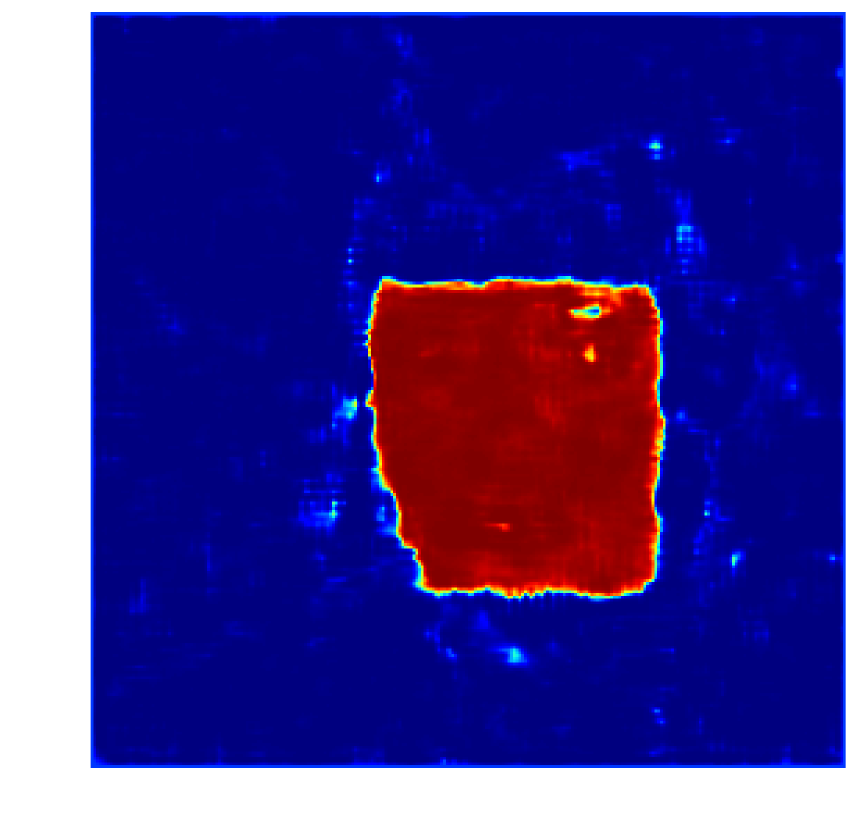}&
			\includegraphics[width=0.1\linewidth]{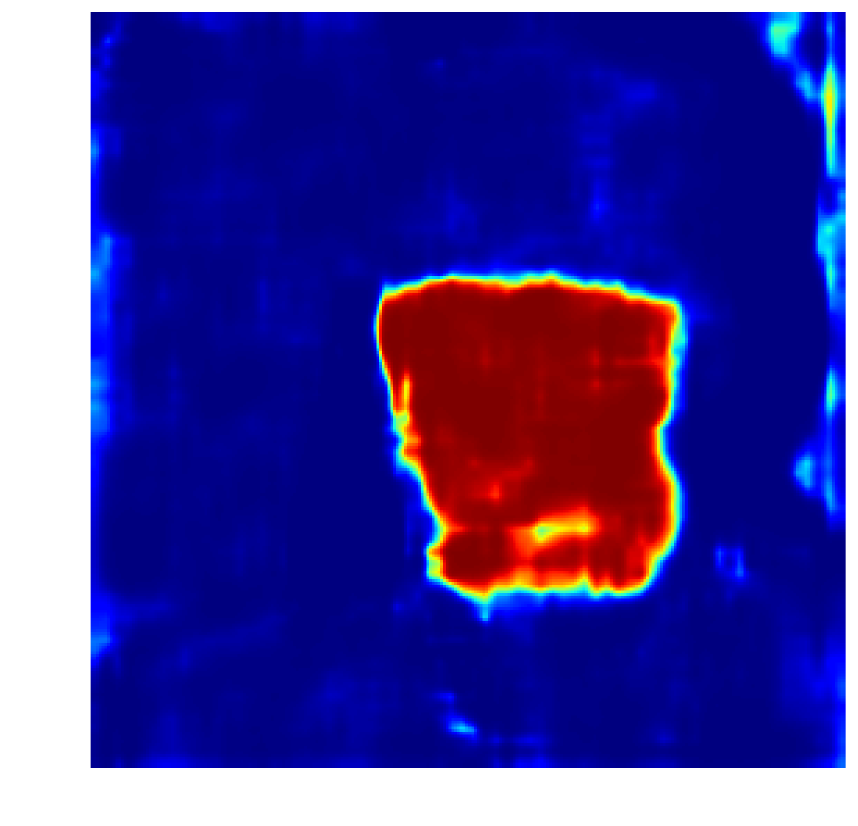}\\
			\hline
			\multirow{2}{*}{F2F}&\includegraphics[width=0.1\linewidth]{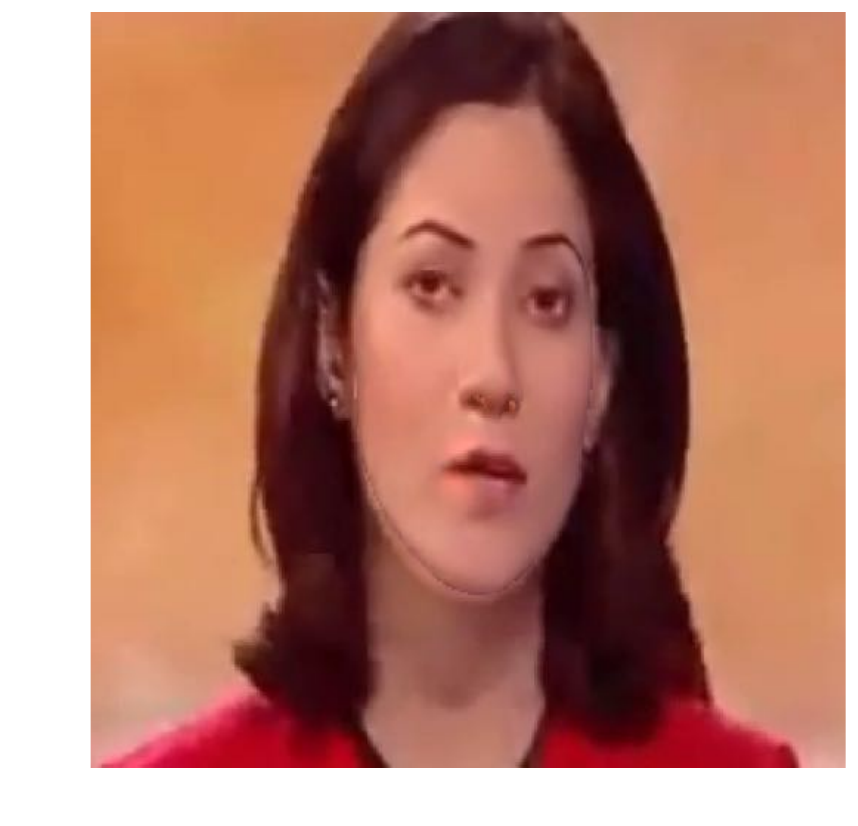}&
			\includegraphics[width=0.1\linewidth]{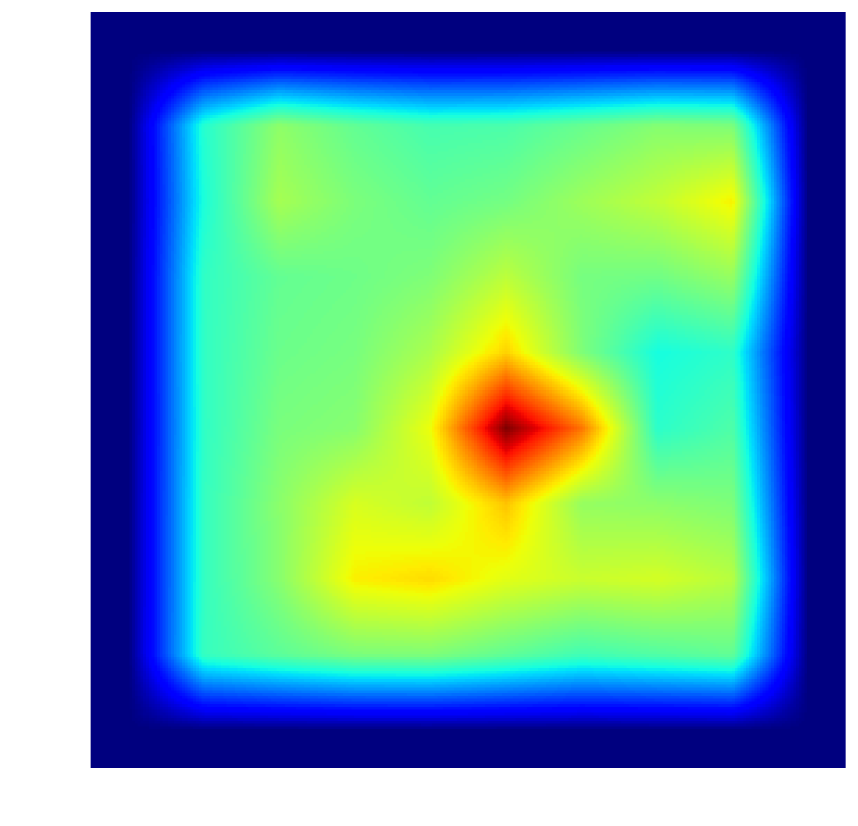}&
			\includegraphics[width=0.1\linewidth]{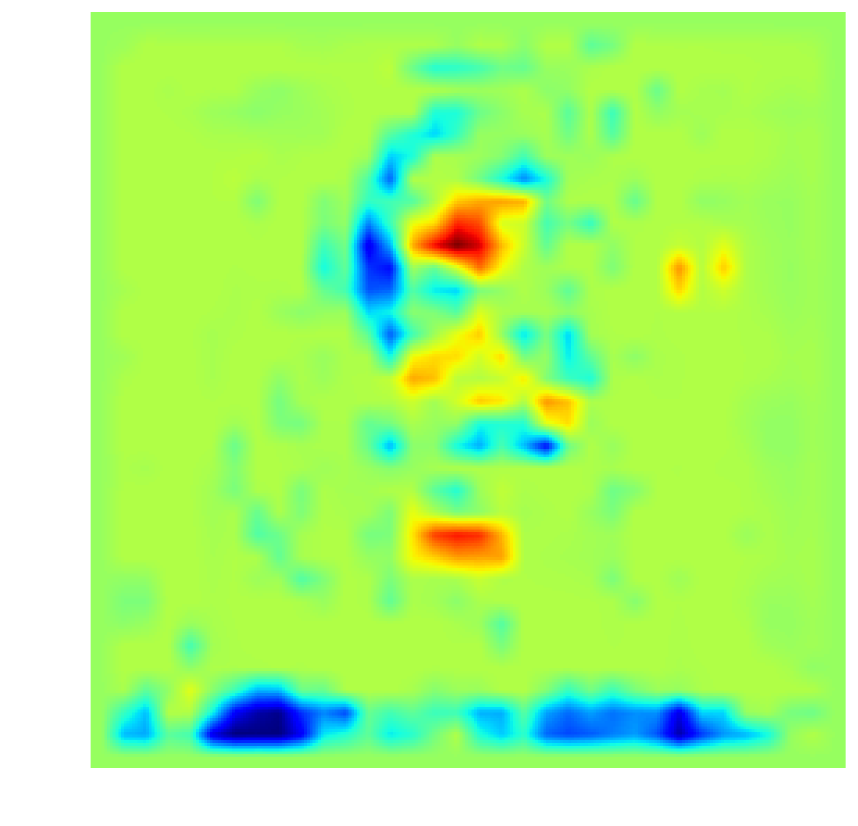}&
			\includegraphics[width=0.1\linewidth]{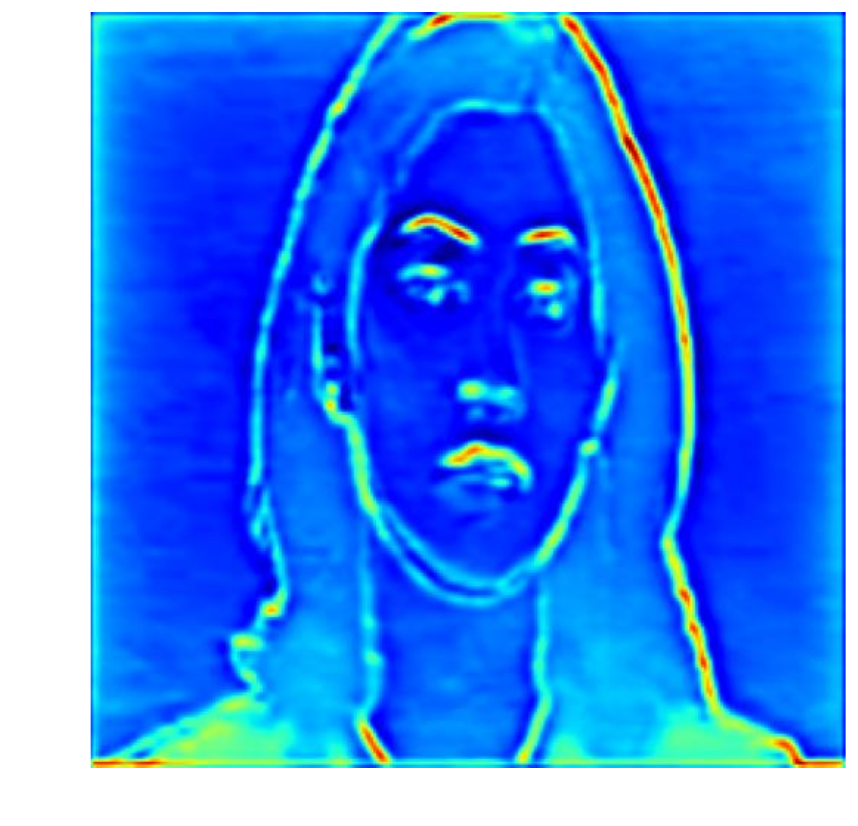}&
			\includegraphics[width=0.1\linewidth]{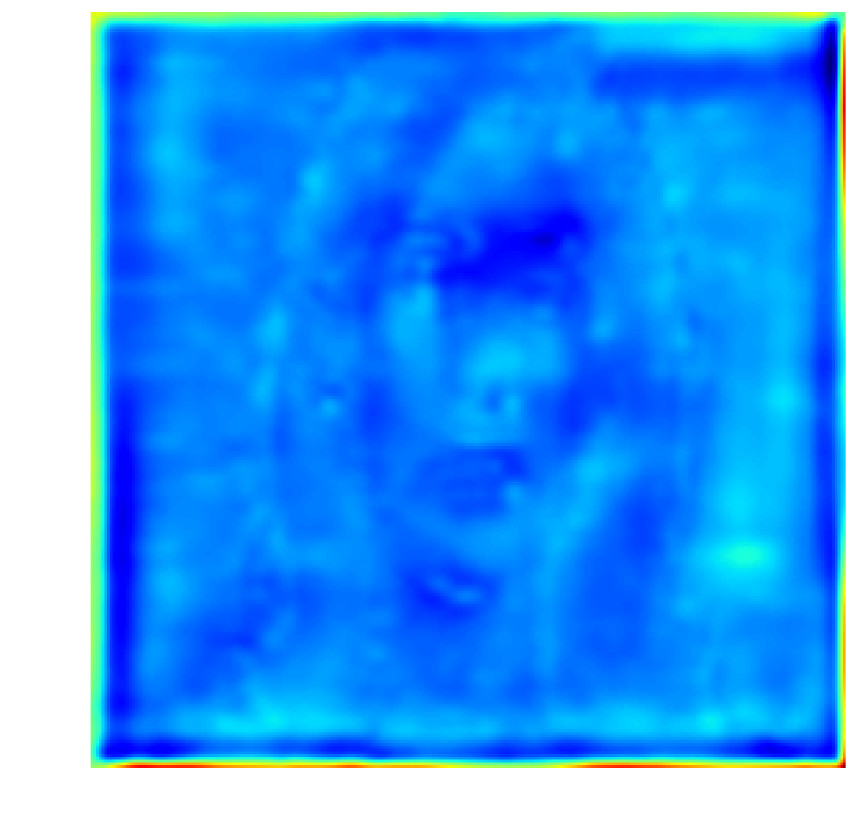}&
			\includegraphics[width=0.1\linewidth]{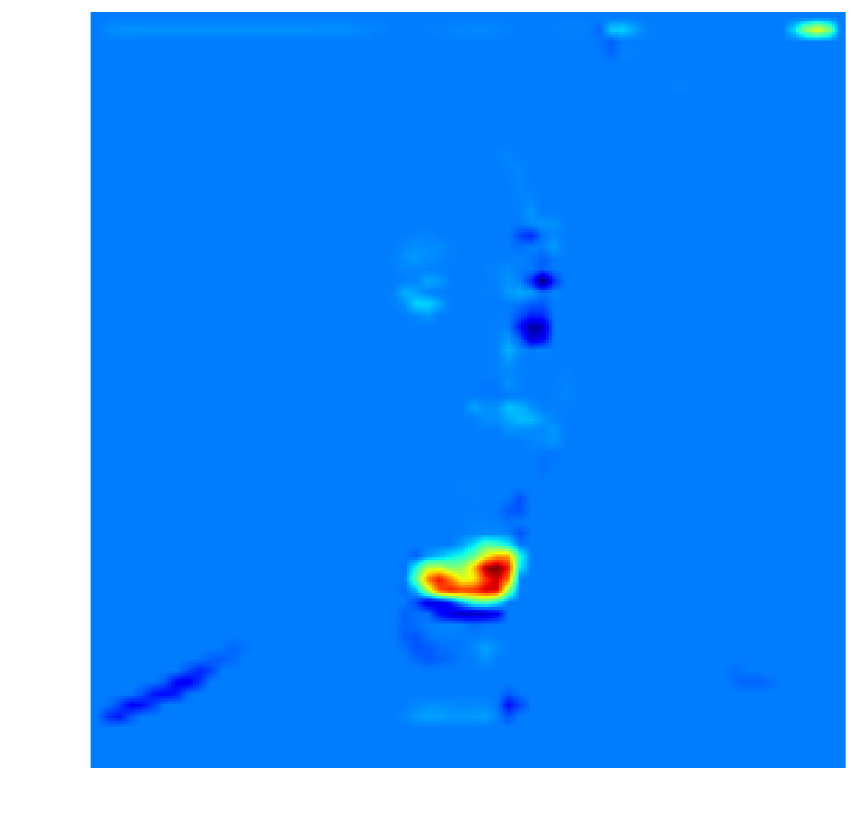}&
			\includegraphics[width=0.1\linewidth]{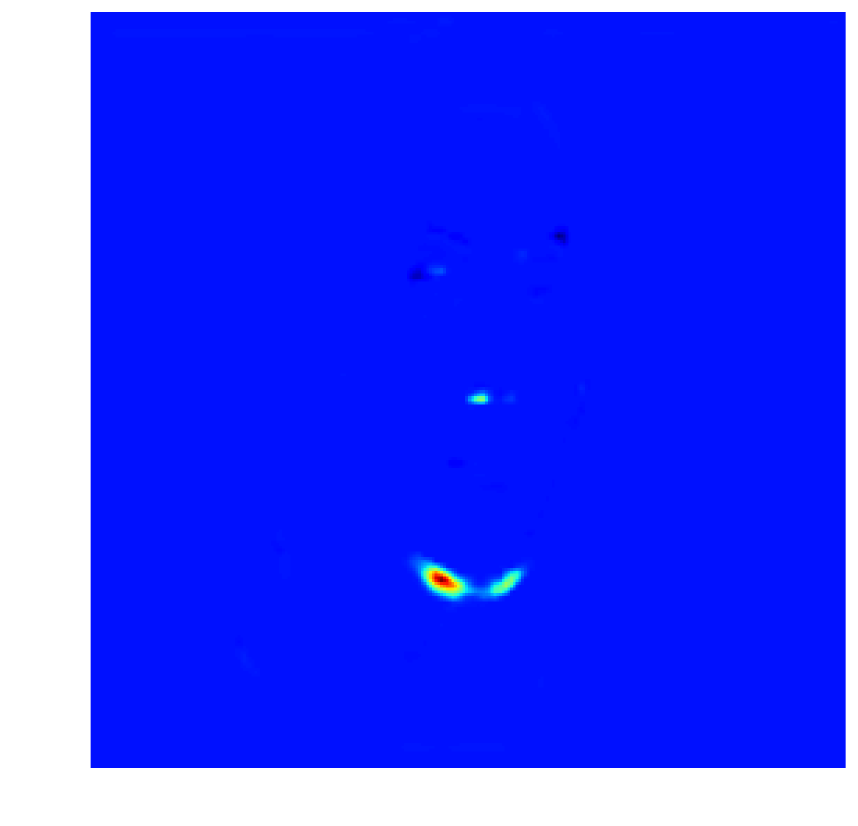}&
			\includegraphics[width=0.1\linewidth]{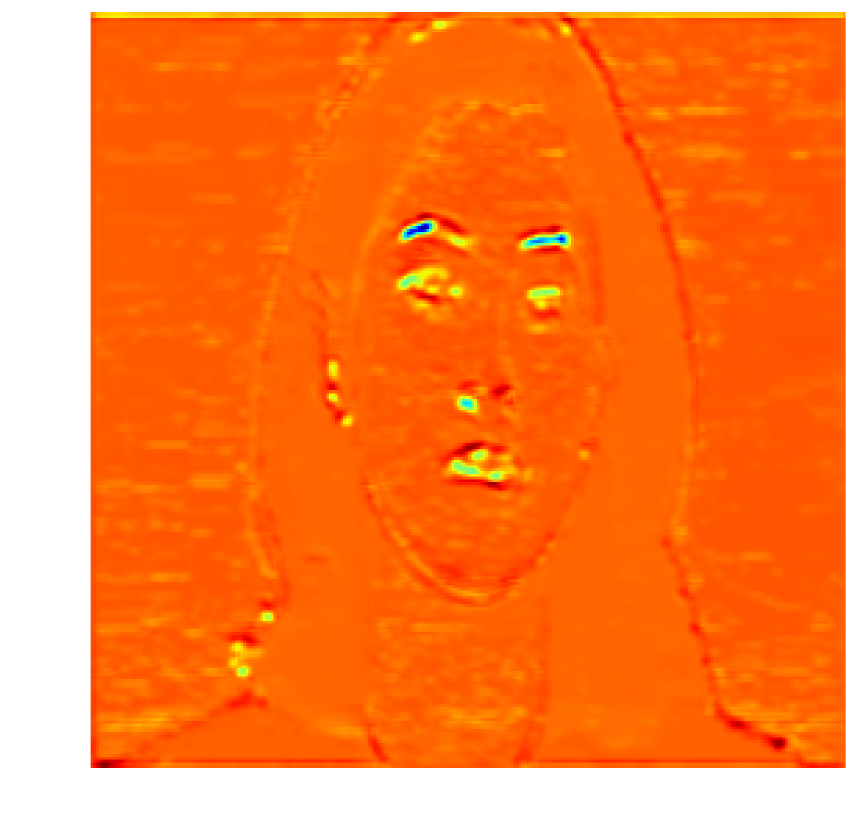}
			\\
			&\includegraphics[width=0.1\linewidth]{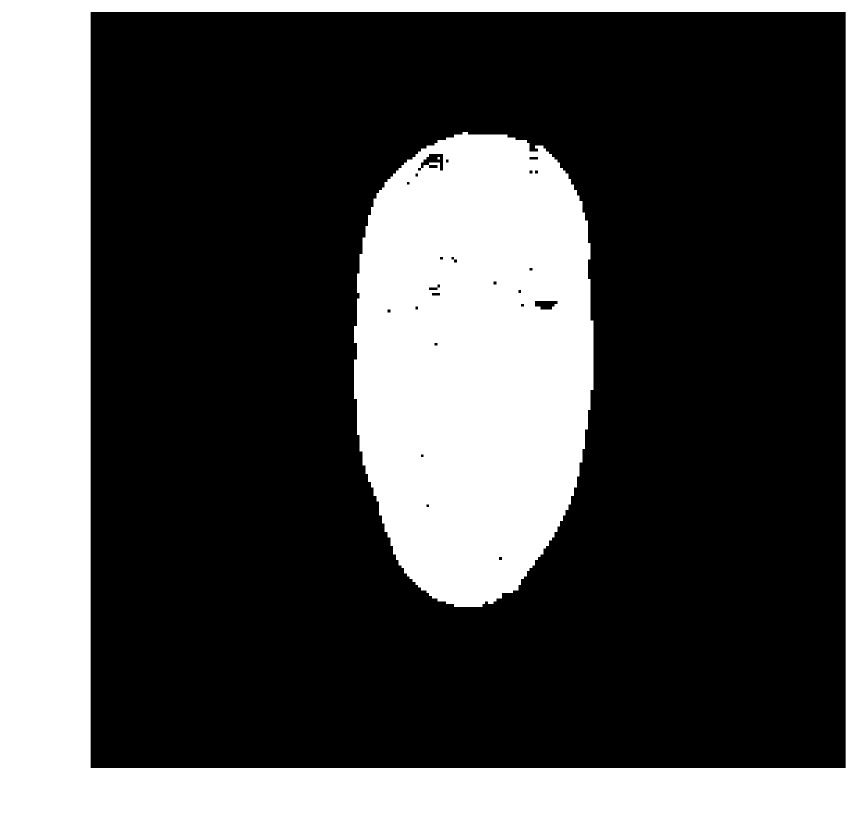}&
			\includegraphics[width=0.1\linewidth]{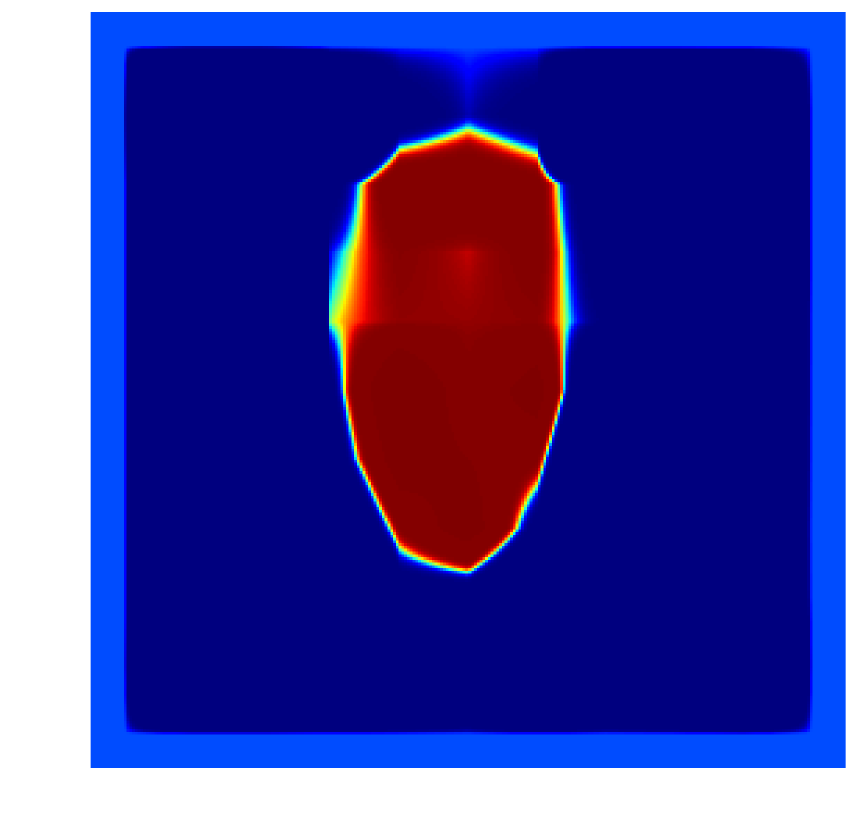}&
			\includegraphics[width=0.1\linewidth]{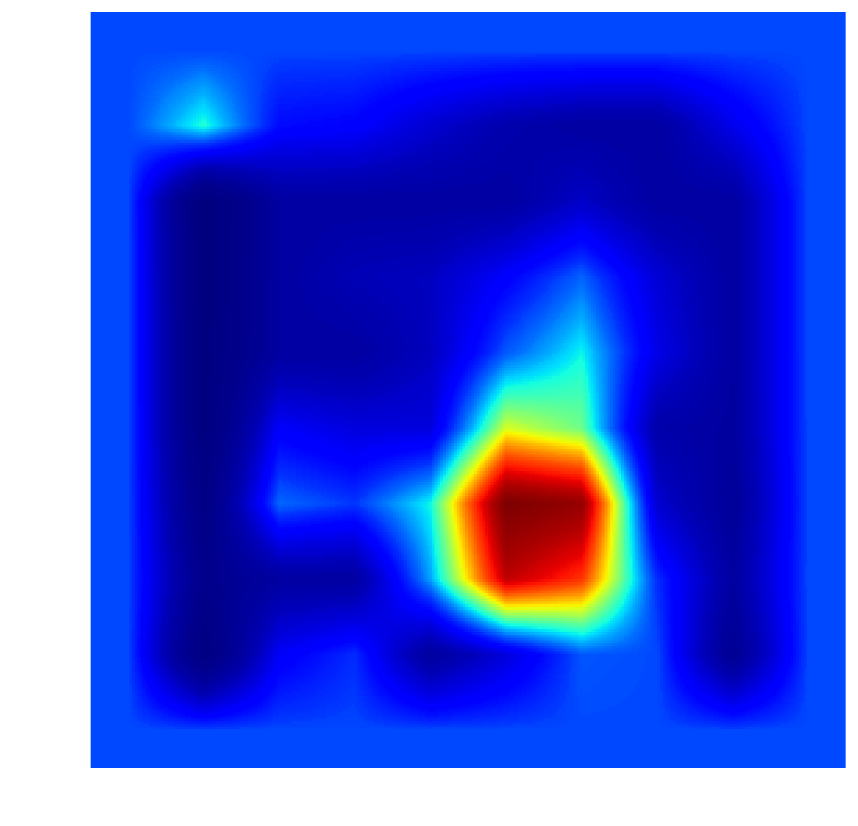}&
			\includegraphics[width=0.1\linewidth]{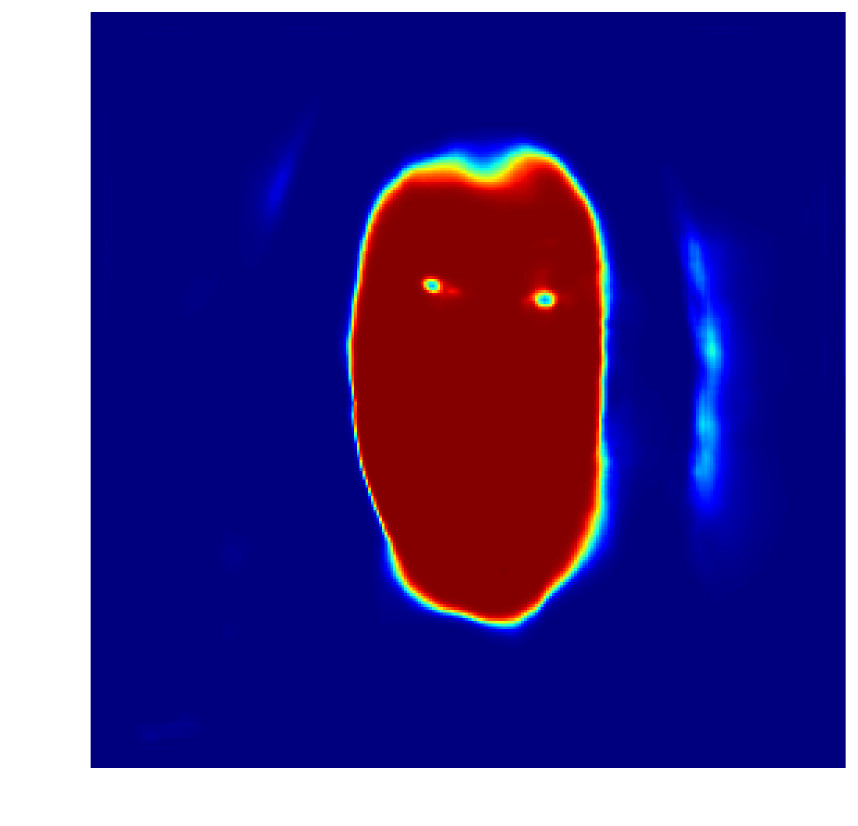}&
			\includegraphics[width=0.1\linewidth]{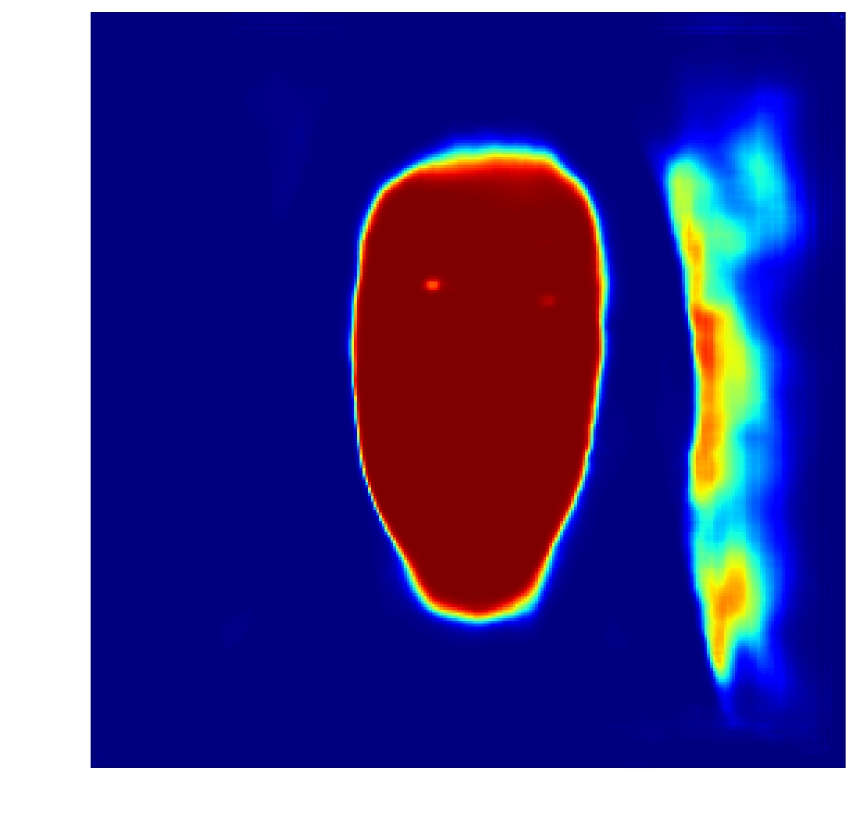}&
			\includegraphics[width=0.1\linewidth]{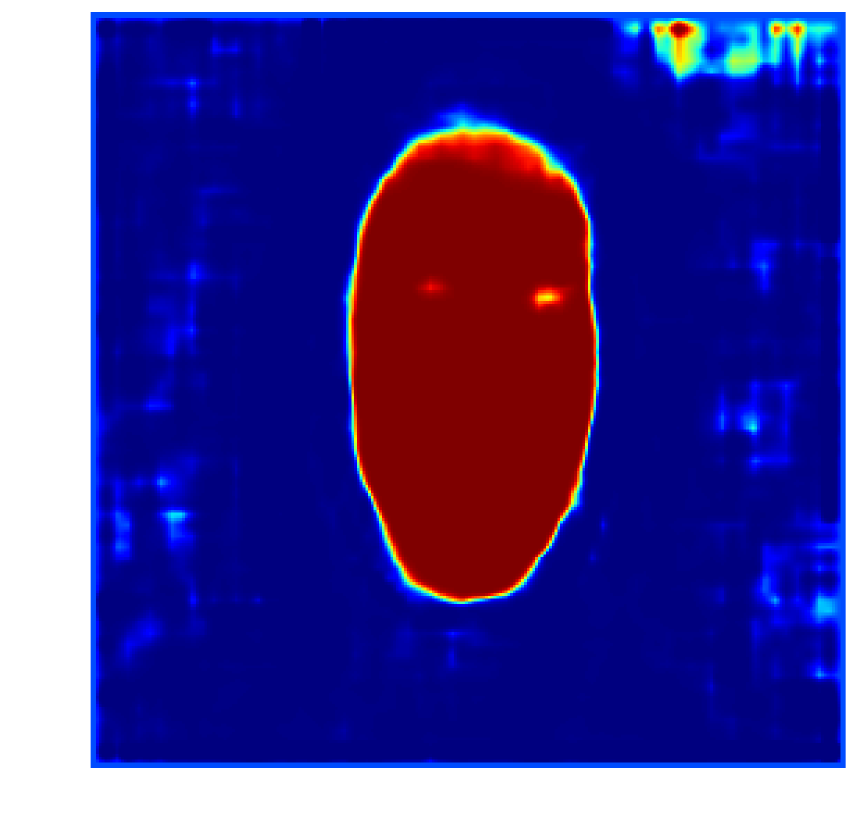}&
			\includegraphics[width=0.1\linewidth]{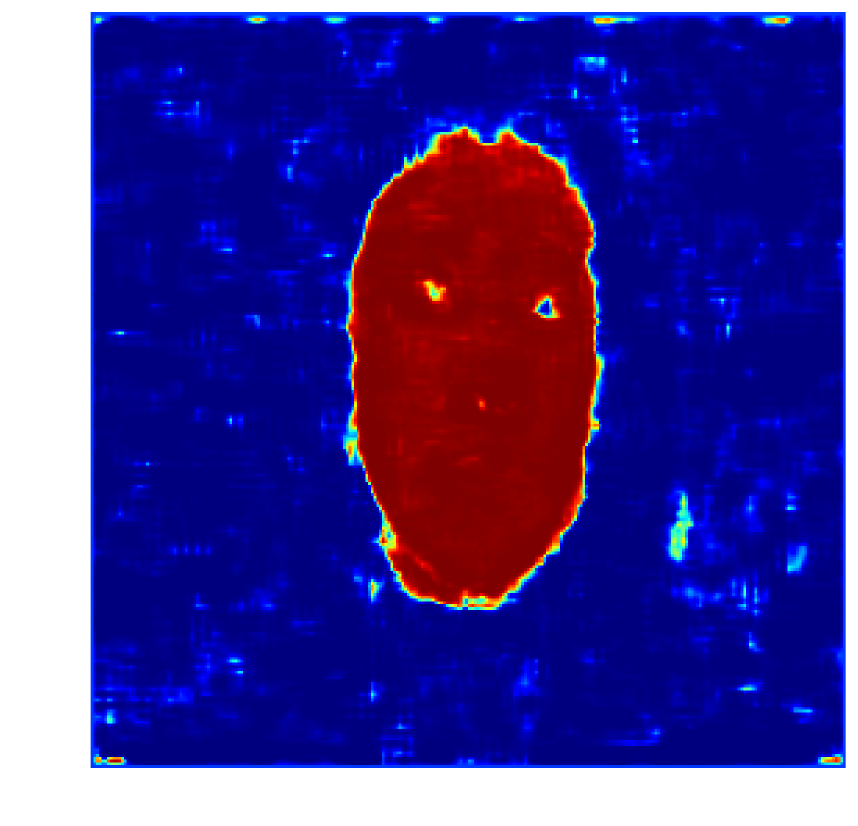}&
			\includegraphics[width=0.1\linewidth]{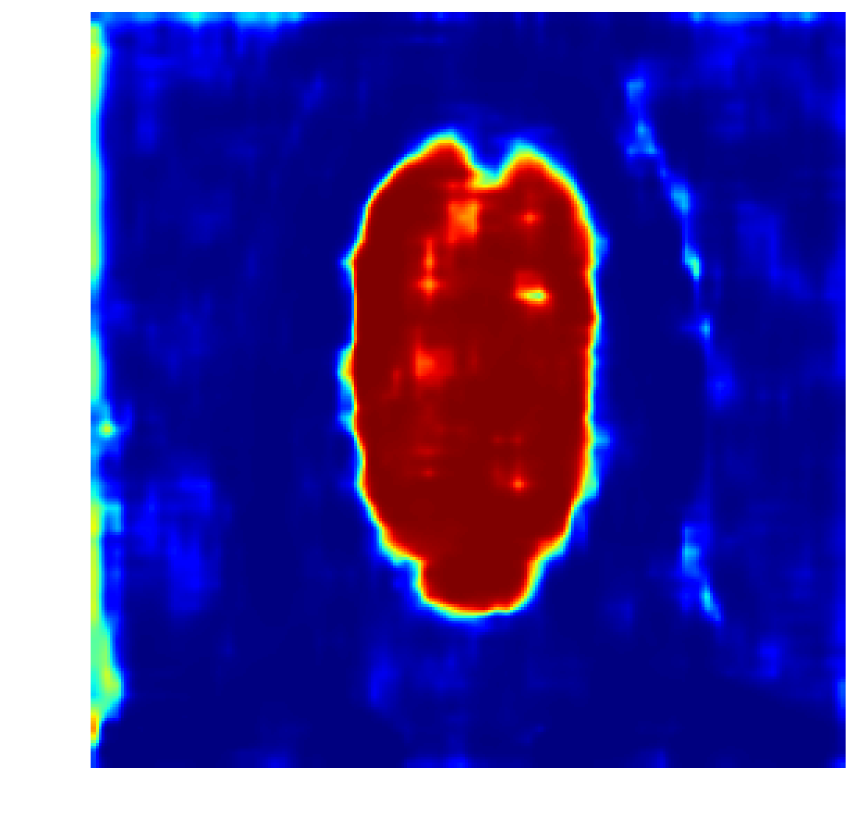}\\
			\hline
			\multirow{2}{*}{FS}&\includegraphics[width=0.1\linewidth]{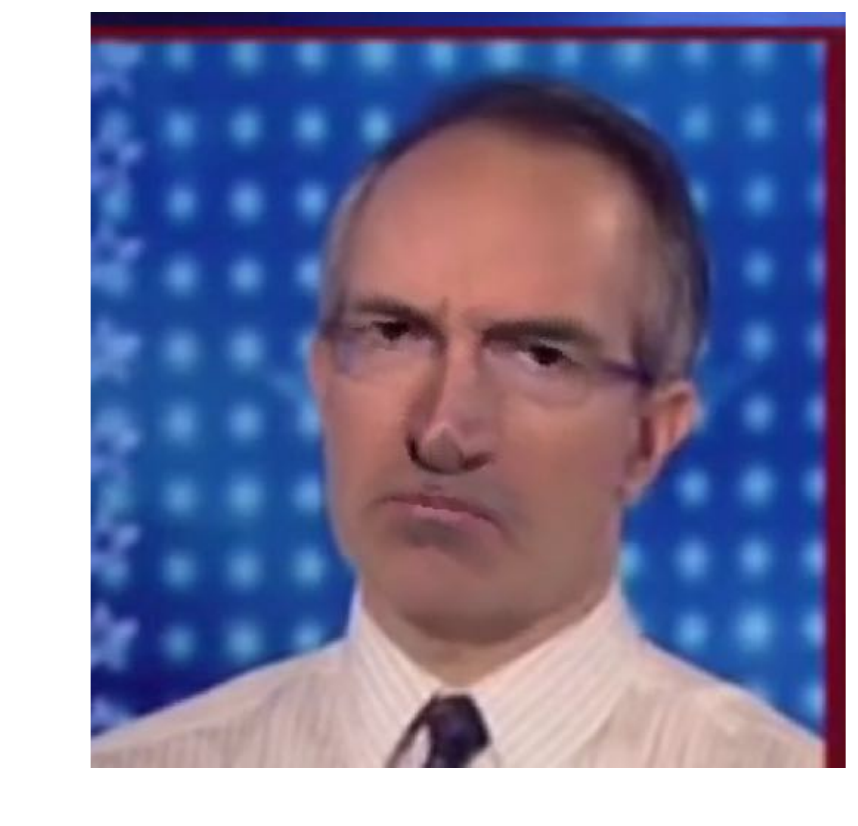}&
			\includegraphics[width=0.1\linewidth]{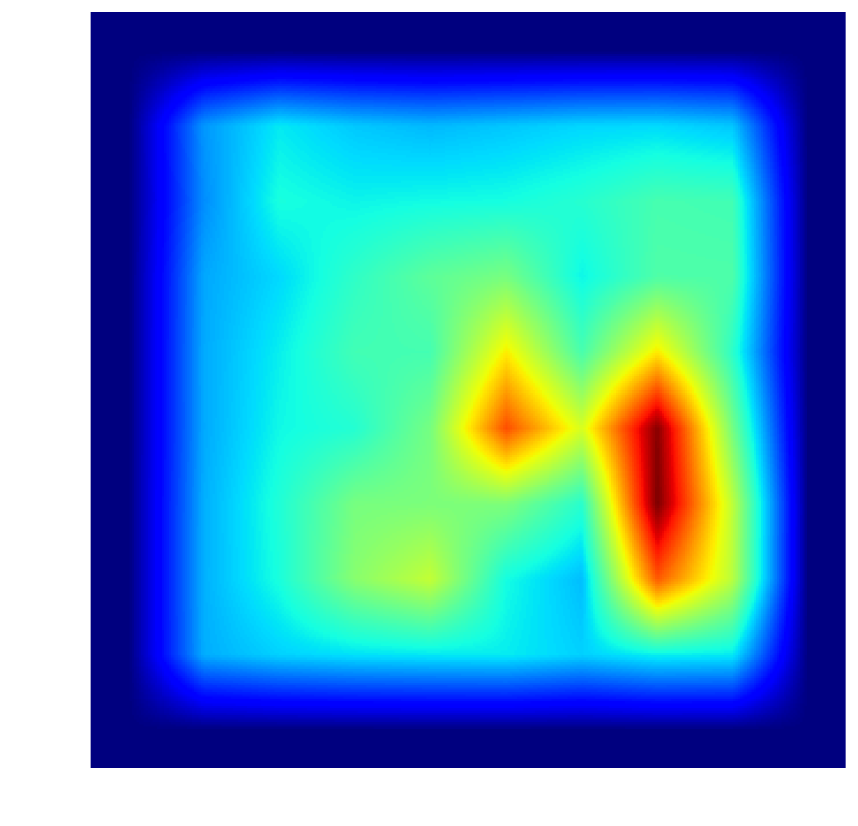}&
			\includegraphics[width=0.1\linewidth]{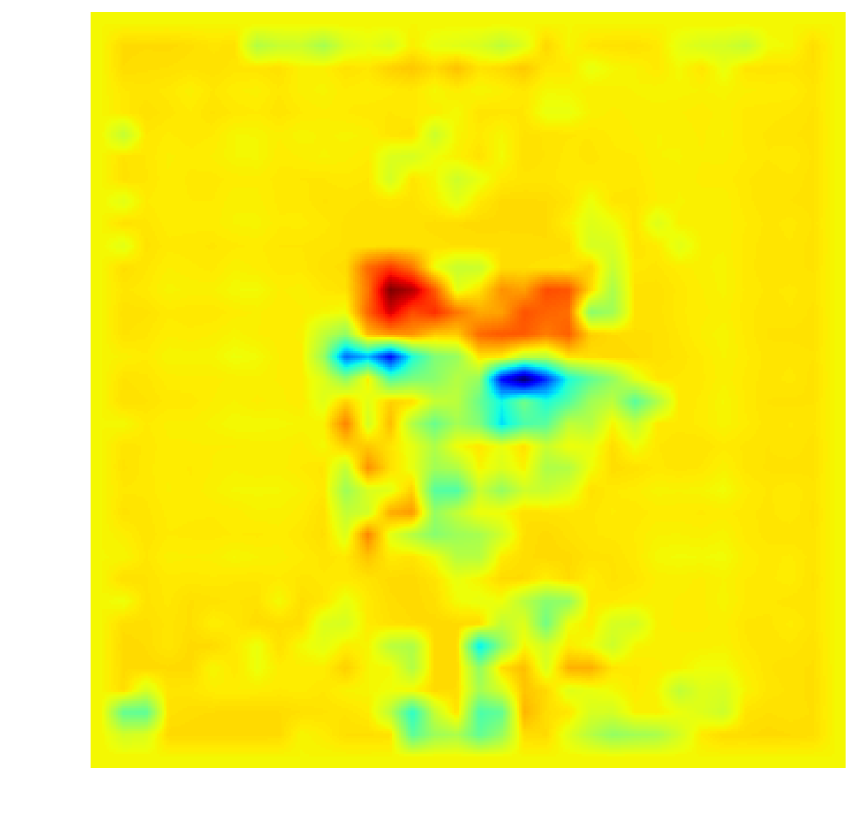}&
			\includegraphics[width=0.1\linewidth]{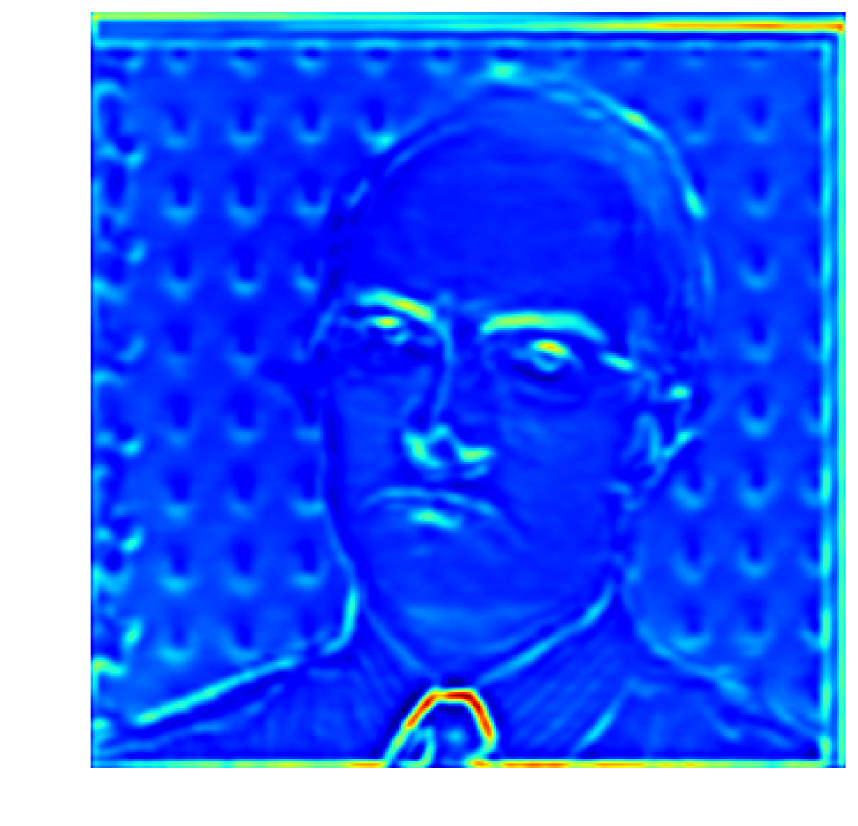}&
			\includegraphics[width=0.1\linewidth]{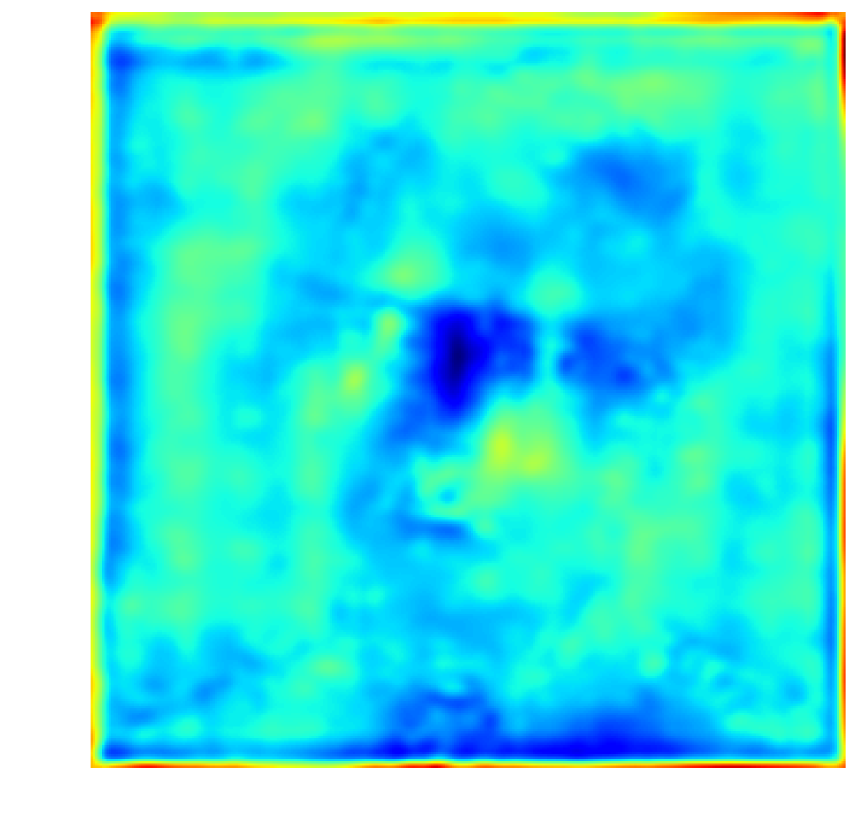}&
			\includegraphics[width=0.1\linewidth]{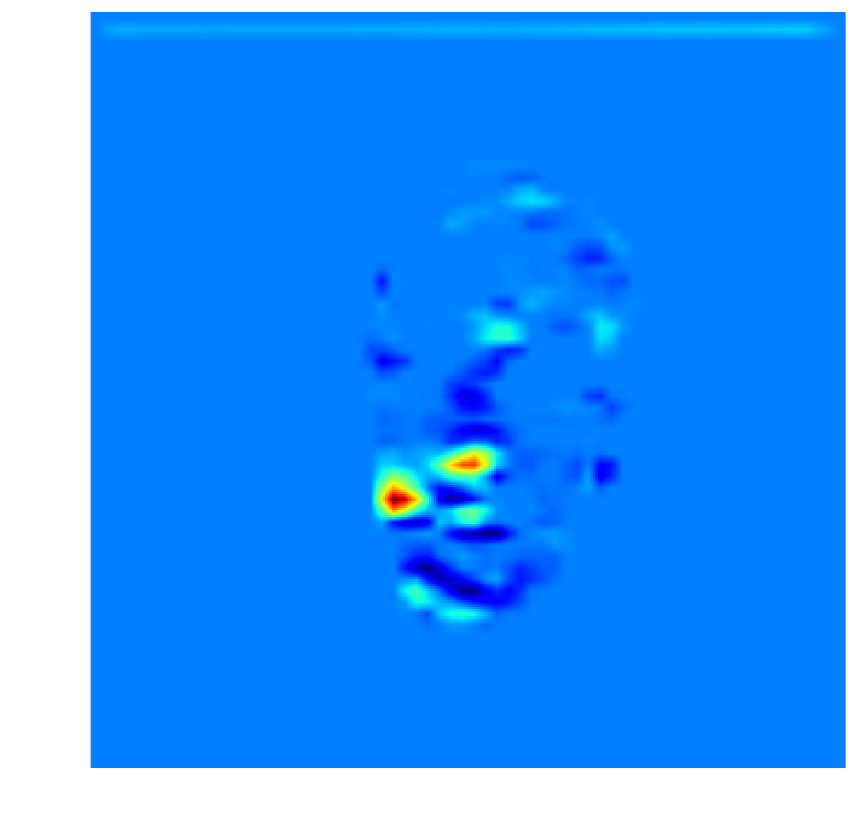}&
			\includegraphics[width=0.1\linewidth]{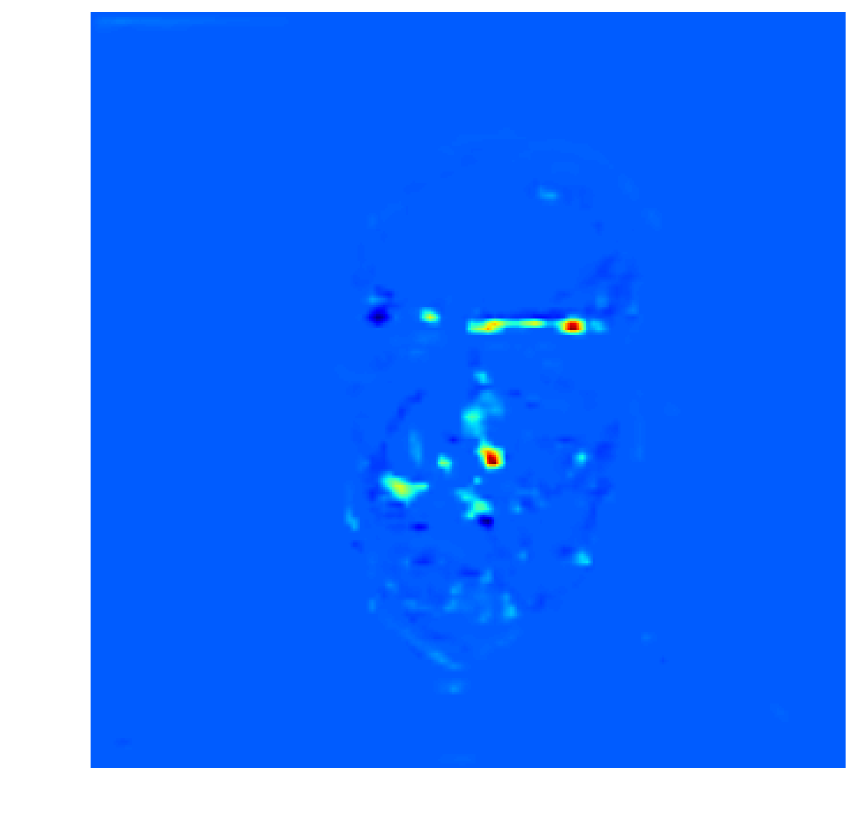}&
			\includegraphics[width=0.1\linewidth]{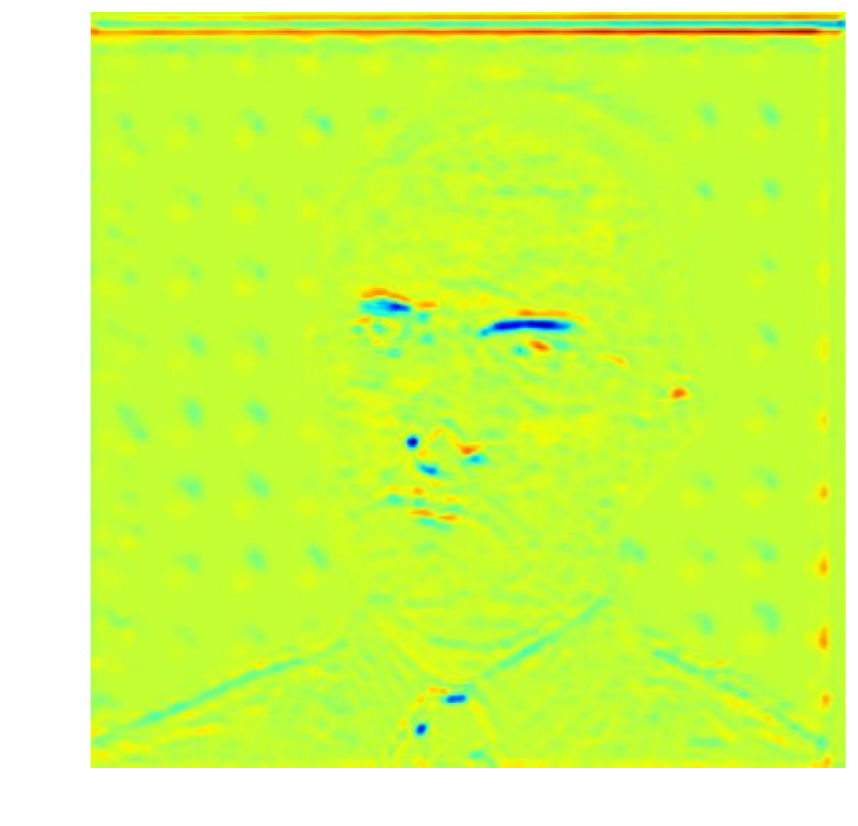}
			\\
			&\includegraphics[width=0.1\linewidth]{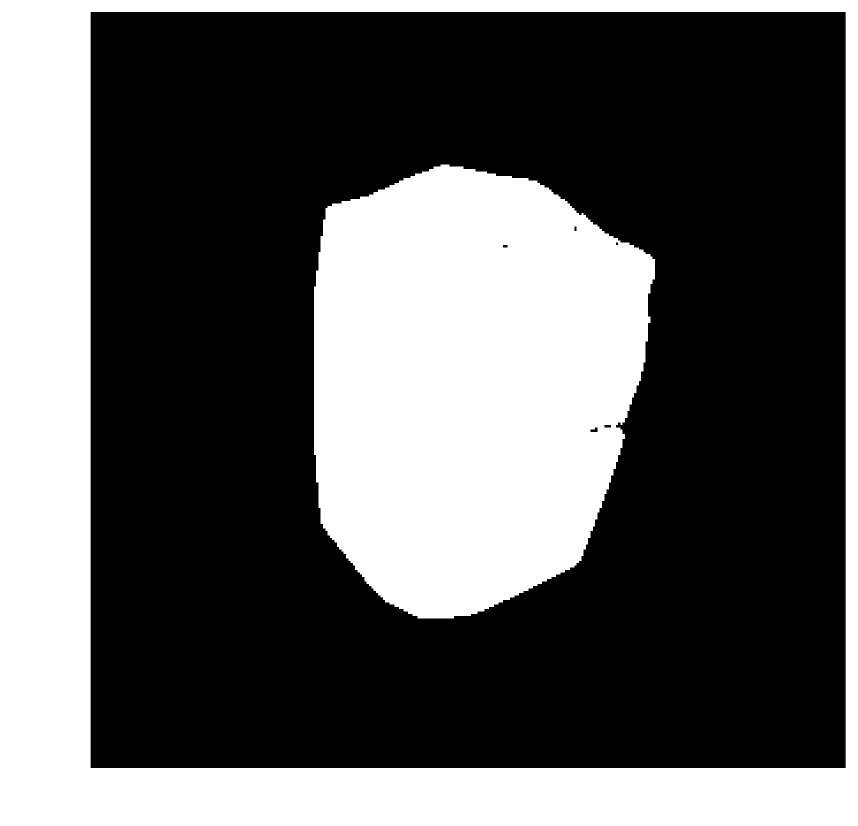}&
			\includegraphics[width=0.1\linewidth]{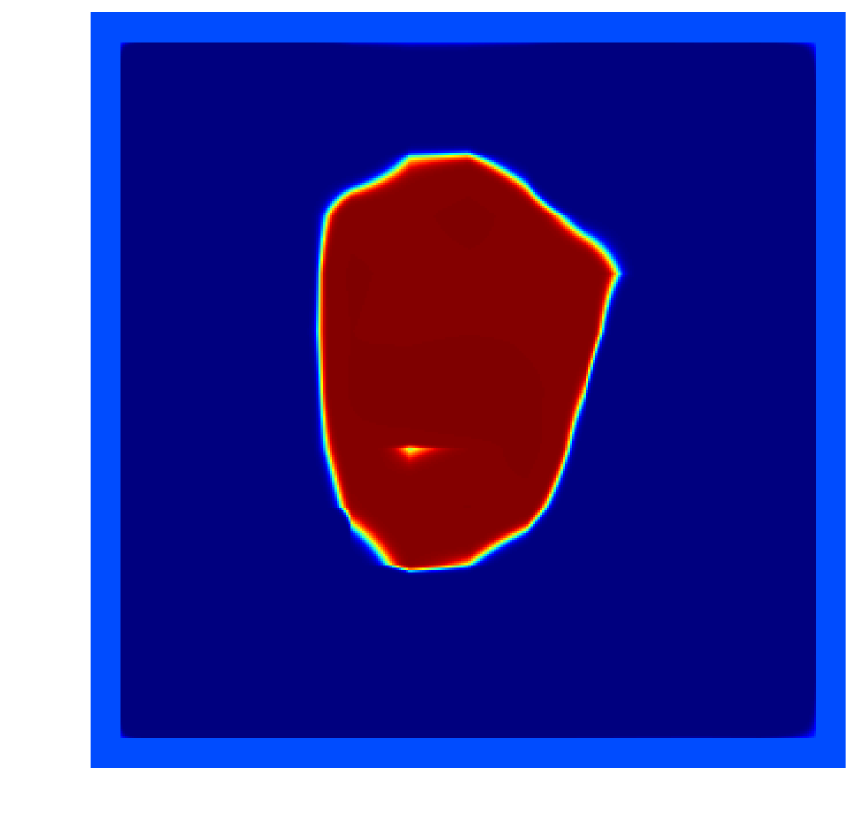}&
			\includegraphics[width=0.1\linewidth]{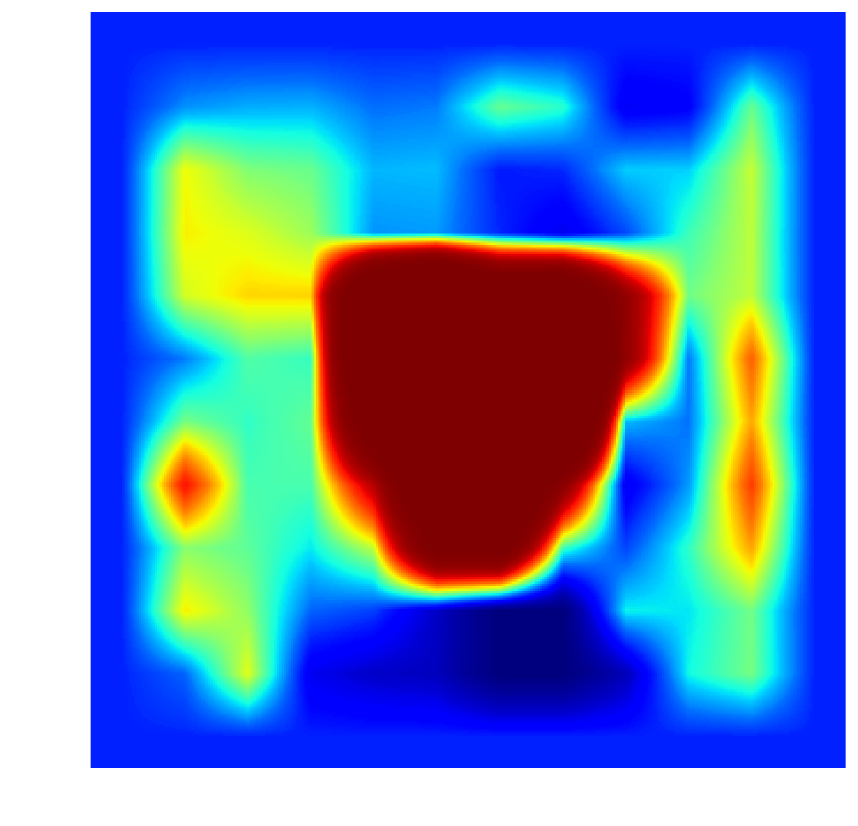}&
			\includegraphics[width=0.1\linewidth]{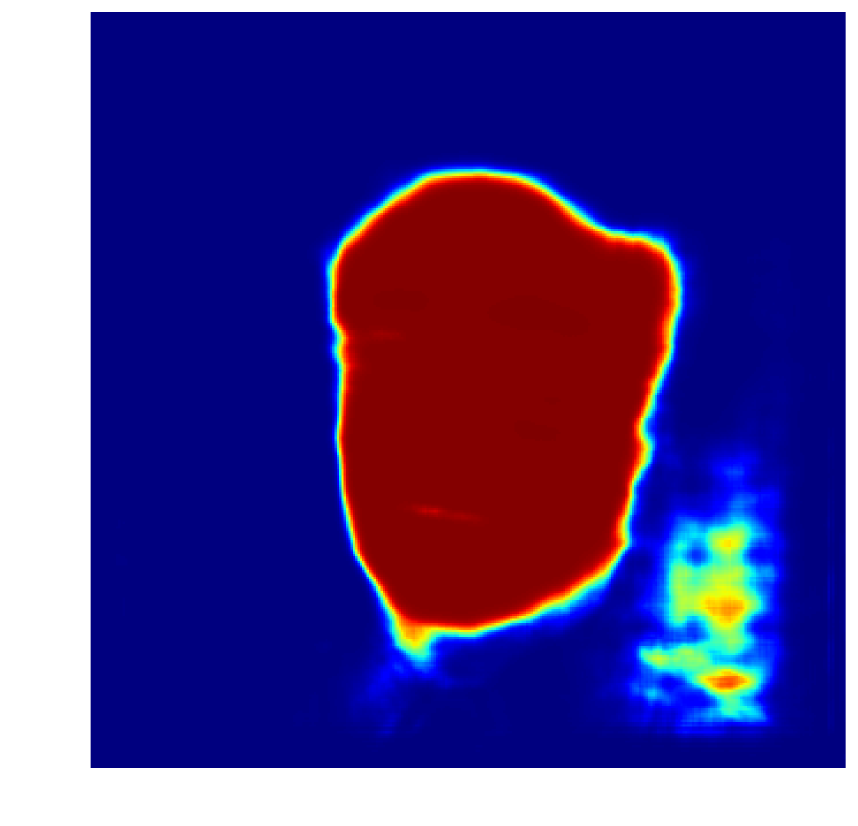}&
			\includegraphics[width=0.1\linewidth]{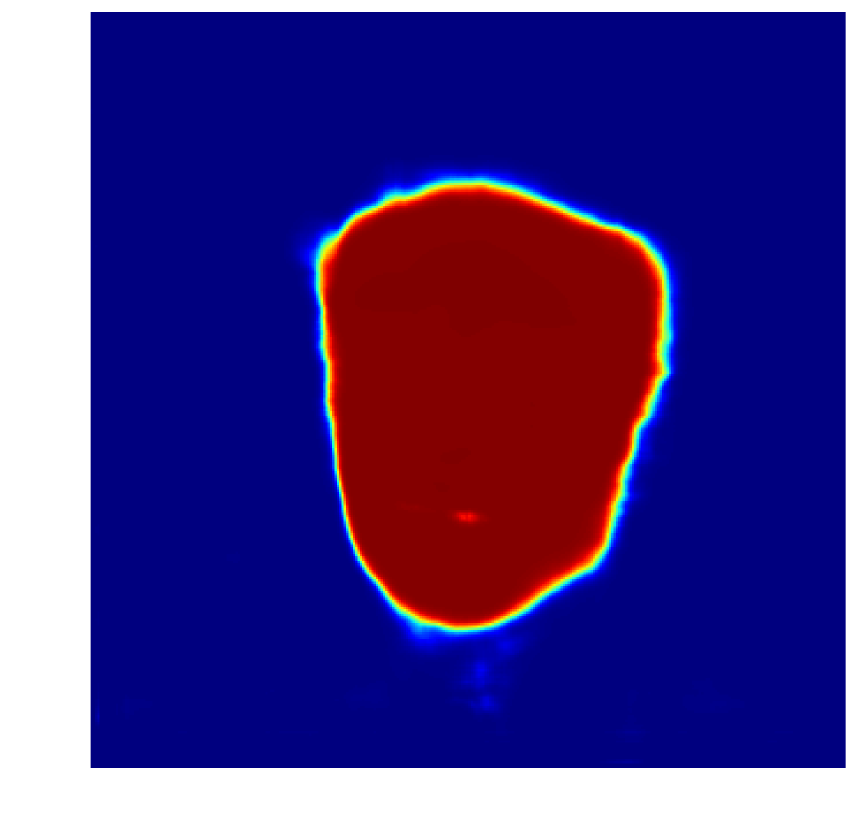}&
			\includegraphics[width=0.1\linewidth]{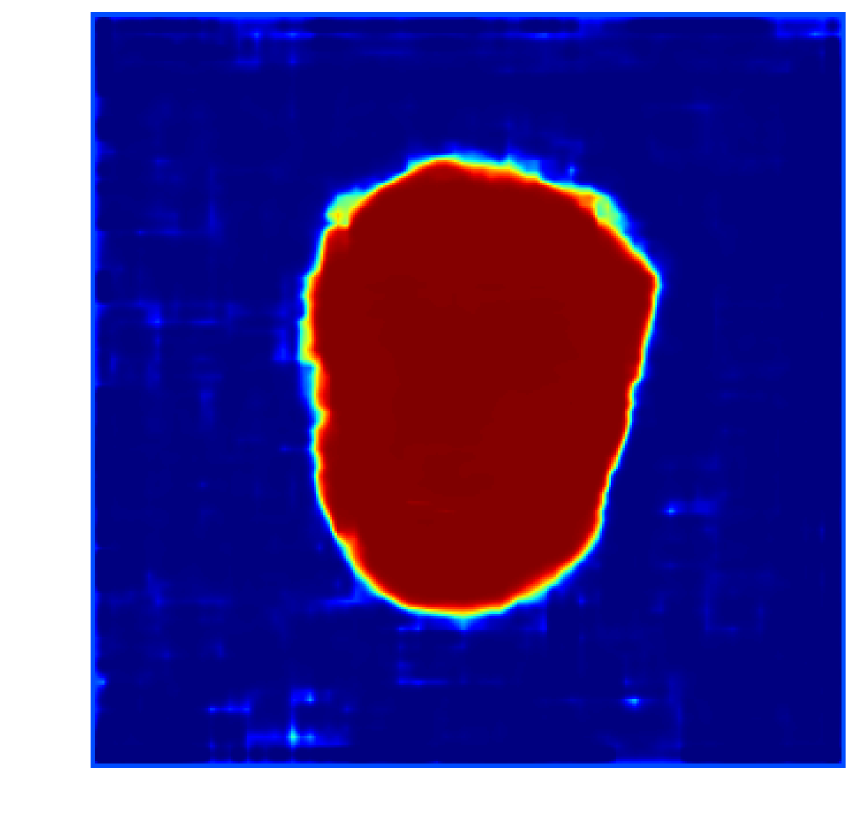}&
			\includegraphics[width=0.1\linewidth]{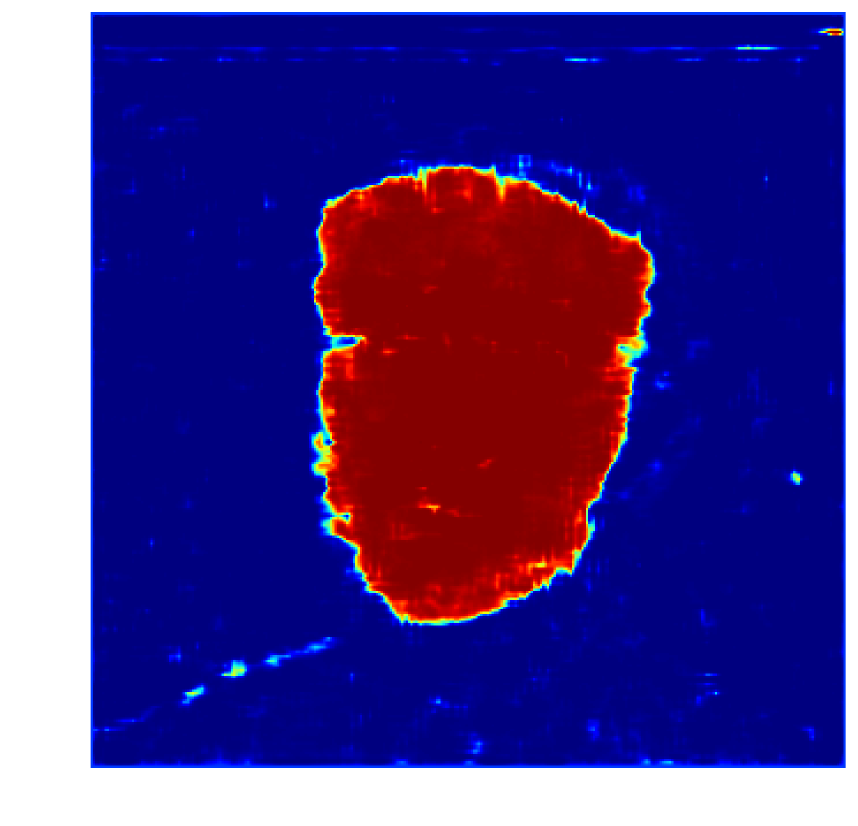}&
			\includegraphics[width=0.1\linewidth]{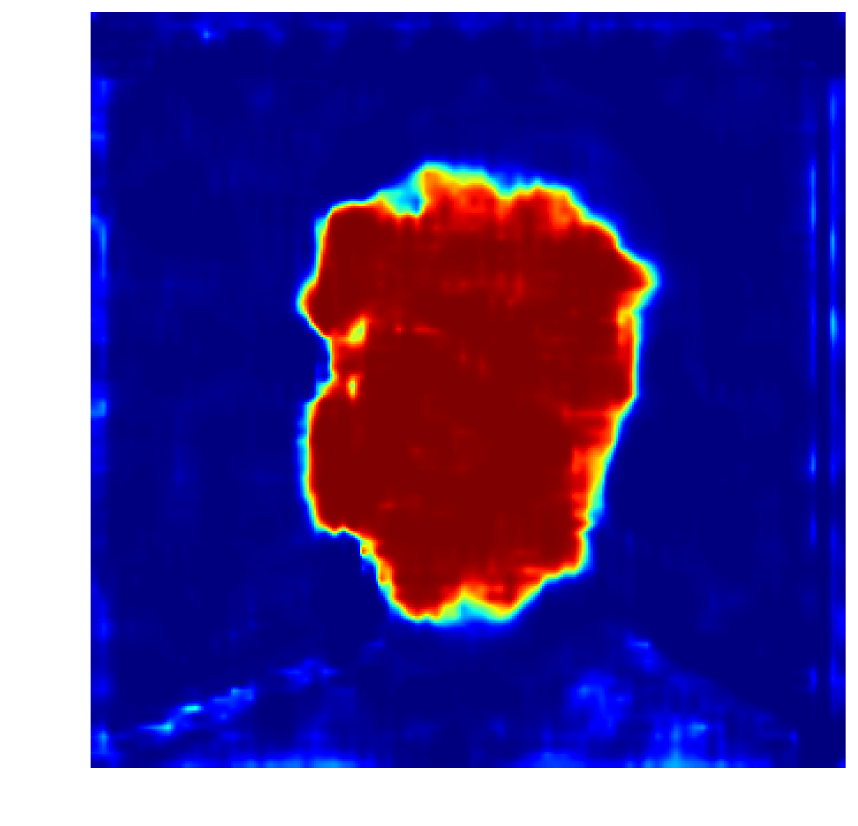}\\
			\hline
			\multirow{2}{*}{NT}&\includegraphics[width=0.1\linewidth]{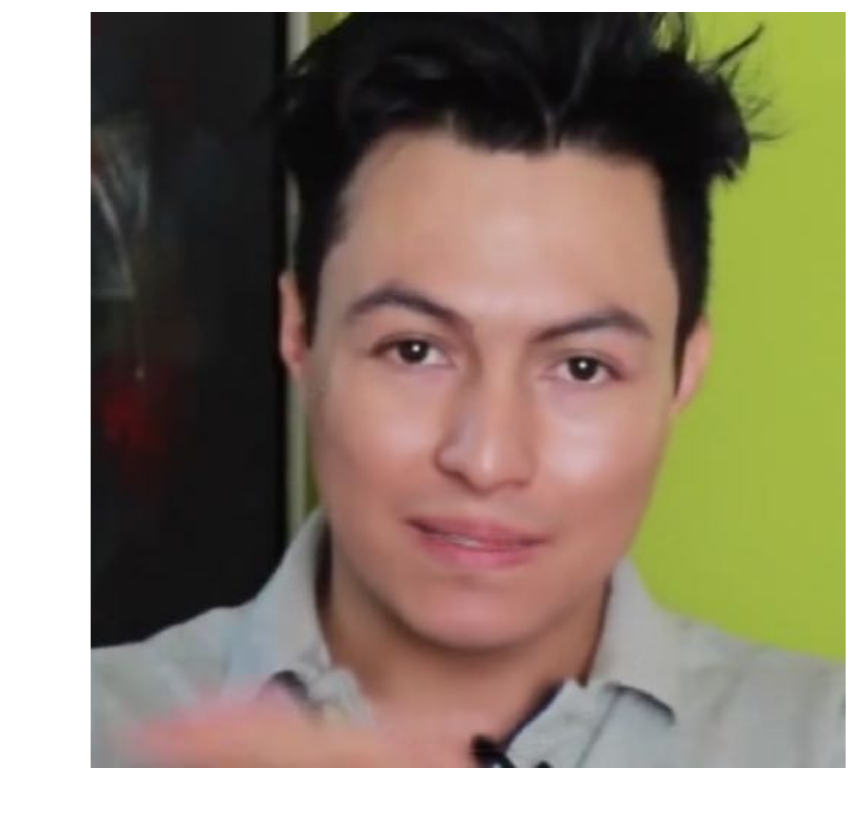}&
			\includegraphics[width=0.1\linewidth]{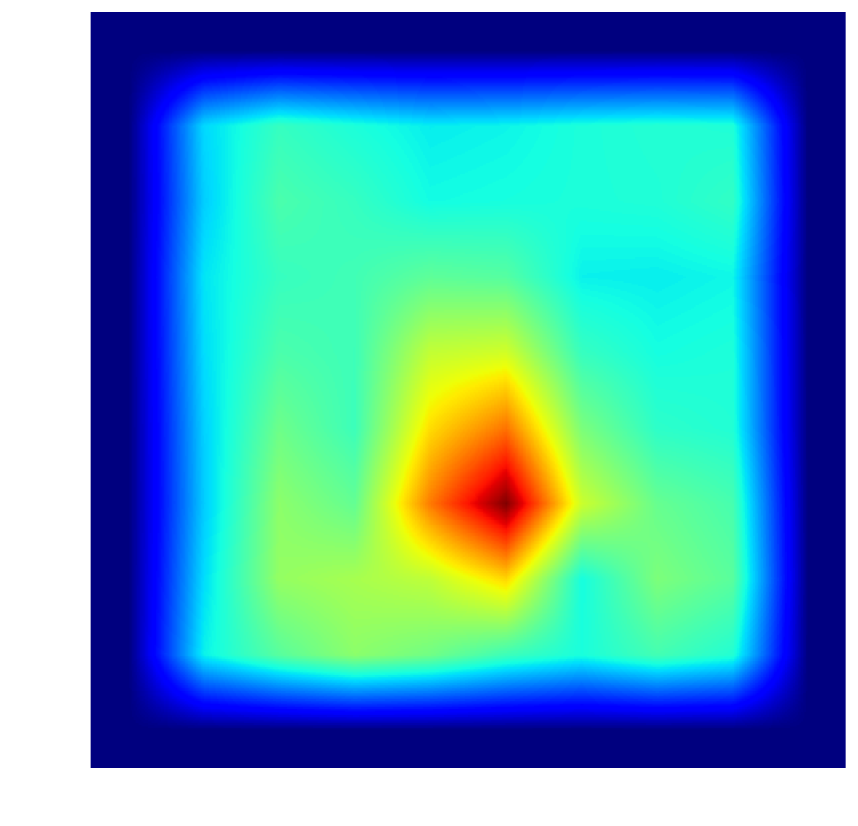}&
			\includegraphics[width=0.1\linewidth]{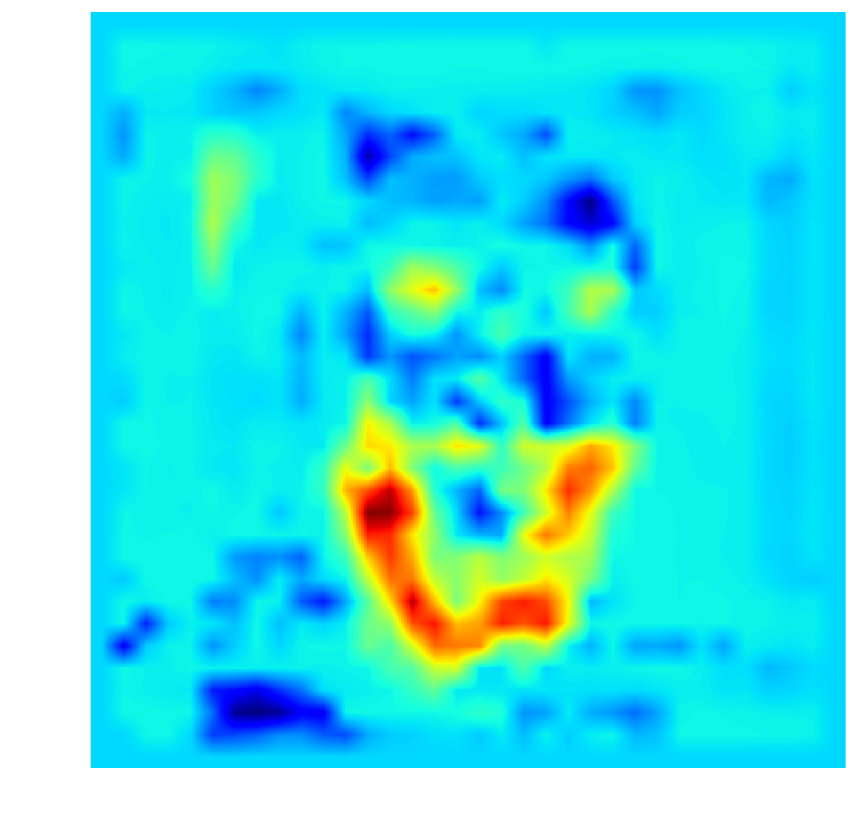}&
			\includegraphics[width=0.1\linewidth]{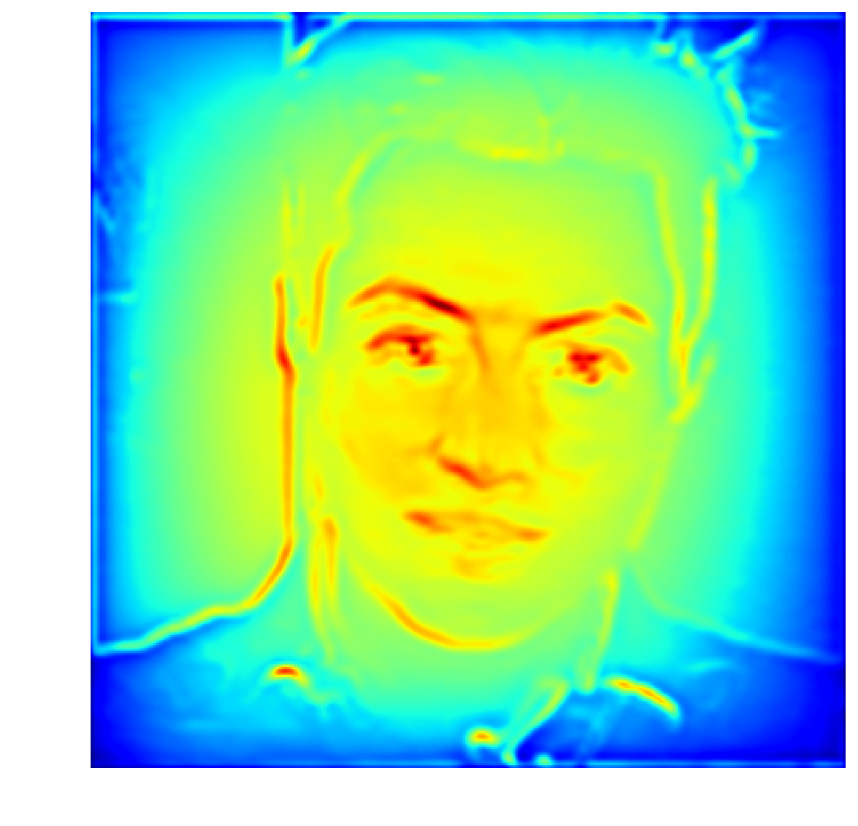}&
			\includegraphics[width=0.1\linewidth]{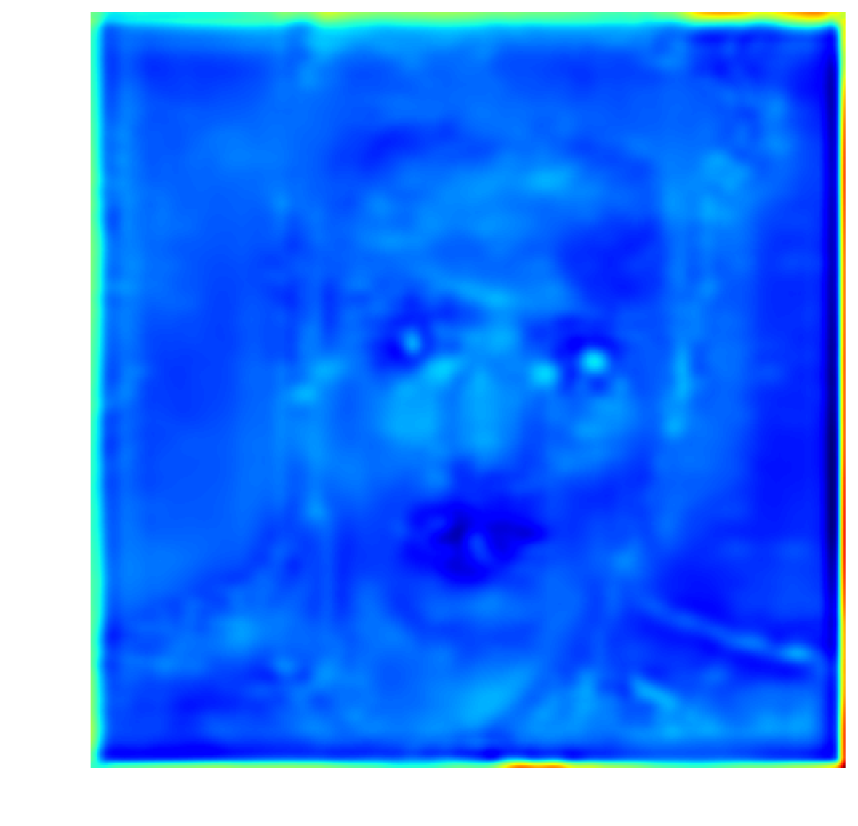}&
			\includegraphics[width=0.1\linewidth]{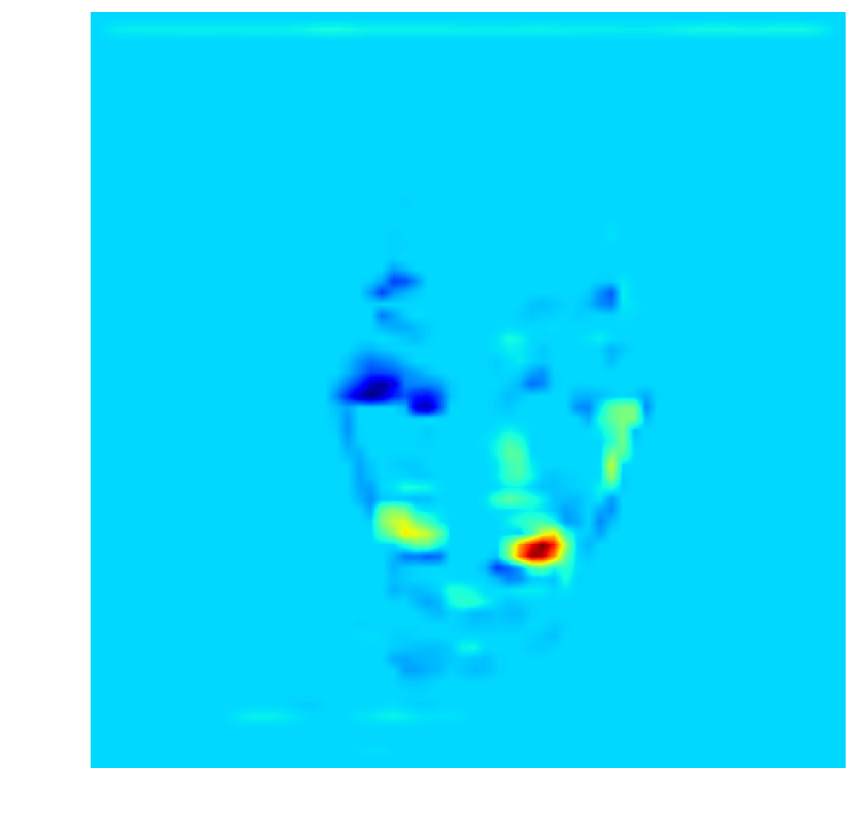}&
			\includegraphics[width=0.1\linewidth]{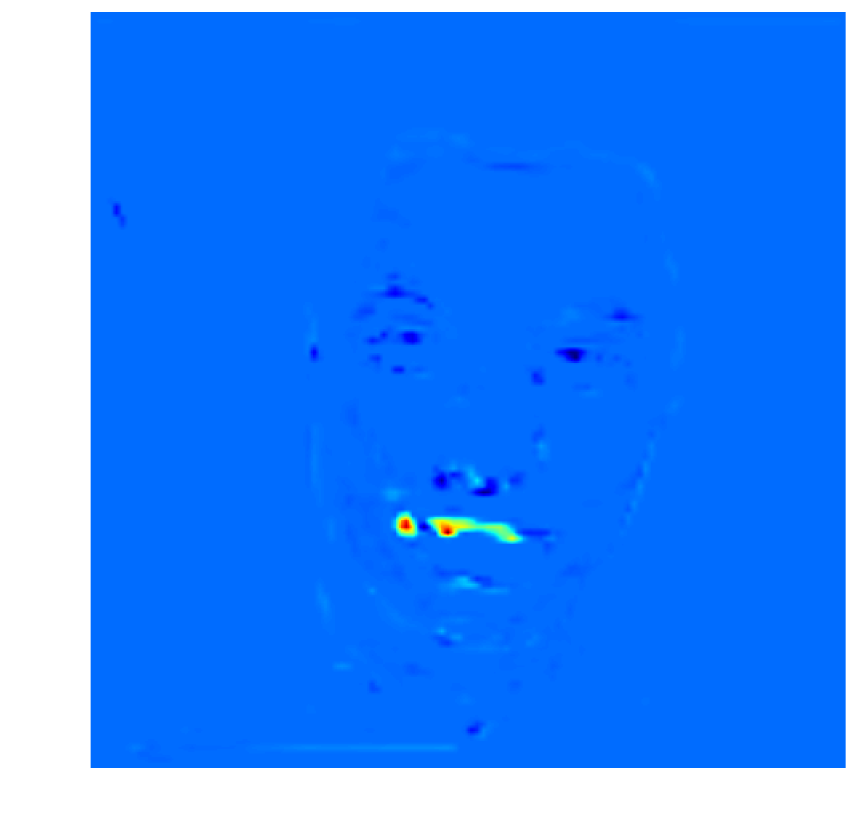}&
			\includegraphics[width=0.1\linewidth]{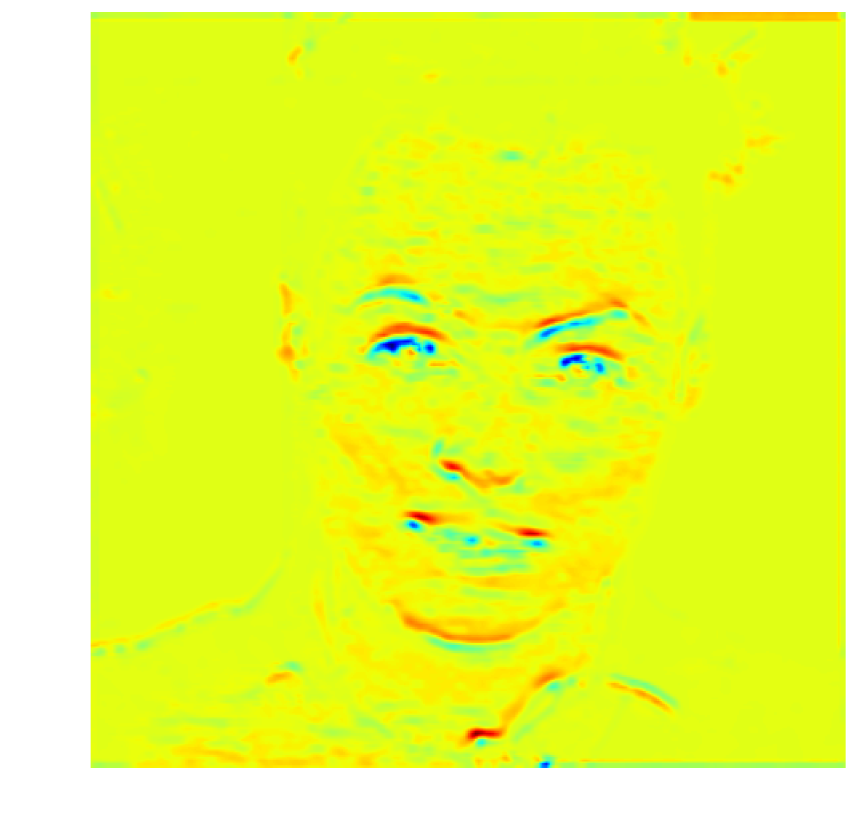}
			\\
			&\includegraphics[width=0.1\linewidth]{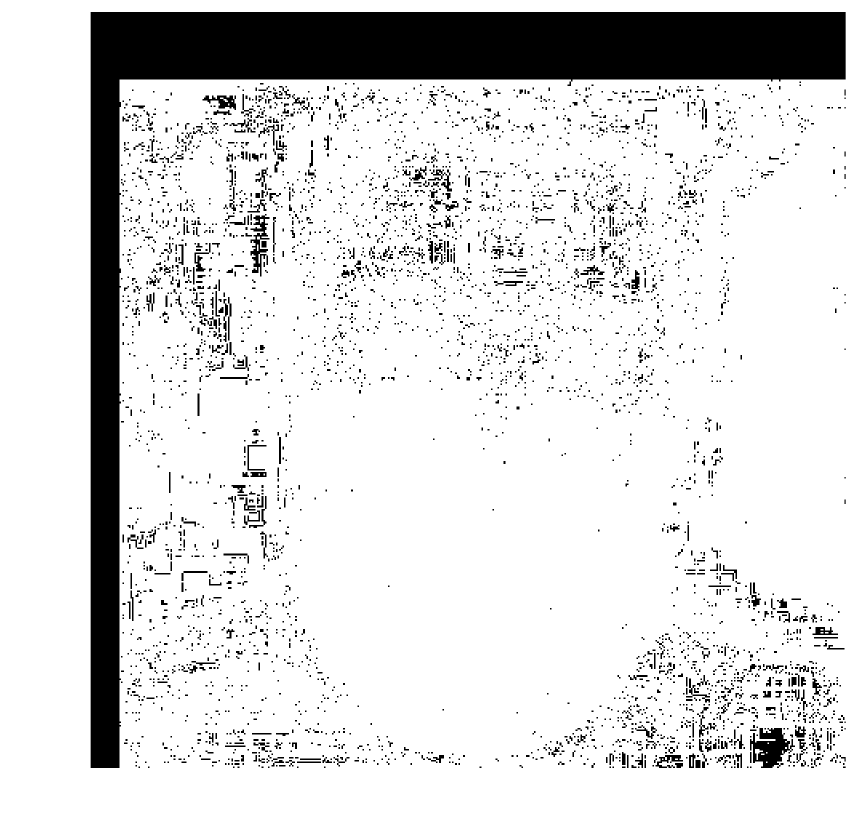}&
			\includegraphics[width=0.1\linewidth]{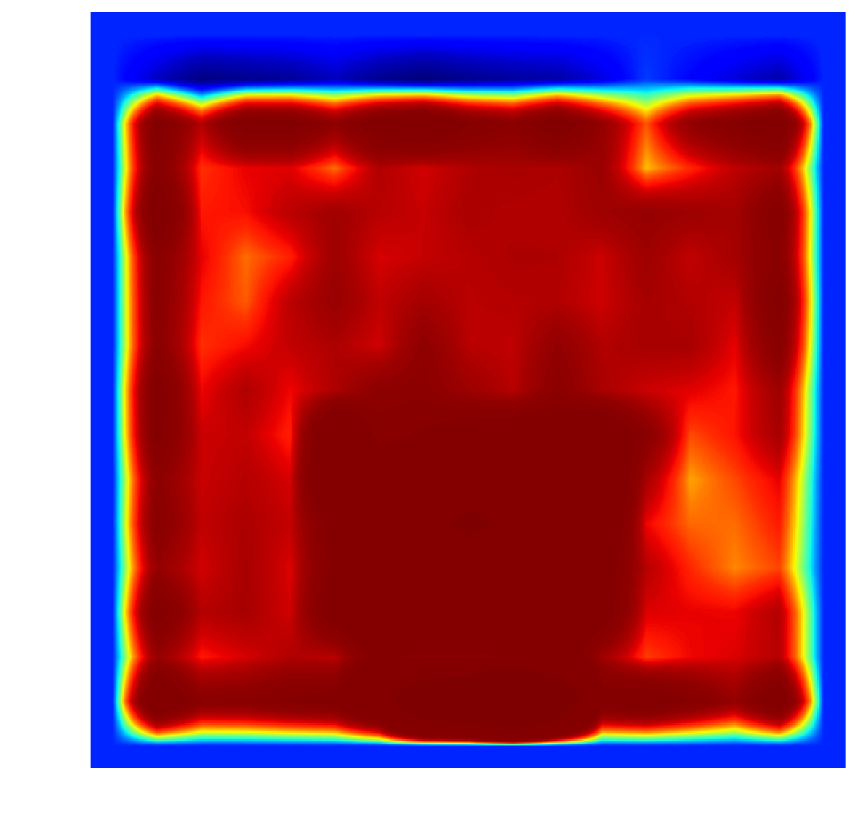}&
			\includegraphics[width=0.1\linewidth]{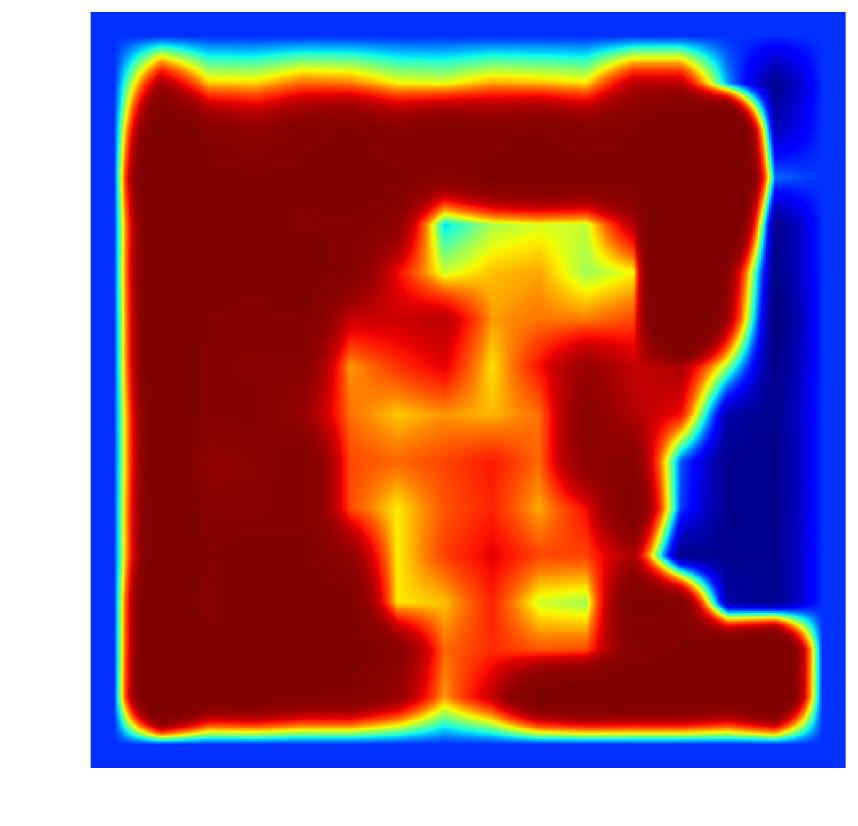}&
			\includegraphics[width=0.1\linewidth]{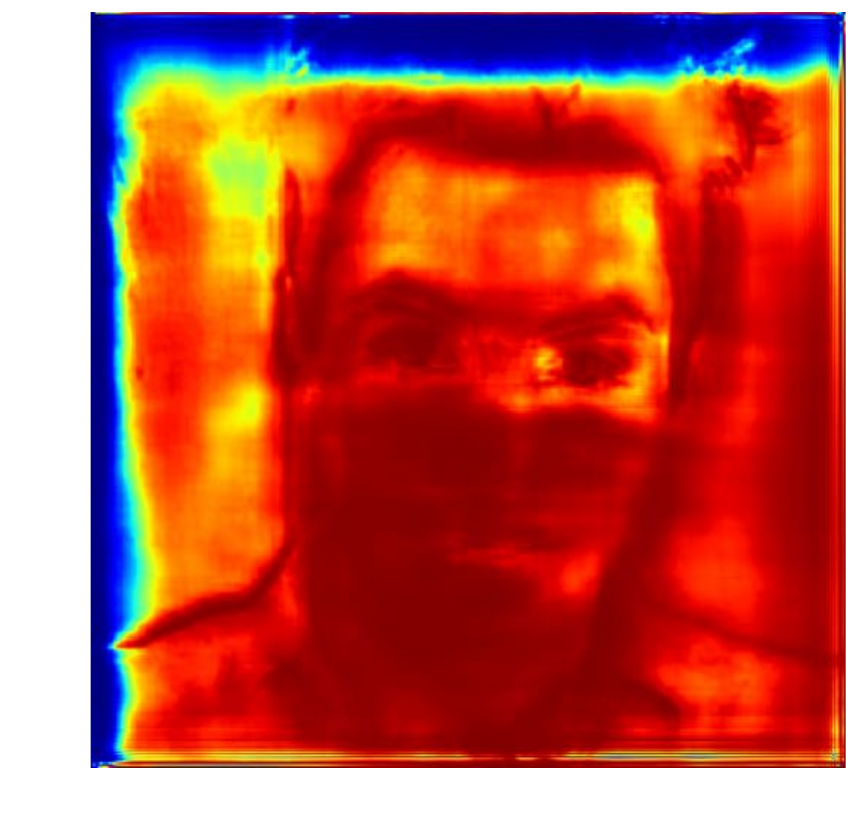}&
			\includegraphics[width=0.1\linewidth]{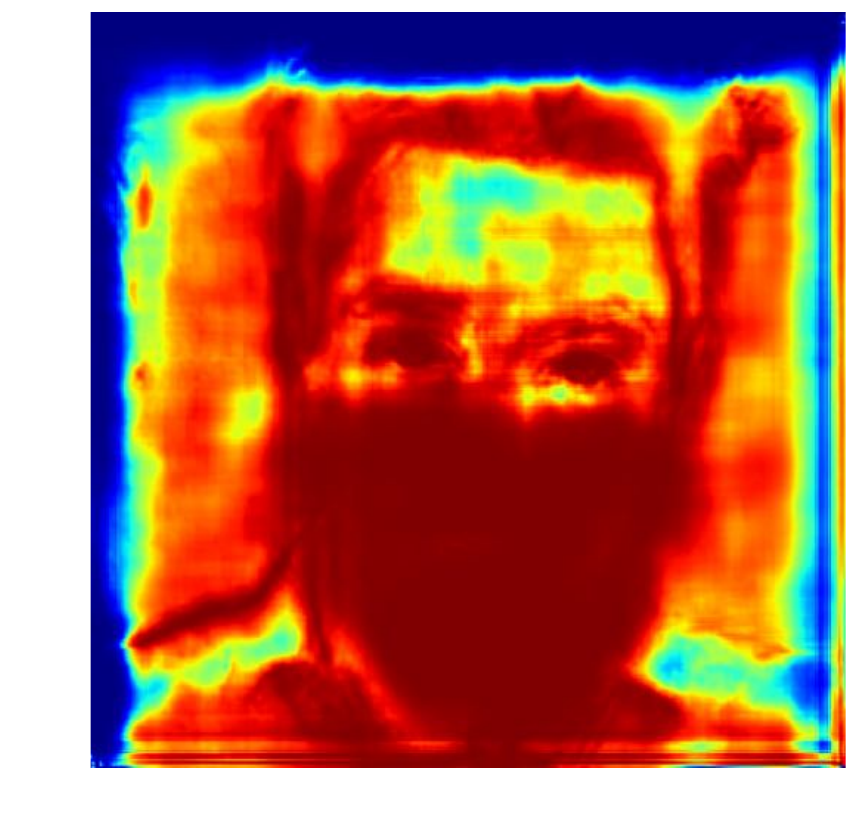}&
			\includegraphics[width=0.1\linewidth]{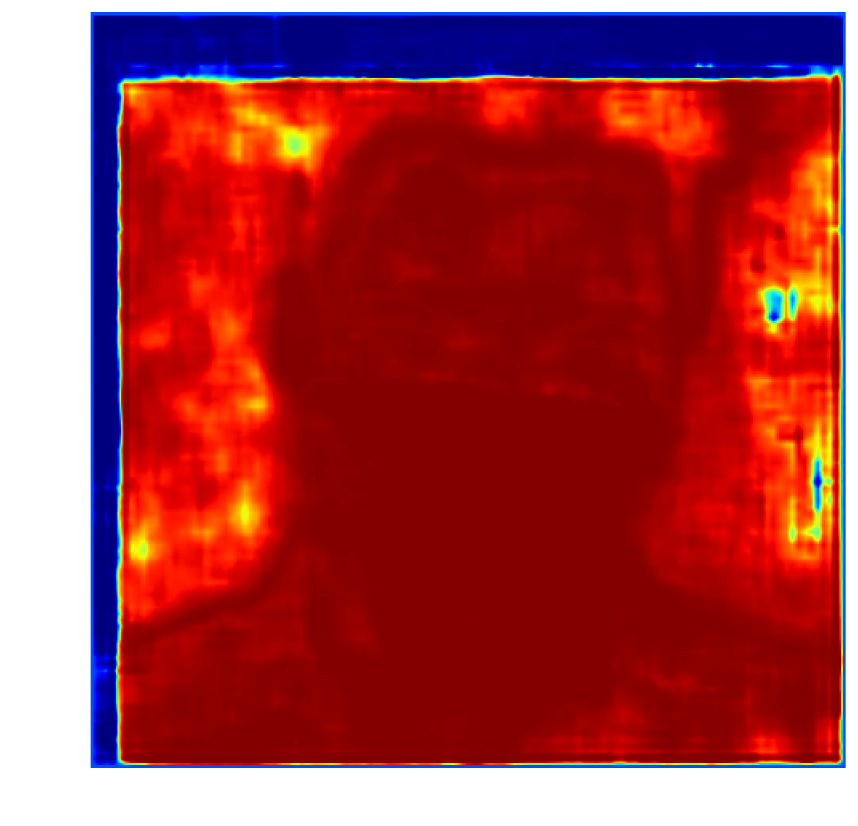}&
			\includegraphics[width=0.1\linewidth]{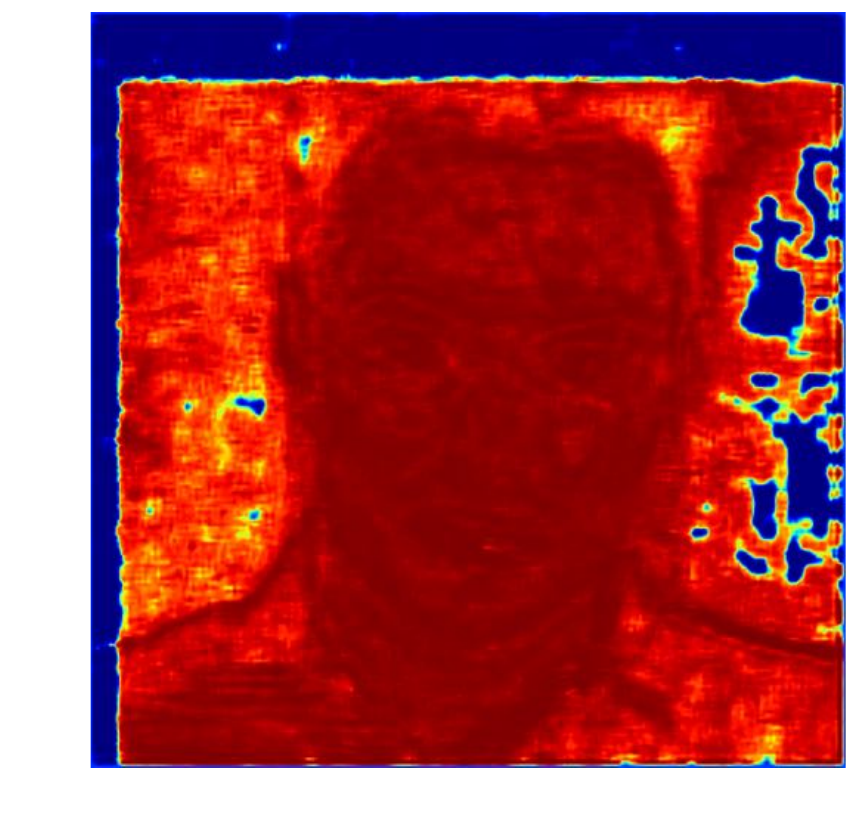}&
			\includegraphics[width=0.1\linewidth]{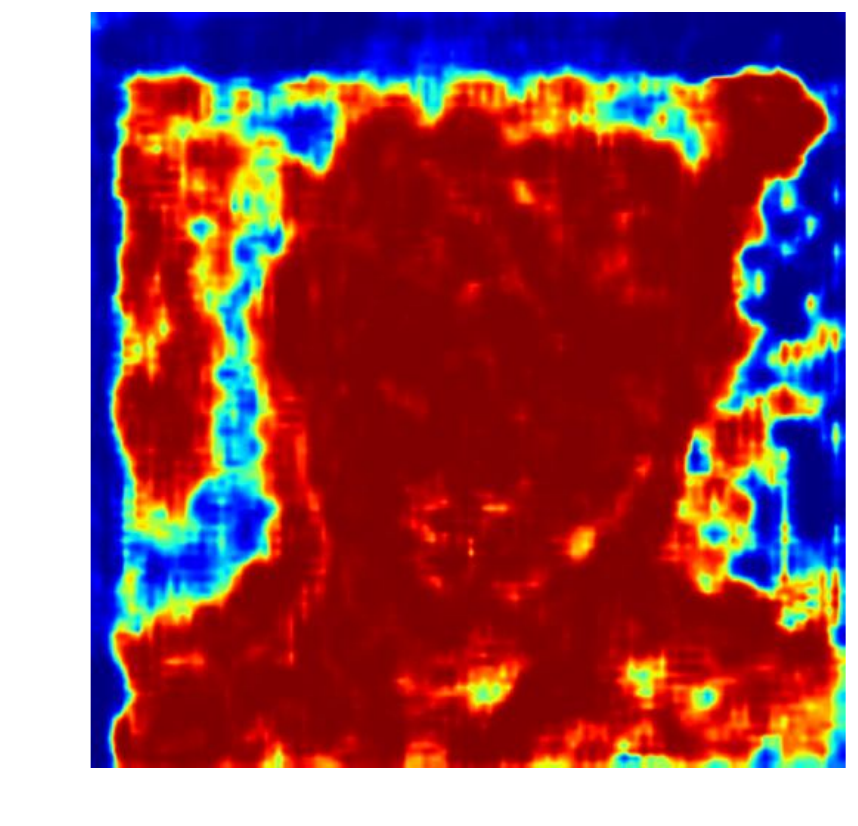}\\
			\hline
			&Image\&GT & Xception & Mesonet & UNet8x & UNet4x&VGG8&VGG5&FN3\\
			\hline
		\end{tabular}
	\end{center}
	\caption{Qualitative results of the classification and segmentation models. Each of two rows relates to a specific manipulation method. For each method, on the left are the input fake image and the ground-truth indicating the manipulated area. The upper row shows the pixel-level results of all the classification models, and the lower row displays the predictions of the segmentation models. (DF: DeepFakes, F2F: Face2Face, FS: FaceSwap, NT: NeuralTextures)}
	\label{fig:comparison}
\end{figure*}

From the results above, obviously segmentation models show superiority over the classification models in terms of both pixel-level prediction and global-level prediction. Figure \ref{fig:comparison} shows outputs of different architectures, which further illustrates the benefit of analyzing fake faces on a pixel-level.

\begin{figure*}[ht]
	\setlength{\abovecaptionskip}{0.cm}
	\setlength{\belowcaptionskip}{-0.1cm}
	\begin{center}
		\scalebox{0.9}{
		\setlength{\tabcolsep}{0.3em}
		\begin{tabular}{ccccc}
			\includegraphics[width=0.15\linewidth]{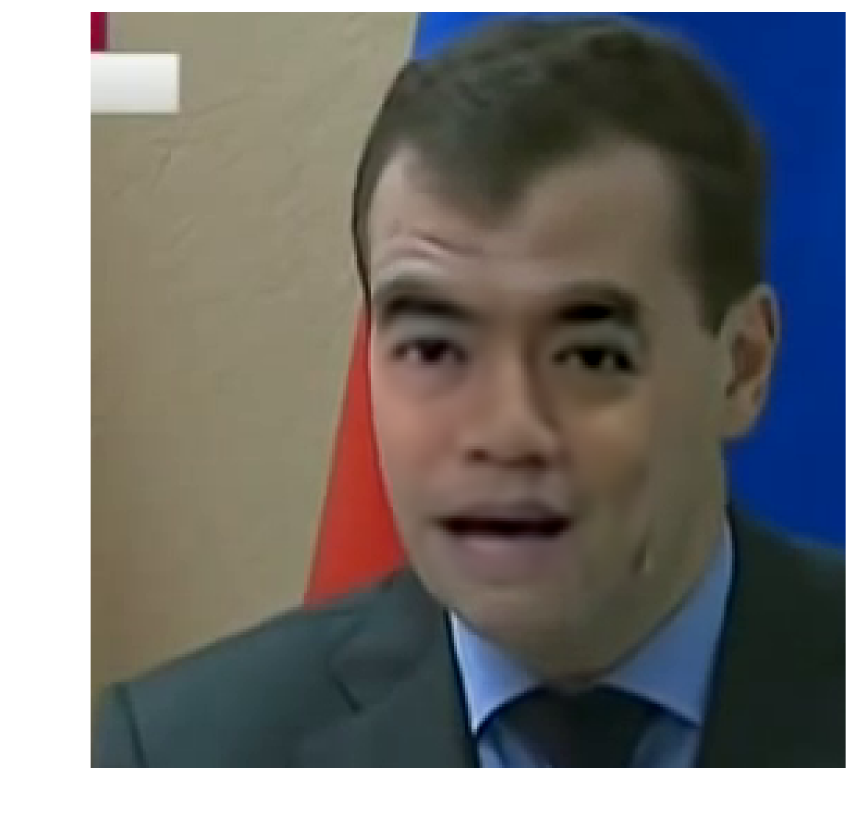} &
			\includegraphics[width=0.15\linewidth]{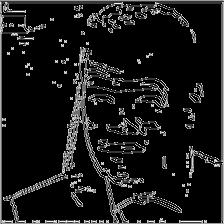} &
			\includegraphics[width=0.15\linewidth]{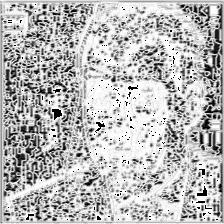} &
			\includegraphics[width=0.15\linewidth]{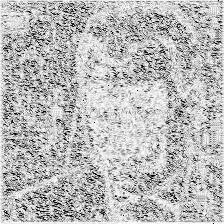} &
			\includegraphics[width=0.15\linewidth]{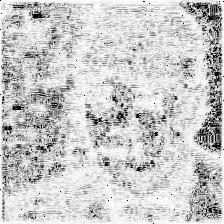} \\
			
			\includegraphics[width=0.15\linewidth]{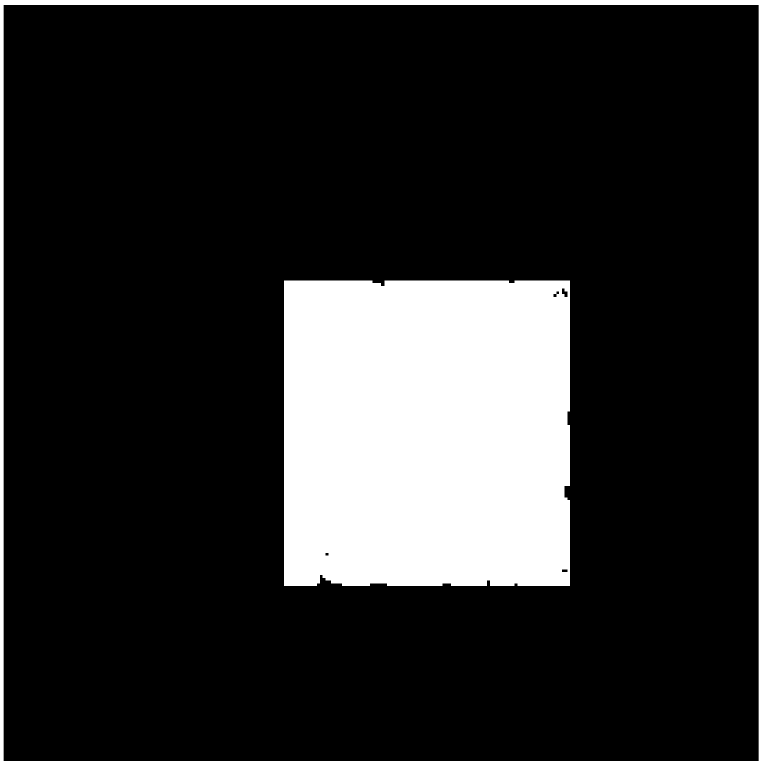} &		
			\includegraphics[width=0.15\linewidth]{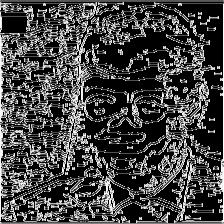} &
			\includegraphics[width=0.15\linewidth]{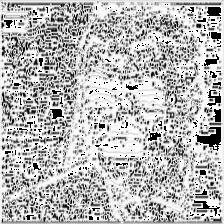} &
			\includegraphics[width=0.15\linewidth]{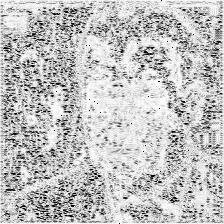} &	
			\includegraphics[width=0.15\linewidth]{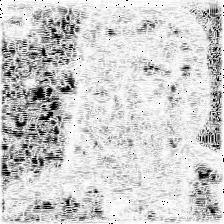} \\
			
			\includegraphics[width=0.15\linewidth]{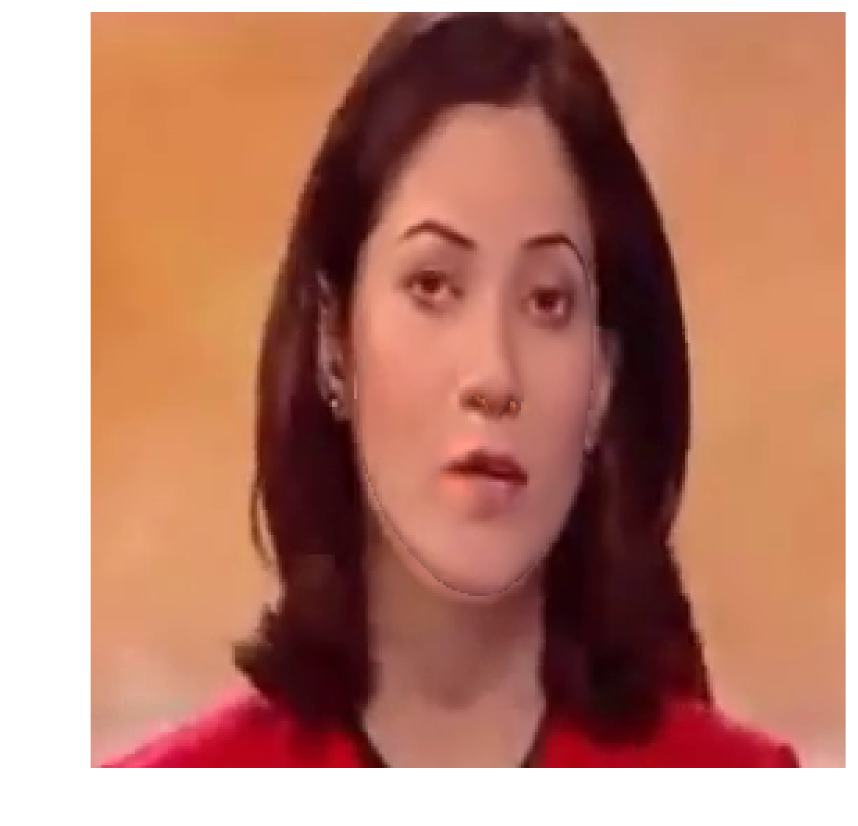} &
			\includegraphics[width=0.15\linewidth]{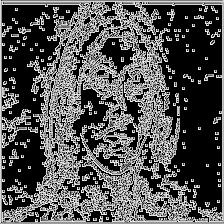} &
			\includegraphics[width=0.15\linewidth]{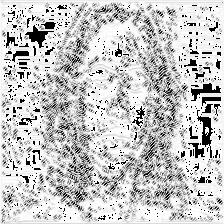} &
			\includegraphics[width=0.15\linewidth]{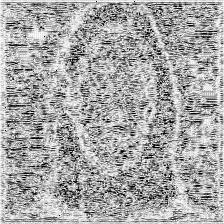} &
			\includegraphics[width=0.15\linewidth]{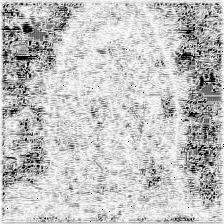} \\
			
			\includegraphics[width=0.15\linewidth]{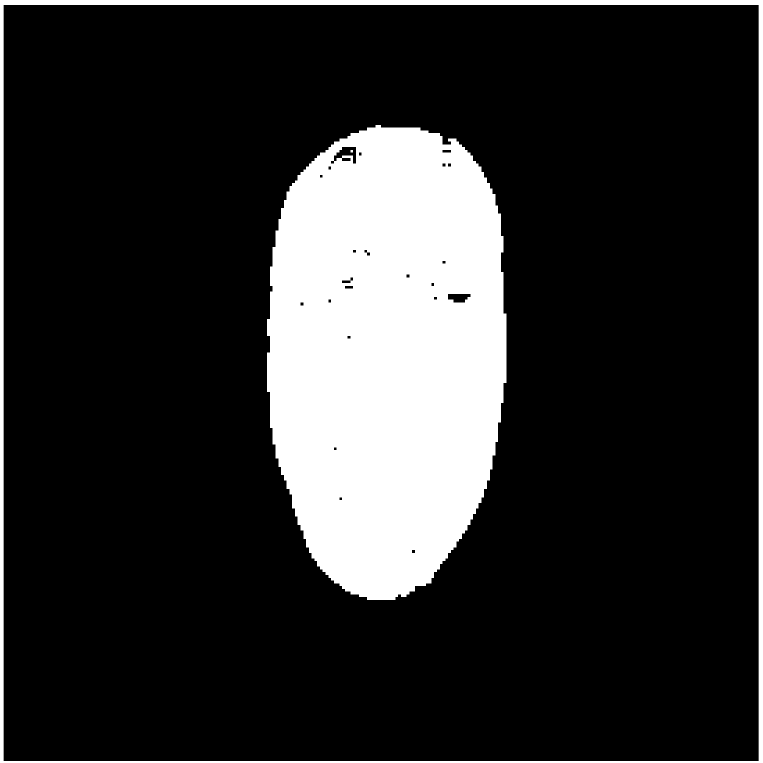} &		
			\includegraphics[width=0.15\linewidth]{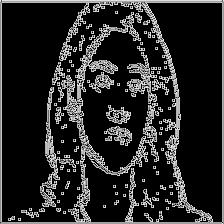} &
			\includegraphics[width=0.15\linewidth]{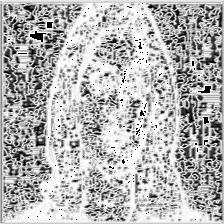} &
			\includegraphics[width=0.15\linewidth]{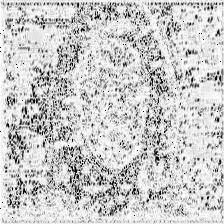} &	
			\includegraphics[width=0.15\linewidth]{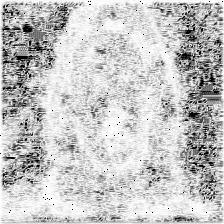} \\	
			Image \& GT & conv1 & conv2 & conv3 & conv4
		\end{tabular}}
	\end{center}
	\caption{Kernel visualization of VGG5. The left column is the input fake image and the ground-truth. Each column on the right shows kernels from a specific convolutional layer.}
	\label{fig:viskernel}
\end{figure*}
\subsection{Analysis}

\noindent\textbf{Deep vs Shallow}

To explore the effect of model depth to the task of face forensics, we also take a closer look at the performance of models with different depth. In Table \ref{tab:shallowdeep}, we summarize the mIoUs of segmentation models with different depth. Apart from VGG8 and VGG5, we also include VGG3, which only uses the first two layers of VGG16 followed by a classifier. It is interesting to see that the deep model, Xception with 36 layers, does not reach to a high score, whereas the shallow models present better abilities. This reveals that face forensics is supposed to be defined as a low-level vision problem than a high-level perception problem. 

\begin{table}[t]
	\renewcommand{\arraystretch}{1.5}
	\begin{center}
		\scalebox{0.7}{
\begin{tabular}{|c|cccccc|}

	\hline 
	& DF & F2F & FS & NT & P & Avg\tabularnewline
	\hline 
	\hline 
	Xception (36) & 89.32 & 88.18 & 87.7 & 62.81 & \textbf{99.95} & 85.59\tabularnewline

	VGG8 (7) & 94.68 & 95.21 & 94.33 & \textbf{76.04} & 99.31 & 91.91\tabularnewline

	VGG5 (4) & \textbf{95.78} & \textbf{96.21} & \textbf{94.81} & 75.6 & 99.86 & \textbf{92.45}\tabularnewline

	FN3 (3) & 92.68 & 93.05 & 89.01 & 64.42 & 99.72 & 87.78\tabularnewline

	VGG3 (3) & 88.79 & 89.92 & 79.65 & 57.93 & 96.64 & 82.58\tabularnewline
	\hline 
\end{tabular}}
\end{center}
\caption{Comparison among models with different depth. The number in the parentheses indicates the depth of the model. The numbers are mIoU.}
\label{tab:shallowdeep}
\end{table}

\noindent\textbf{Pretrained or From Scratch}

As implied by the analysis in the last section, face forensics is more like a low-level vision task. Another question is that ``can the models benefit from the features used for general vision recognition tasks?'' We conduct another ablation study where we compare the performance on the segmentation task using models with and without ImageNet-pretraining. The results are shown in Table \ref{tab:pretrainedscratch}. According to the numbers, there is little difference between the pretrained model and the trained-from-scratch model.  The features learned in a general vision recognition task such as ImageNet did not help quickly find a better local optima. 

\begin{table}[ht]
	\renewcommand{\arraystretch}{1.5}
	\begin{center}
		\scalebox{0.7}{
\begin{tabular}{|c|ccccc|c|}
	\hline 
	& DF & F2F & FS & NT & P & Avg\tabularnewline
	\hline 
	\hline 
	Xception (pretrained) & 89.32 & 88.18 & 87.7 & 62.81 & 99.95 & \textbf{85.59}\tabularnewline
	Xception (non-pretrained) & 88.72 & 87.88 & 88.70 & 62.84 & 99.74 & 85.57\tabularnewline
	\hline 
	VGG5 (pretrained) & 95.78 & 96.21 & 94.81 & 75.6 & 99.86 & \textbf{92.45}\tabularnewline
	VGG5 (non-pretrained) & 95.69 & 96.2 & 94.75 & 75.35 & 99.86 & 92.37\tabularnewline
	\hline 
	VGG8 (pretrained) & 94.68 & 95.21 & 94.33 & 76.04 & 99.31 & 91.91\tabularnewline
	VGG8 (non-pretrained) & 95.67 & 95.93 & 95.06 & 75.18 & 99.83 & \textbf{92.33}\tabularnewline
	\hline 
\end{tabular}}
\end{center}
\caption{Comparison between a pretrained model and the model trained from scratch. The numbers are mIoU.}
\label{tab:pretrainedscratch}
\end{table}

\noindent\textbf{Kernel visualization}

In order to have a better understanding of the features learned by the model, we analyze the kernels by visualizing them using the technique in \cite{DB15a}.  In Figure \ref{fig:viskernel}, for each fake image, we visualize two kernels in each convolutional layer. Apart from the features in conv1, which are mostly low-level edges and corners, the kernels in following layers do not make much sense to us humans. Intuitively, the model tries to learn subtle features, to which humans are not sensitive to. Humans are good at recognizing things on a semantic level, but fake faces, generated by advanced manipulation methods, seem beyond humans ability. This further emphasizes the demand of a good face forensics model.

\section{Conclusion}

Face forensics has become increasingly important as face manipulation methods have made stunning progress to enable effortless generation of indistinguishable fake face images. Most previous works cast the problem as a classification task, which suffers from limitations. In this paper, we analyze the problem from pixel-level perspective by using segmentation methods to complement the traditional classification methods. With comprehensive experiments, we show the superiority of formulating it as a segmentation problem instead of a classification problem. In addition, we also perform different ablation studies to analyze the important factors of being an effective face forensics model, which establishes a strong new baseline for the benchmark. We hope that our analysis can provide more insight to the field of face forensics. 

%

\newpage

{\small
\bibliographystyle{ieee_fullname}
\bibliography{egbib}

\begin{thebibliography}{10}\itemsep=-1pt

\bibitem{deepfake}
Deepfakes github.
\newblock \url{https://github.com/deepfakes/faceswap}.

\bibitem{facerecognition}
facerecognition github.
\newblock \url{https://github.com/ageitgey/face_recognition}.

\bibitem{Faceswap}
Faceswap.
\newblock \url{https://github.com/MarekKowalski/FaceSwap}.

\bibitem{mesonet}
D. {Afchar}, V. {Nozick}, J. {Yamagishi}, and I. {Echizen}.
\newblock Mesonet: a compact facial video forgery detection network.
\newblock In {\em 2018 IEEE International Workshop on Information Forensics and
  Security (WIFS)}, pages 1--7, Dec 2018.

\bibitem{2dVideo_face_replacement}
Mahmoud Afifi, Khaled Hussain, Hosny Ibrahim, and Nagwa Omar.
\newblock Video face replacement system using a modified poisson blending
  technique.
\newblock {\em 2014 International Symposium on Intelligent Signal Processing
  and Communication Systems, ISPACS 2014}, 12 2014.

\bibitem{face_age}
G. {Antipov}, M. {Baccouche}, and J. {Dugelay}.
\newblock Face aging with conditional generative adversarial networks.
\newblock In {\em 2017 IEEE International Conference on Image Processing
  (ICIP)}, pages 2089--2093, Sep. 2017.

\bibitem{elor2017bringingPortraits}
Hadar Averbuch-Elor, Daniel Cohen-Or, Johannes Kopf, and Michael~F. Cohen.
\newblock Bringing portraits to life.
\newblock {\em ACM Transactions on Graphics (Proceeding of SIGGRAPH Asia
  2017)}, 36(6):196, 2017.

\bibitem{Double_JPEG_Detection}
Mauro Barni, Luca Bondi, Nicolò Bonettini, Paolo Bestagini, Andrea Costanzo,
  Marco Maggini, Benedetta Tondi, and Stefano Tubaro.
\newblock Aligned and non-aligned double jpeg detection using convolutional
  neural networks.
\newblock {\em Journal of Visual Communication and Image Representation}, 49,
  08 2017.

\bibitem{new_cnn_Bayar:2016}
Belhassen Bayar and Matthew~C. Stamm.
\newblock A deep learning approach to universal image manipulation detection
  using a new convolutional layer.
\newblock In {\em Proceedings of the 4th ACM Workshop on Information Hiding and
  Multimedia Security}, IH\&\#38;MMSec '16, pages 5--10, New York, NY, USA,
  2016. ACM.

\bibitem{Video_Rewrite_Bregler}
Christoph Bregler, Michele Covell, and Malcolm Slaney.
\newblock Video rewrite: Driving visual speech with audio.
\newblock In {\em Proceedings of the 24th Annual Conference on Computer
  Graphics and Interactive Techniques}, SIGGRAPH '97, pages 353--360, New York,
  NY, USA, 1997. ACM Press/Addison-Wesley Publishing Co.

\bibitem{Chollet_2017_CVPR}
Francois Chollet.
\newblock Xception: Deep learning with depthwise separable convolutions.
\newblock In {\em The IEEE Conference on Computer Vision and Pattern
  Recognition (CVPR)}, July 2017.

\bibitem{xception_2017_CVPR}
Francois Chollet.
\newblock Xception: Deep learning with depthwise separable convolutions.
\newblock In {\em The IEEE Conference on Computer Vision and Pattern
  Recognition (CVPR)}, July 2017.

\bibitem{Weakly_Supervised_Object_Localization}
R.~G. {Cinbis}, J. {Verbeek}, and C. {Schmid}.
\newblock Weakly supervised object localization with multi-fold multiple
  instance learning.
\newblock {\em IEEE Transactions on Pattern Analysis and Machine Intelligence},
  39(1):189--203, Jan 2017.

\bibitem{Cordts2016Cityscapes}
Marius Cordts, Mohamed Omran, Sebastian Ramos, Timo Rehfeld, Markus Enzweiler,
  Rodrigo Benenson, Uwe Franke, Stefan Roth, and Bernt Schiele.
\newblock The cityscapes dataset for semantic urban scene understanding.
\newblock In {\em Proc. of the IEEE Conference on Computer Vision and Pattern
  Recognition (CVPR)}, 2016.

\bibitem{Illumination_Color}
T.~J. d. {Carvalho}, C. {Riess}, E. {Angelopoulou}, H. {Pedrini}, and A. d.
  R.~{Rocha}.
\newblock Exposing digital image forgeries by illumination color
  classification.
\newblock {\em IEEE Transactions on Information Forensics and Security},
  8(7):1182--1194, July 2013.

\bibitem{3dVideo_face_replacement}
Kevin Dale, Kalyan Sunkavalli, Micah~K. Johnson, Daniel Vlasic, Wojciech
  Matusik, and Hanspeter Pfister.
\newblock Video face replacement.
\newblock In {\em Proceedings of the 2011 SIGGRAPH Asia Conference}, SA '11,
  pages 130:1--130:10, New York, NY, USA, 2011. ACM.

\bibitem{expressions_variation}
D. {Dang-Nguyen}, G. {Boato}, and F.~G.~B. {De Natale}.
\newblock Identify computer generated characters by analysing facial
  expressions variation.
\newblock In {\em 2012 IEEE International Workshop on Information Forensics and
  Security (WIFS)}, pages 252--257, Dec 2012.

\bibitem{Farid:2016}
Hany Farid.
\newblock {\em Photo Forensics}.
\newblock The MIT Press, 2016.

\bibitem{Vdub}
P. Garrido, Levi Valgaerts, Hamid Sarmadi, I. Steiner, Kiran Varanasi, P.
  Pérez, and C. Theobalt.
\newblock Vdub: Modifying face video of actors for plausible visual alignment
  to a dubbed audio track.
\newblock {\em Computer Graphics Forum}, 34, 05 2015.

\bibitem{rnn_tamper}
D. {Güera} and E.~J. {Delp}.
\newblock Deepfake video detection using recurrent neural networks.
\newblock In {\em 2018 15th IEEE International Conference on Advanced Video and
  Signal Based Surveillance (AVSS)}, pages 1--6, Nov 2018.

\bibitem{face_rotation_Huang_2017_ICCV}
Rui Huang, Shu Zhang, Tianyu Li, and Ran He.
\newblock Beyond face rotation: Global and local perception gan for
  photorealistic and identity preserving frontal view synthesis.
\newblock In {\em The IEEE International Conference on Computer Vision (ICCV)},
  Oct 2017.

\bibitem{fight_fake_2018_ECCV}
Minyoung Huh, Andrew Liu, Andrew Owens, and Alexei~A. Efros.
\newblock Fighting fake news: Image splice detection via learned
  self-consistency.
\newblock In {\em The European Conference on Computer Vision (ECCV)}, September
  2018.

\bibitem{pix2pix2016}
Phillip Isola, Jun-Yan Zhu, Tinghui Zhou, and Alexei~A Efros.
\newblock Image-to-image translation with conditional adversarial networks.
\newblock {\em arxiv}, 2016.

\bibitem{kim2018DeepVideo}
H. Kim, P. Garrido, A. Tewari, W. Xu, J. Thies, N. Nie{\ss}ner, P. P{\'e}rez,
  C. Richardt, M. Zollh{\"o}fer, and C. Theobalt.
\newblock {Deep Video Portraits}.
\newblock {\em ACM Transactions on Graphics 2018 (TOG)}, 2018.

\bibitem{kingma:adam}
Diederick~P Kingma and Jimmy Ba.
\newblock Adam: A method for stochastic optimization.
\newblock In {\em International Conference on Learning Representations (ICLR)},
  2015.

\bibitem{eye_blinking}
Y. {Li}, M. {Chang}, and S. {Lyu}.
\newblock In ictu oculi: Exposing ai created fake videos by detecting eye
  blinking.
\newblock In {\em 2018 IEEE International Workshop on Information Forensics and
  Security (WIFS)}, pages 1--7, Dec 2018.

\bibitem{FCN_CVPR2015}
J. {Long}, E. {Shelhamer}, and T. {Darrell}.
\newblock Fully convolutional networks for semantic segmentation.
\newblock In {\em 2015 IEEE Conference on Computer Vision and Pattern
  Recognition (CVPR)}, pages 3431--3440, June 2015.

\bibitem{attribute_cyclegan_Lu_2018_ECCV}
Yongyi Lu, Yu-Wing Tai, and Chi-Keung Tang.
\newblock Attribute-guided face generation using conditional cyclegan.
\newblock In {\em The European Conference on Computer Vision (ECCV)}, September
  2018.

\bibitem{DBN2012CVPR}
Ping Luo.
\newblock Hierarchical face parsing via deep learning.
\newblock In {\em Proceedings of the 2012 IEEE Conference on Computer Vision
  and Pattern Recognition (CVPR)}, CVPR '12, pages 2480--2487, Washington, DC,
  USA, 2012. IEEE Computer Society.

\bibitem{Obrien:2012:EPM}
James~F. O'Brien and Hany Farid.
\newblock Exposing photo manipulation with inconsistent reflections.
\newblock {\em ACM Transactions on Graphics}, 31(1):4:1--11, Jan. 2012.
\newblock Presented at SIGGRAPH 2012.

\bibitem{Mid_level_Image_Representations}
M. {Oquab}, L. {Bottou}, I. {Laptev}, and J. {Sivic}.
\newblock Learning and transferring mid-level image representations using
  convolutional neural networks.
\newblock In {\em 2014 IEEE Conference on Computer Vision and Pattern
  Recognition}, pages 1717--1724, June 2014.

\bibitem{paszke2017automatic}
Adam Paszke, Sam Gross, Soumith Chintala, Gregory Chanan, Edward Yang, Zachary
  DeVito, Zeming Lin, Alban Desmaison, Luca Antiga, and Adam Lerer.
\newblock Automatic differentiation in {PyTorch}.
\newblock In {\em NIPS Autodiff Workshop}, 2017.

\bibitem{Popescu:2005}
A.C. Popescu and H. Farid.
\newblock Exposing digital forgeries by detecting traces of resampling.
\newblock {\em Trans. Sig. Proc.}, 53(2):758--767, Feb. 2005.

\bibitem{print_scanned_morphed}
R. {Raghavendra}, K.~B. {Raja}, S. {Venkatesh}, and C. {Busch}.
\newblock Transferable deep-cnn features for detecting digital and
  print-scanned morphed face images.
\newblock In {\em 2017 IEEE Conference on Computer Vision and Pattern
  Recognition Workshops (CVPRW)}, pages 1822--1830, July 2017.

\bibitem{unet_RFB15a}
O. Ronneberger, P.Fischer, and T. Brox.
\newblock U-net: Convolutional networks for biomedical image segmentation.
\newblock In {\em Medical Image Computing and Computer-Assisted Intervention
  (MICCAI)}, volume 9351 of {\em LNCS}, pages 234--241. Springer, 2015.
\newblock (available on arXiv:1505.04597 [cs.CV]).

\bibitem{2019faceforensicspp}
Andreas Rossler, Davide Cozzolino, Luisa Verdoliva, Christian Riess, Justus
  Thies, and Matthias Niesner.
\newblock Face{F}orensics++: Learning to detect manipulated facial images.
\newblock In {\em International Conference on Computer Vision (ICCV)}, 2019.

\bibitem{schroff2015facenet}
Florian Schroff, Dmitry Kalenichenko, and James Philbin.
\newblock Facenet: A unified embedding for face recognition and clustering.
\newblock In {\em CVPR}, 2015.

\bibitem{grad_cam_2017_ICCV}
Ramprasaath~R. Selvaraju, Michael Cogswell, Abhishek Das, Ramakrishna Vedantam,
  Devi Parikh, and Dhruv Batra.
\newblock Grad-cam: Visual explanations from deep networks via gradient-based
  localization.
\newblock In {\em The IEEE International Conference on Computer Vision (ICCV)},
  Oct 2017.

\bibitem{vgg}
K. Simonyan and A. Zisserman.
\newblock Very deep convolutional networks for large-scale image recognition.
\newblock In {\em International Conference on Learning Representations}, May
  2015.

\bibitem{DB15a}
J.T. Springenberg, A. Dosovitskiy, T. Brox, and M. Riedmiller.
\newblock Striving for simplicity: The all convolutional net.
\newblock In {\em ICLR (workshop track)}, 2015.

\bibitem{sun2014deep}
Yi Sun, Yuheng Chen, Xiaogang Wang, and Xiaoou Tang.
\newblock Deep learning face representation by joint
  identification-verification.
\newblock In {\em Advances in Neural Information Processing Systems}, pages
  1988--1996, 2014.

\bibitem{Synthesize}
Supasorn Suwajanakorn, Steven Seitz, and Ira Kemelmacher.
\newblock Synthesizing obama: learning lip sync from audio.
\newblock {\em ACM Transactions on Graphics}, 36:1--13, 07 2017.

\bibitem{Suwajanakorn:2017}
Supasorn Suwajanakorn, Steven~M. Seitz, and Ira Kemelmacher-Shlizerman.
\newblock Synthesizing obama: Learning lip sync from audio.
\newblock {\em ACM Trans. Graph.}, 36(4):95:1--95:13, July 2017.

\bibitem{inception_CVPR}
Christian Szegedy, Wei Liu, Yangqing Jia, Pierre Sermanet, Scott Reed, Dragomir
  Anguelov, Dumitru Erhan, Vincent Vanhoucke, and Andrew Rabinovich.
\newblock Going deeper with convolutions.
\newblock In {\em The IEEE Conference on Computer Vision and Pattern
  Recognition (CVPR)}, June 2015.

\bibitem{taigman2013deepface}
Yaniv Taigman, Ming Yang, Marc'Aurelio Ranzato, and Lior Wolf.
\newblock Deepface: Closing the gap to human-level performance in face
  verification.
\newblock In {\em Proceedings of the IEEE Conference on Computer Vision and
  Pattern Recognition}, pages 1701--1708, 2013.

\bibitem{2019Neural_Textures}
Justus Thies, Michael Zollh\"ofer, and Matthias Niessner.
\newblock Deferred neural rendering: Image synthesis using neural textures.
\newblock {\em ACM Trans. Graph.}, 38(4):66:1--66:12, July 2019.

\bibitem{Thies_2016_CVPR}
Justus Thies, Michael Zollhofer, Marc Stamminger, Christian Theobalt, and
  Matthias Niessner.
\newblock Face2face: Real-time face capture and reenactment of rgb videos.
\newblock In {\em The IEEE Conference on Computer Vision and Pattern
  Recognition (CVPR)}, June 2016.

\bibitem{Feature_Interpolation2017deep}
Paul Upchurch, Jacob Gardner, Geoff Pleiss, Robert Pless, Noah Snavely, Kavita
  Bala, and Kilian~Q. Weinberger.
\newblock Deep feature interpolation for image content changes.
\newblock In {\em The IEEE Conference on Computer Vision and Pattern
  Recognition (CVPR)}, July 2017.

\bibitem{PS2019detecting}
Sheng-Yu Wang, Oliver Wang, Andrew Owens, Richard Zhang, and Alexei~A Efros.
\newblock Detecting photoshopped faces by scripting photoshop.
\newblock {\em arXiv preprint arXiv:1906.05856}, 2019.

\bibitem{Zhou_2016_CVPR}
Bolei Zhou, Aditya Khosla, Agata Lapedriza, Aude Oliva, and Antonio Torralba.
\newblock Learning deep features for discriminative localization.
\newblock In {\em The IEEE Conference on Computer Vision and Pattern
  Recognition (CVPR)}, June 2016.

\bibitem{two_stream_tamper}
P. {Zhou}, X. {Han}, V.~I. {Morariu}, and L.~S. {Davis}.
\newblock Two-stream neural networks for tampered face detection.
\newblock In {\em 2017 IEEE Conference on Computer Vision and Pattern
  Recognition Workshops (CVPRW)}, pages 1831--1839, July 2017.

\bibitem{2018face_star}
Michael Zollhofer, Justus Thies, Pablo Garrido, Derek Bradley, Thabo Beeler,
  Patrick Pérez, Marc Stamminger, Matthias Nießner, and Christian Theobalt.
\newblock {State of the Art on Monocular 3D Face Reconstruction, Tracking, and
  Applications}.
\newblock {\em Computer Graphics Forum}, 2018.

\end{thebibliography}
}

\end{document}